\documentclass[table, preprint]{article}

\usepackage{ifthen}

\newboolean{isArXiv}

            \setboolean{isArXiv}{true}           

\ifthenelse{\boolean{isArXiv}}
{
  \usepackage{neurips_2023}
  
  \author{%
      Narun Raman \\
      \parbox{.45\textwidth}{\centering University of British Columbia}\\
      \texttt{narunram@cs.ubc.ca} \\
      \And
      Taylor Lundy \\
      \parbox{.45\textwidth}{\centering University of British Columbia} \\
      \texttt{tlundy@cs.ubc.ca} \\
      \And  Samuel Amouyal\\
      \parbox{.45\textwidth}{\centering Tel Aviv University}  \\
      \texttt{samuel.amouyal@cs.tau.ac.il} \\
      \And
      Yoav Levine \\
      \parbox{.45\textwidth}{\centering Stanford University \& AI21 Labs}\\
      \texttt{yoavl@stanford.edu} \\
      \And
      Kevin Leyton-Brown \\
      \parbox{.45\textwidth}{\centering University of British Columbia \& AI21 Labs}\\
      \texttt{kevinlb@cs.ubc.ca}
      \And
      Moshe Tennenholtz \\
      \parbox{.45\textwidth}{\centering Technion \& AI21 Labs} \\
      \texttt{moshet@technion.ac.il} \\
  }
  \title{STEER: Assessing the \\Economic Rationality of Large Language Models}
}
{
  \usepackage{icml2024}
  
  \icmltitlerunning{STEER}
}
\newif\ifcomments
\commentstrue
\ifcomments
    \providecommand{\narun}[1]{{\protect\color{blue}{[Narun: #1]}}}
    \providecommand{\sa}[1]{{\protect\color{teal}{[Samuel: #1]}}} 
    \providecommand{\tl}[1]{{\protect\color{magenta}{[Taylor: #1]}}} 
    \providecommand{\moshe}[1]{{\protect\color{violet}{[Moshe: #1]}}} 
    \providecommand{\klb}[1]{{\protect\color{red}{[Kevin: #1]}}} 
    \providecommand{\yoav}[1]{{\protect\color{orange}{[Yoav: #1]}}} 
\else
    \providecommand{\narun}[1]{}
    \providecommand{\sa}[1]{} 
    \providecommand{\tl}[1]{}
    \providecommand{\moshe}[1]{} 
    \providecommand{\klb}[1]{} 
    \providecommand{\yoav}[1]{} 
\fi

\usepackage[utf8]{inputenc} 
\usepackage[T1]{fontenc}    
\usepackage{hyperref}       
\usepackage{url}            
\usepackage{booktabs}       
\usepackage{amsfonts}       
\usepackage{nicefrac}       
\usepackage{microtype}      
\usepackage{xspace}
\usepackage{xcolor}         
\usepackage{graphicx}
\usepackage{amssymb}
\usepackage{natbib}
\usepackage{fontawesome}

\usepackage{chngcntr}
\newcounter{element}
\counterwithin*{element}{subsubsection}

\newtheorem{element}{Element}

\newtheorem{illustration}{Illustration}[section]

\newcommand{\Parent}{Setting\xspace}
\newcommand{\parent}{setting\xspace}
\newcommand{\Child}{Module\xspace}
\newcommand{\child}{module\xspace}
\newcommand{\parentSec}[1]{\subsection{\textsc{\Parent \arabic{subsection}: #1}}}
\newcommand{\childSec}[1]{\subsubsection{\textsc{\Child \arabic{subsection}.\arabic{subsubsection}: #1}}}
\newcommand{\parents}{{\parent}s\xspace}
\newcommand{\children}{{\child}s\xspace}

\newcommand{\firstParent}{\textsc{Foundations}\xspace}
\newcommand{\secondParent}{\textsc{Decisions in Single-Agent Environments}\xspace}
\newcommand{\thirdParent}{\textsc{Decisions in Multi-Agent Environments}\xspace}
\newcommand{\fourthParent}{\textsc{Decisions on Behalf of Other Agents}\xspace}

\newcommand{\character}{agent\xspace}
\newcommand{\acharacter}{an agent\xspace}

\newcommand{\rc}{SRC\xspace}
\newcommand{\rcs}{SRCs\xspace}
\newcommand{\reportcard}{STEER Report Card\xspace}

\usepackage{multiaudience}
\SetNewAudience{arxiv}
\SetNewAudience{icml}

\usepackage{comment}
\usepackage{tikz}
\usepackage{pgfplots}
\usepackage{pgfplotstable}
\pgfplotsset{compat=1.7}
\usepackage{array,multirow,multicol}
\usepackage{enumitem}
\usepackage{xcolor}
\usepackage{color}
\usepackage{amsmath}
\usepackage{cleveref}
\usepackage{nameref}
\usepackage{pdfpages}
\usepackage{caption}
\usepackage{subcaption}
\usepackage{wrapfig}
\usepackage{csquotes}

\newcommand*{\fullref}[1]{\hyperref[{#1}]{\ref*{#1} \nameref*{#1}}} 

\ifthenelse{\boolean{isArXiv}}{
    \newenvironment{exampleref}[1]{%
    \let\egtheillustration=\theillustration%
    \renewcommand{\theillustration}{\ref*{#1}}
    \begin{illustration}
    }
    {\end{illustration}
    \let\theillustration\egtheillustration
    \addtocounter{illustration}{-1}
    }
}{
    \newenvironment{exampleref}[1]{%
    \let\egtheillustration=\theillustration%
    \renewcommand{\theillustration}{\ref*{#1}}
    \begin{illustration}
    }
    {\end{illustration}
    \let\theillustration\egtheillustration
    \addtocounter{illustration}{-1}
}
}

\makeatletter
\def\@opargbegintheorem#1#2#3{\trivlist
\item[]{\bfseries #1\ #2\ (#3)} \itshape}
\makeatother

\begin{document}

\begin{shownto}{icml}
\twocolumn[
  \icmltitle{STEER: \\Assessing the Economic Rationality of Large Language Models}
  \begin{icmlauthorlist}
      \icmlauthor{Narun Raman}{yyy}
      \icmlauthor{Taylor Lundy}{yyy}
      \icmlauthor{Samuel Joseph Amouyal}{sch}
      \icmlauthor{Yoav Levine}{comp}
      \icmlauthor{Kevin Leyton-Brown}{yyy}
      \icmlauthor{Moshe Tennenholtz}{tech}
  \end{icmlauthorlist}

    \icmlaffiliation{yyy}{Department of Computer Science, University of British Columbia, Vancouver, Canada}
    \icmlaffiliation{comp}{Stanford & AI21 Labs, Palo Alto, California, United States}
    \icmlaffiliation{sch}{Tel Aviv University, Aviv, Israel}
    \icmlaffiliation{tech}{Technion & AI21 Labs, Tel Aviv, Israel}
    
    \icmlcorrespondingauthor{Narun Raman}{narunram@cs.ubc.ca}
    \icmlcorrespondingauthor{Taylor Lundy}{tlundy@cs.ubc.ca}

]
\end{shownto}
\begin{shownto}{arxiv}
    \maketitle
\end{shownto}

\begin{abstract}
There is increasing interest in using LLMs as decision-making ``agents.'' Doing so includes many degrees of freedom: which model should be used; how should it be prompted; should it be asked to introspect, conduct chain-of-thought reasoning, etc? Settling these questions---and more broadly, determining whether an LLM agent is reliable enough to be trusted---requires a methodology for assessing such an agent's economic rationality. In this paper, we provide one. We begin by surveying the economic literature on rational decision making, taxonomizing a large set of fine-grained ``elements'' that an agent should exhibit, along with dependencies between them. We then propose a benchmark distribution called STEER (Systematic and Tuneable Evaluation of Economic Rationality) that quantitatively scores an LLMs performance on these elements and, combined with a user-provided rubric, produces a ``STEER report card.'' Finally, we describe the results of a large-scale empirical experiment with 14 different LLMs, characterizing the both current state of the art and the impact of different model sizes on models' ability to exhibit rational behavior.
\end{abstract}

\section{Introduction}
\label{sec:intro}

Recently, much research has worked to leverage Large Language Models (LLMs) to create decision-making engines, configuring them either to act directly as economic agents \citep{cai2023large, homo_silicus, voyager} or to serve as key elements of broader systems that do so \citep{zhuge2023mindstorms, wang2023unleashing, huggingGPT}. 
LLM-based agents are already showing strength in planning (e.g., for personal finance \citep{agentGPT}), solving complex problems  \citep[e.g., medical diagnostics;][]{mcduff2023towards}, leveraging tools \citep[e.g.,][]{toolformer} and playing games \citep[e.g., chess;][]{babyagi}.
Better decision-making capabilities will be critical for advancing the use of  Reinforcement Learning from AI Feedback (RLAIF) to fine-tune chatbots, such as constitution-based approaches \citep{bai2022constitutional, metagpt}.
LLMs may soon take the place of humans in some social science experiments \citep{homo_silicus,aher2023using, park2023generative}.
Eventually, this research agenda offers the promise of realizing the longstanding AI dream of personalized economic agents. 

How can we best configure LLMs to maximize their performance on decision-making tasks (e.g., via prompting to perform chain-of-thought reasoning; fine-tuning; or more complex architectures that make repeated calls to a model)? 
After we have done so, how well do LLM-based agents perform?
Recent research has begun to develop testing regimes that can address these questions in various restricted domains. 
These include various narrowly defined tasks \citep{agentbench, toolbench, webarena,react}; 
limited economic settings \citep{flue, finqa_benchmark, sent_analysis_benchmark_1, finbert, akata2023playing}; and open-world video games \citep{voyager, ghost_in_minecraft}.
Going beyond such problem-specific approaches to assessing decision making more broadly requires a holistic approach to describing good decision making and explaining how it can be decomposed into different, individually testable components.
One approach is to divide decision making into distinct, ad hoc tasks, emphasizing those that have been clasically studied in NLP \citep{liang2022holistic, gehrmann-etal-2021-gem}.

We advocate a different approach: enumerating first principles that describe how agents \emph{should} make decisions, and then evaluating an agent's degree of adherence with these principles.
Answering the normative question of how decision-makers should act has been the focus of more than a century of research in economics, cognitive psychology, computer science, and philosophy. 
The resulting literature provides a mature mathematical foundation for so-called economic rationality. 
The grounding principle is that agents should (implicitly or explicitly) quantify their preferences according to a utility function and make decisions to maximize their own expected utility. 
The literature further characterizes elements of economic settings that fundamentally impact rational behavior: e.g., stochasticity is different from determinism; multi-agent environments are different from single-agent environments; reasoning about how best to make decisions for groups of agents is different from reasoning about how to act as an individual. In some cases, the theory is prescriptive: e.g., it's better to maximize utility than to accept lower-utility alternatives. In other cases, things become more complicated: e.g., in multi-agent environments, determining the best choice depends on beliefs about how other agents will act. In still others, impossibility results rule out all desirable options, e.g., when deciding how to aggregate multiple agents' preferences. Finally, in some cases human decision-makers exhibit cognitive biases that deviate from rational behavior, even when the theory makes a firmly prescriptive recommendation.

How can we hope to assess rationality when the landscape is so complex? 
Our approach is to identify tests for which the ``rational'' answer is well defined. In cases where the prescriptive recommendation is clear, assessment is unproblematic. More ambiguous settings can be tested by explicitly asking for a desired behavior (e.g., eliciting a Nash equilibrium strategy). Such settings also often admit prescriptive special cases: e.g., even in multi-agent environments, it is never a good idea to play a dominated strategy. Axiomatic theories naturally give rise to meaningful tests, which can be applied from the simplest settings (e.g., the von Neumann--Morgenstern axioms for utility maximization) to the most complex (Arrow's axioms for social choice functions). Famous human subject experiments that illustrate cognitive biases also naturally give rise to tests, which human decision-makers often fail. In the end, we obtain a \emph{STEER Report Card (SRC)}. We leave it to the end user to determine the scoring rubric: e.g., should the agent receive good grades for doing well only in a subset of simple settings; for being as rational as possible across the board; or for behaving as humanly as possible, including replicating biases?

More specifically, our work  begins by identifying a rich and hierarchical taxonomy of 64 ``elements of rationality'' for which some notion of a ``right answer'' is well defined (Section~\ref{sec:elements-of-rationality}). 
We define each element, giving an example of each in an appendix.
We then move on (in Section~\ref{sec:benchmark}) to describe how we used this taxonomy to derive  a fine-grained benchmark distribution that serves as the basis for STEER report cards. Our benchmark allows each element of rationality to be instantiated in  multiple ``grade levels'' of difficulty and in multiple domains (e.g., asking questions about finance vs.\ medicine). 
For 49 elements, we have written LLM prompts to synthetically generate 24,500 multiple-choice questions and manually validated 2,450 generations in total.
We also discuss how \rcs can be graded.
We built a web interface for generating benchmark questions,  validating them, and visualizing experimental results; it allows elements to be filtered by position in the taxonomy, by logical dependence (e.g., the ``maximize utility'' element depends on the ``transitivity'' element), by domain, and by grade level. 
To demonstrate the utility of our system, we generated full \rcs for 14 language models, ranging from Llama 7B to GPT-4 Turbo, evaluated on 134,750 test questions. This experimental setup is described in Section~\ref{sec:exp-setup}. 
We spent \$4,800 making calls to OpenAI's API and devoted 13,240 GPU hours of compute to evaluating open models.
Section~\ref{sec:results} describes our results; here are some highlights.
Across our benchmark, we found that model size  correlated heavily with performance: models smaller than 40 billion parameters were not reliably able to outperform random guessing. 
Model performance consistently decreased with grade level. GPT-4 Turbo was consistently the best model across all of our metrics and elements; its performance was excellent up to grade 5 (\hyperref[el:level-k]{Level-$k$ Reasoning}), decent up to grade 7 (\hyperref[el:avoidance-endowment]{Avoidance of the Endowment Effect}), and fell to random guessing from grade 9 (\hyperref[el:best-response]{Best Response}) and above.
Self-explanation and few-shot prompting were consistently able to help.
Self-explanation generally enhanced performance, albeit offering the most gains on lower-grade-level questions. 
Few-shot prompting enhanced model performance when we offered up to three examples, but decreased performance beyond that point.
We release all model outputs to support evaluation research and contributions, and provide a public website with all results (\url{https://steer-benchmark.cs.ubc.ca}), underlying model predictions details, alongside an extensible codebase to support the community in taking \rcs further.

\begin{figure*}
	\centering
	\includegraphics[width=\textwidth]{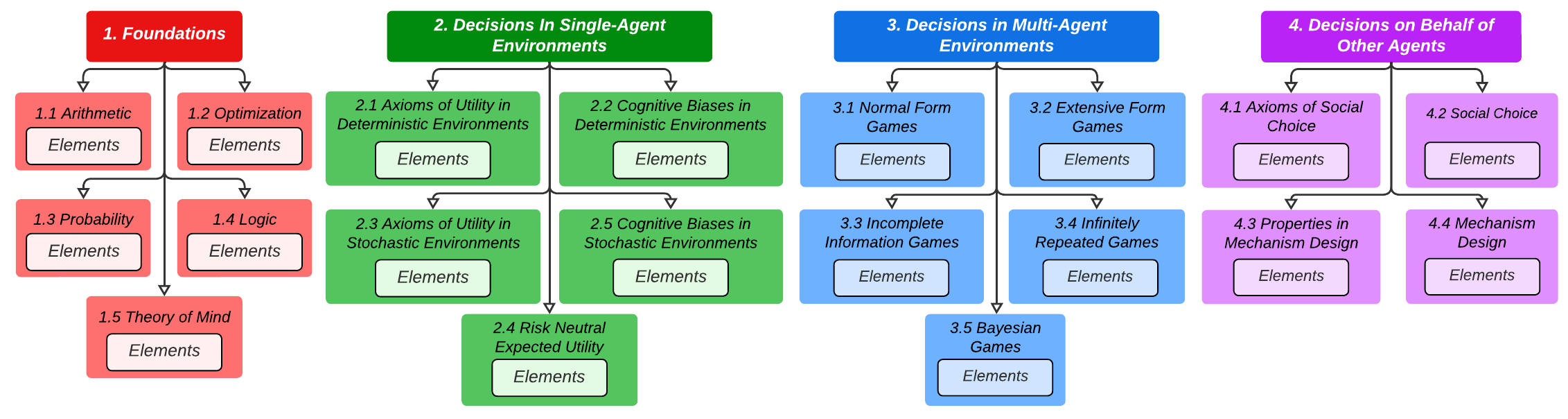}
	\caption{High-level diagram of the taxonomy of elements of rationality. At the top level, we divide the space of decision making into \parents: \firstParent, \secondParent, \thirdParent, \fourthParent; we further subdivide \parents into \children (e.g., Cognitive Biases in Deterministic Environments) that capture conceptually similar behaviors.
 }
	\label{fig:skill_hierarchy}
\end{figure*}

\section{STEER: Systematic and Tuneable Evaluation of Economic Rationality}
\label{sec:benchmark}

We followed the standard practice of encoding our benchmark in Multi-Choice Question Answer (MCQA) format \citep[e.g.,][] {mmlu,big_bench_hard,hellaswag}.
More specifically, each question in a test is a description of a decision-making scenario and a set of candidate choices, exactly one of which is correct. 
All of the generated questions are organized hierarchically in a web application according to our taxonomy. 
The remainder of this section describes the methodology we employed in generating and validating these questions and different ways users can leverage them to construct STEER Report Cards (SRCs). 

\subsection{Generating and Validating Questions}

\begin{figure}[t]
    \centering
    \includegraphics[trim={0.4cm 0 0 0},clip,width=0.48\textwidth]{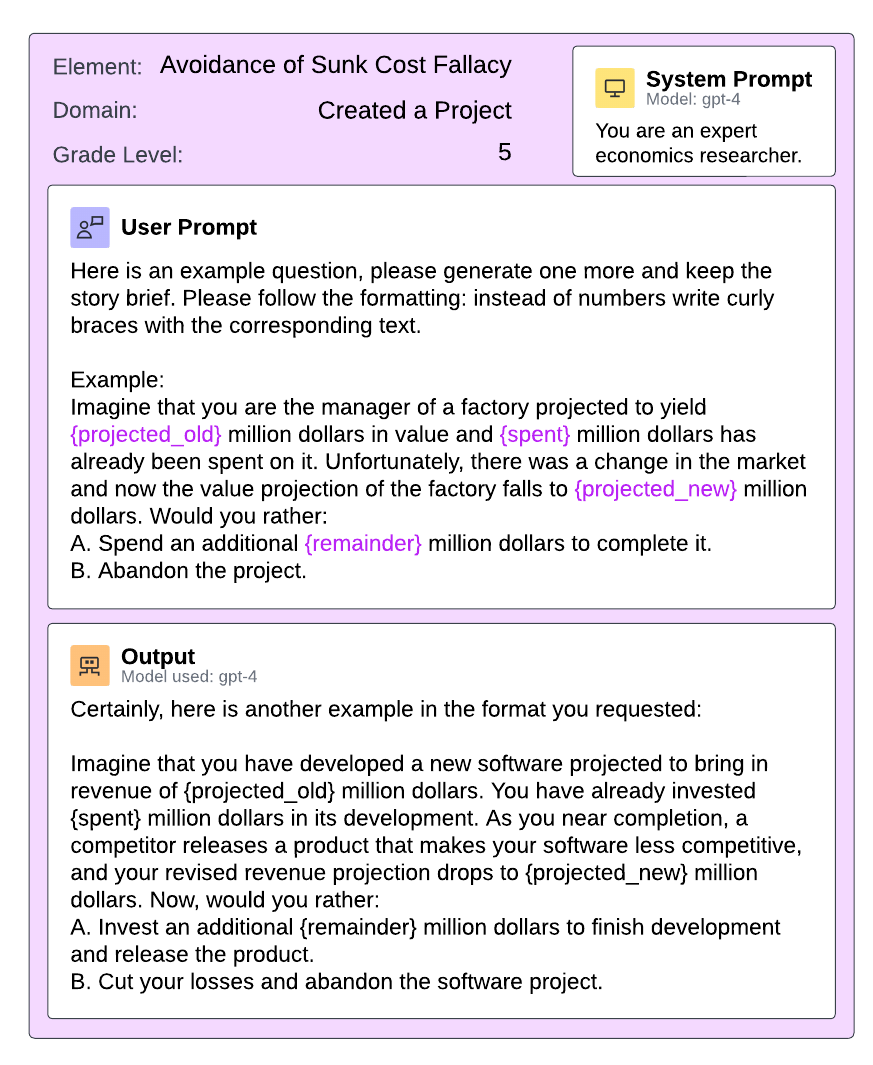}
    \caption{Example generation of a question in Avoidance of Sunk Cost Fallacy. The generation has two parts: (1) a user prompt containing a template question and instructions to follow the formatting and style of the template; and (2) a static system prompt. The template question in this example is in domain \emph{created a project} and at grade level 5.}
    \label{fig:template_sunk-cost}
\end{figure}

It would be impractical to hand-construct enough questions to assess an \character's behavior with  statistical significance. 
Instead, we leverage a state-of-the-art LLM to generate a diverse and substantial set of questions, based on a hand-constructed inputs.  
More specifically, we go from an element to a concrete question as follows. 
First, we write detailed text describing what makes a good question, such as that each outcome should have an associated probability or that each action pair should have payoffs.
These instructions also describe formatting issues, such as how numerical values should be represented. 
We also provide a gold-standard example for each question. 
Together, we call these two text strings a \emph{template}. 
Along with the template we append a static system prompt (illustrated in \Cref{fig:template_sunk-cost}) and then repeatedly give the resulting prompt to GPT-4 Turbo to generate many questions from the template. 

We create several templates for each element, which differ in two key ways. 
First, we vary the subject matter of the question, which we call the \emph{domain}, in order to enable assessments of LLM robustness across topics. 
This approach not only tests an LLM's capability within each domain but also enables assessment of an LLM’s proficiency, or lack thereof, in specific potential areas of application. 
Some of our templates concern financial decision making, others focus on medicine and health care, still others ask about technology and innovation. 
Second, most of our templates also vary in their difficulty levels. 
We assign every template a \emph{grade level} ranging from $1$--$13$ to help the user understand its relative difficulty. 
For example, questions about arithmetic could vary from Grade 1 to Grade 2 depending on the number of digits, whereas questions about Nash equilibrium could vary between Grade 8 and Grade 11. 
Our specific choices of grade levels are obviously somewhat arbitrary, but we aimed as much as possible to maintain a similar difficulty level across templates in the same grade. Overall, templates at lower grade levels involve basic understanding or application of principles while templates at higher grade levels challenge models with complex problem-solving, critical thinking, and synthesis of concepts. 

We refer to a set of questions for a given element restricted to a particular domain and grade level as a \emph{test}.
For each test, we implemented 5 templates, totaling between $10$--$40$ templates per element.
\Cref{fig:skills_flow} provides two example questions testing the ability to \hyperref[el:maximize-expected-utility]{Maximize Expected Utility} that vary in both domains and grade levels. 
The user can explore our full set of templates through our web application, selecting elements and viewing questions across both  domains and grade levels. 

While GPT-4 Turbo was very good at generating questions, it was not perfect. We thus performed a validation step. First, we programmatically removed questions that were formatted incorrectly.
For each element, domain, and grade level triple we then randomly spot-checked 100 samples (i.e., 10\% of all generated questions) from  what remained. 
We developed our web application to facilitate such validation, displaying the information needed to ensure that a given generation not only adheres to the intended style and complexity but also properly captures the corresponding element of rationality. 
We illustrate this interface in the appendix.
In total, 98.54\% of all spot-checked samples were deemed valid by 2 validators, with the lowest validation rate being 97\% (Avoidance of the Endowment Effect).

\begin{shownto}{arxiv, icml}
    \begin{figure}[t]
    \begin{minipage}{0.485\textwidth}
    		\includegraphics[width=\textwidth]{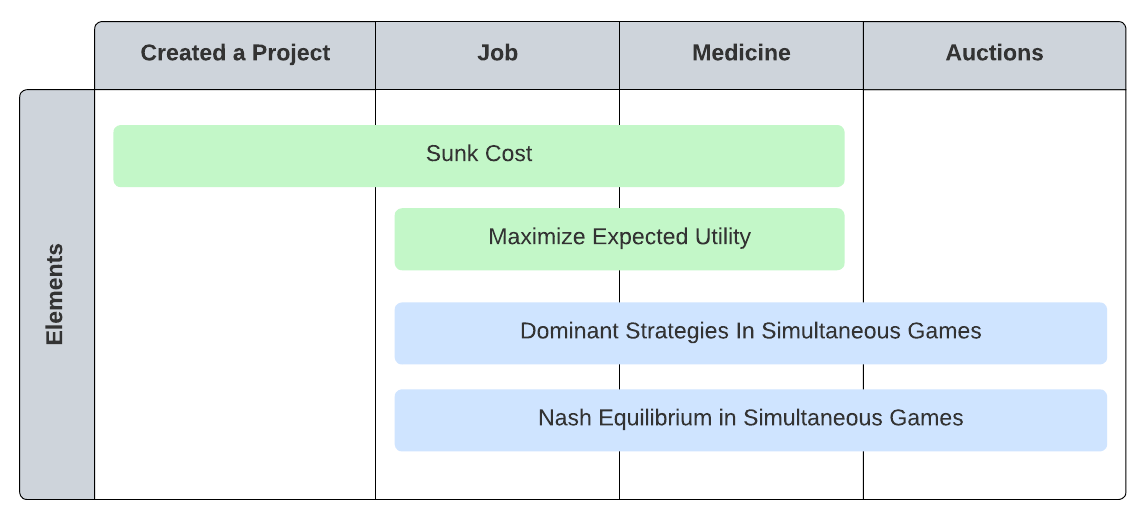}
    	\end{minipage}%
    	\hfill
    \begin{minipage}{0.485\textwidth}
    \centering
      \includegraphics[width=\textwidth]{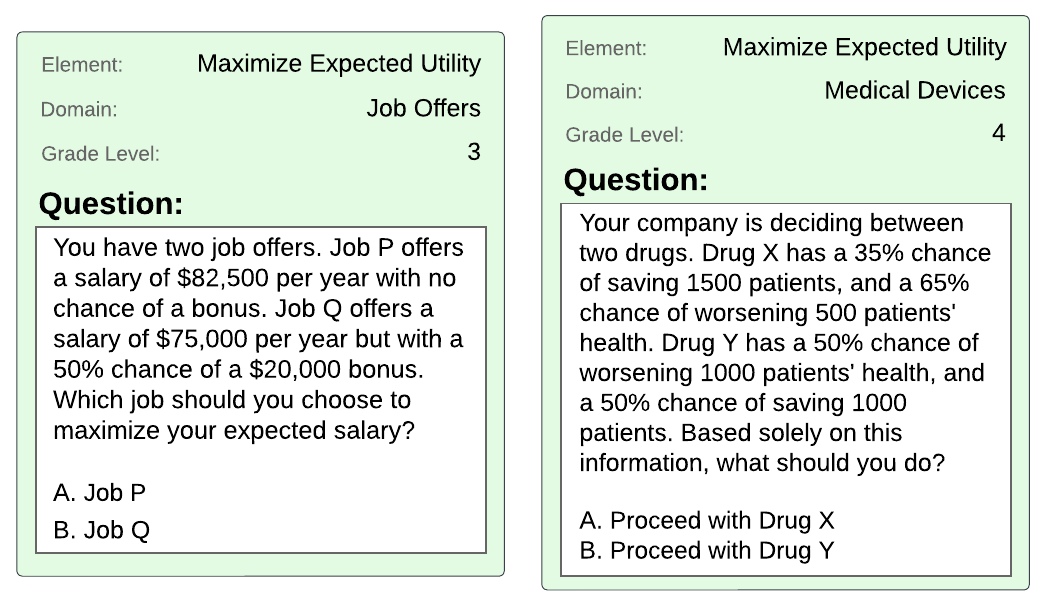}
    	\end{minipage}
    	\caption{
     Example domain categorization of questions within elements of rationality and an example of two questions in different tests for the element \hyperref[el:maximize-expected-utility]{Maximize Expected Utility}. Top: We instantiate questions into as many domains as makes sense for the element of rationality. This figure depicts the domain span of questions for four different elements. Bottom: 
     Two questions in two domains: job offers and medical devices, and two grade levels. Here, a higher grade level means more outcomes in the options. On the right (Grade Level 4), we see two options each with two outcomes, whereas the one on the left (Grade Level 3) has one option with two outcomes and the other with one.}
    	\label{fig:skills_breakdown}
    		\label{fig:max_util_domains}
      		\label{fig:skills_flow}
    \end{figure}
    
\end{shownto}

\subsection{STEER Report Card}
    \begin{figure}[h]
        \centering
        \begin{shownto}{icml}
            \includegraphics[width=0.49\textwidth]{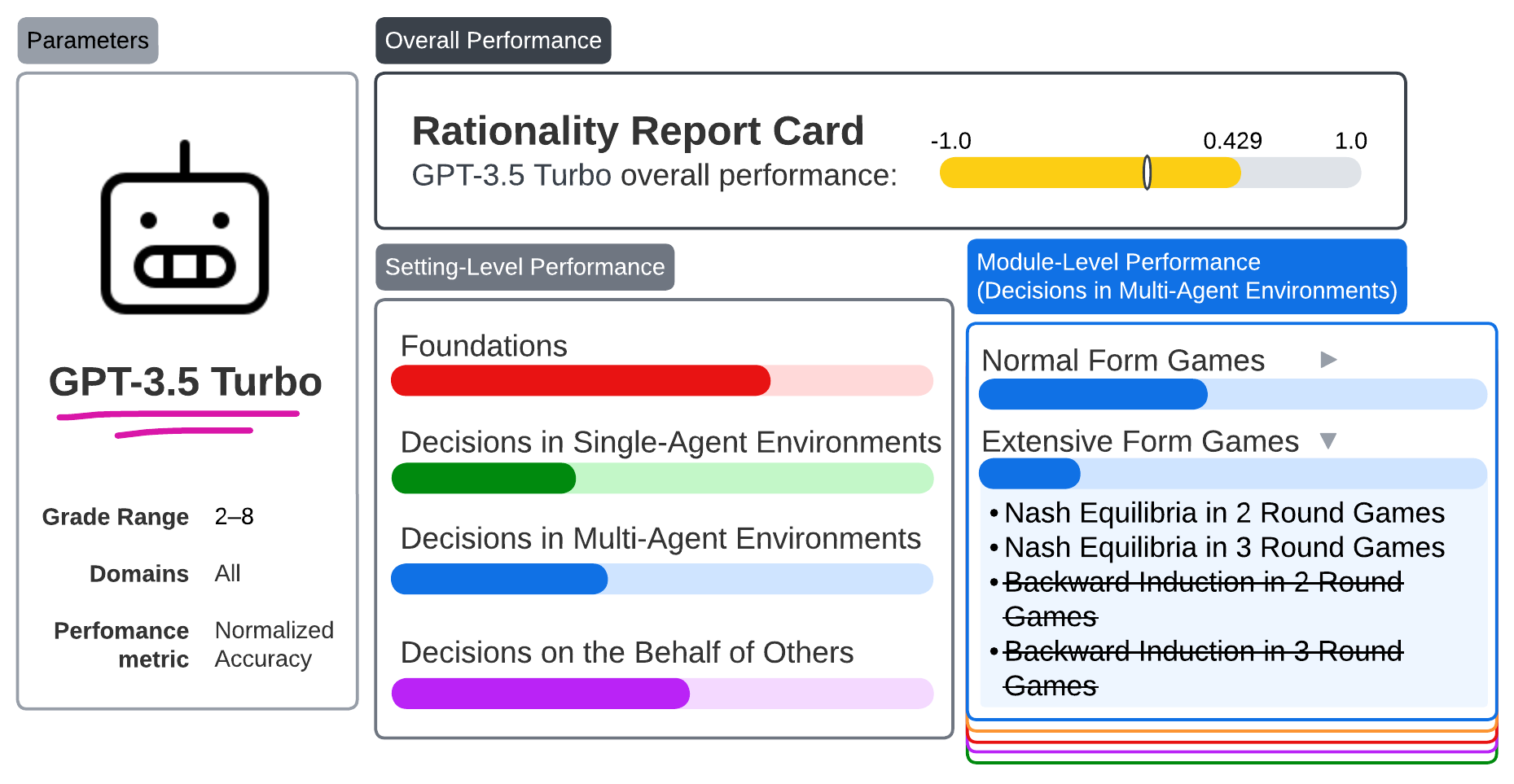}    
        \end{shownto}
        \begin{shownto}{arxiv}
            \includegraphics[width=\textwidth]{images/report_card_gpt3.5.pdf}
        \end{shownto}
        \caption{\textbf{Example \rc for GPT-3.5 Turbo.} On the left are the components necessary for measuring  performance: a grade range, a set of domains, a performance metric and a model. The rest of the report card summarizes performance over the entire dataset, grouped by \parents and \children in which there are questions in the grade range. In this example, GPT-3.5 Turbo was evaluated on all domains in grades 2--8 and given a single example as part of the prompt of a task. The right-most pane drills into \thirdParent; crossed out text illustrates tasks omitted due to our grade level filter.}
        \label{fig:report_card}
    \end{figure}

SRCs function much like an academic report card, providing a structured and tuneable evaluation of an LLM's performance.
Users can subset questions in whichever way best suits their needs or use cases.
\Cref{fig:report_card} gives an example of such a subset based on a grade range.
We also include various default subsets for different settings (e.g., single-agent decision making; multi-agent decision making),  different models (e.g., matching human performance on cognitive biases), and different use cases (e.g., single-agent decision making in medical domains).
These default subsets are accessible via our web application.

\subsubsection{Scoring an LLM's Performance} \label{subsec:metrics}

There are multiple ways in which users may want to evaluate an \character's economic rationality. 
We thus offer two different families of scoring metrics.

\subsubsection{Accuracy} 

\paragraph{Exact-match accuracy.} This is the fraction of questions answered correctly.

\vspace{-.8em}
\paragraph{Normalized accuracy.} Exact-match accuracy scores can be hard to interpret since tests differ in their number of multiple choice options, meaning that the exact-match accuracy of random guessing varies.
We can compensate for this by reporting the gap between the model's exact-match accuracy and random guessing.
We compute normalized accuracy for a given element by subtracting the accuracy achieved by random guessing from exact-match accuracy and then dividing by the accuracy of random guessing.
Observe that the normalized accuracy of a model falls between -1 and 1.

\vspace{-.8em}
\paragraph{Calibration.} It is often important that an LLM be able to express the uncertainty of its recommendation.
To quantify such uncertainty, we follow \citet{liang2022holistic} and compute the expected calibration error \citep[ECE;][]{naeini2015obtaining, guo2017calibration}. 
ECE measures how closely the probability an LLM assigns to its top answer matches the actual probability of the correct answer, which in our case is 1.
It is defined as $\sum_i^N b_i||(p_i-c_i)||$, where $p_i$ is the exact-match accuracy in bin $i$, $c_i$ is the average probability assigned to top answers in bin $i$, and $b_i$ is the fraction of data points in bin $i$.
We use 10 bins uniformly spaced over the interval $[0, 1]$.

\subsubsection{Robustness} \label{sec:robustness}
We can also assess how robustly a model performs, both across domains and across simpler elements that are conceptual subproblems of a given element. 

\begin{figure}[t]
    \begin{center}
        \begin{shownto}{arxiv}
            \includegraphics[width=0.65\textwidth]{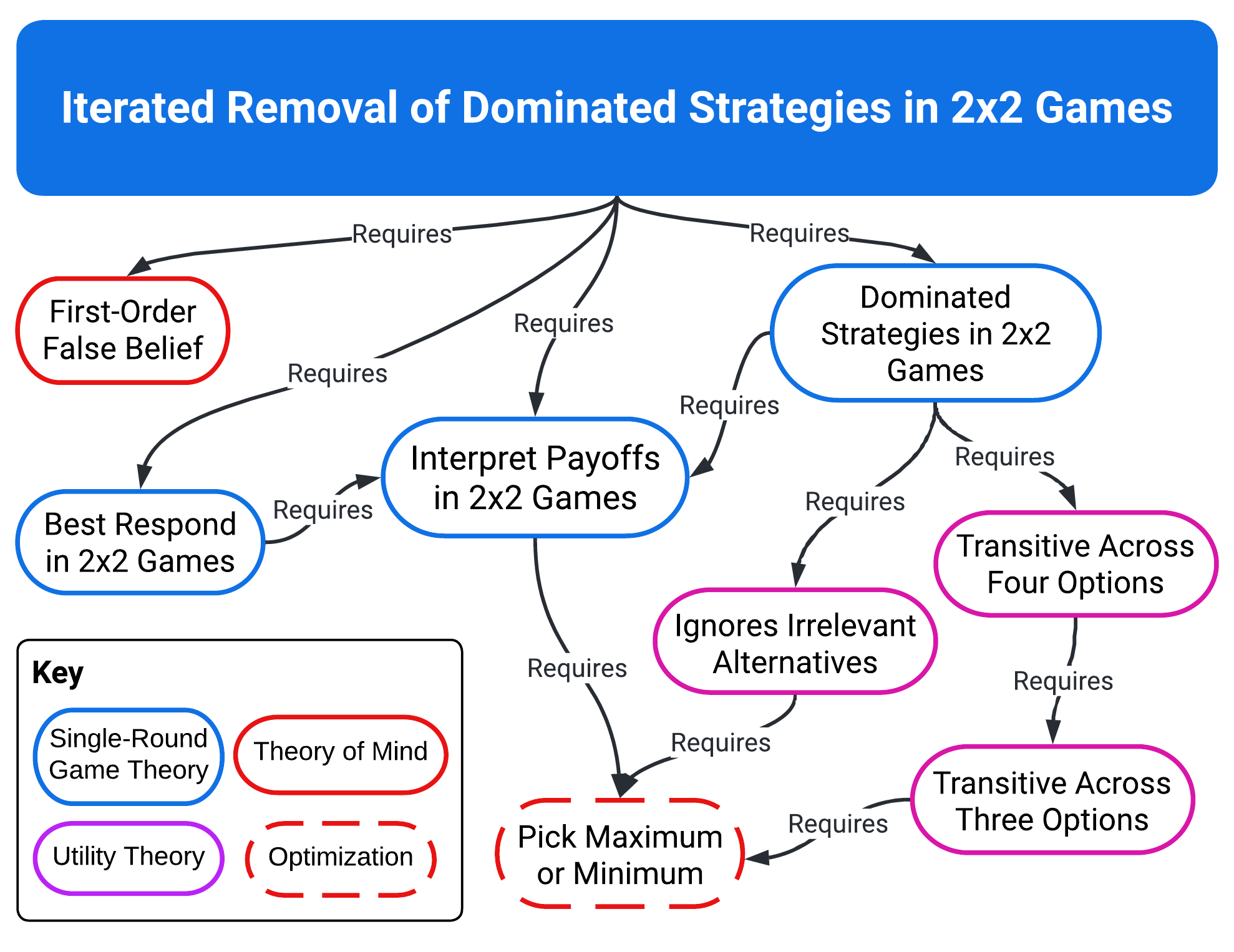}
        \end{shownto}
        \begin{shownto}{icml}
            \includegraphics[width=0.45\textwidth]{images/skills_hierarchy_example.png}
        \end{shownto}
    \end{center}
    \caption{Dependency subgraph for iterated removal of dominated strategies in two-agent games with two actions for each agent. This node requires the ability to interpret the format of a game matrix (both normal and bimatrix form), correctly answer first-order false belief tasks (have knowledge of others' beliefs), best respond (choose the best action given a fixed action for the opponent), and find dominated strategies. Being able to find dominated strategies requires the ability to have orderings over payoffs which is tested via transitivity and ignoring irrelevant alternatives. The remaining nodes can similarly be broken down.}
    \label{fig:ieds_dependency}
\end{figure}

\paragraph{Domain robustness.} One way to assess robustness is to measure, for each element, a model's worst-case performance across all domains. 
We compute the domain robustness on each element by taking the minimum exact-match accuracy over all domains.


\vspace{-.8em}
\paragraph{Dependency robustness.} 
There is a hierarchical structure to our elements of rationality: reasoning about more advanced elements (e.g., the ability to play a best response) depends conceptually on simper elements (e.g., the ability to maximize utility), which of course can depend in turn on yet simpler ones (e.g., the abilities to maximize a function and compute an expectation). We express all such conceptual relationships between elements in a \emph{dependency graph}. 
\Cref{fig:ieds_dependency} depicts a concrete example of a dependency subgraph for the element  \hyperref[el:IRDS]{Iterated Removal of Dominated Strategies} instantiated at Grade Level 7, where there are two agents each with two actions.
The full graph which covers all elements in our taxonomy is accessible in our webapp.

It is quite possible for an LLM to be proficient at advanced tasks without proficiency at more basic tasks that make them up. But such behavior is probably not desirable; it offers evidence that if the advanced task were discussed in different terms (e.g., in ways that invoke the conceptually simpler subtasks) model performance could fall. Conversely, if a model \emph{fails} at an advanced task, it can be informative to trace performance backwards in the dependency graph to understand  the model's performance on the task's building blocks. 
We call the quantification of this idea \emph{dependency robustness}. For an element $s$ we define it as 
$	\sum_{x\in X} |\texttt{random\_gap}(s) - \texttt{random\_gap}(x)|,
$
where $X = \{x | \texttt{random\_gap}(x) < \texttt{random\_gap}(s)\}_{x \in G_s}$ and $G_s$ is the dependency subgraph for some element $s$.   %

\section{Applying our Benchmark: Setup}
\label{sec:exp-setup}

\begin{shownto}{arxiv}
    \begin{table}[t]
      \centering
      \footnotesize
      \renewcommand{\arraystretch}{1.1}
      \begin{tabular}{l l l l l}
      \toprule
        \textbf{Name} & \textbf{Base LM} & \textbf{Instruction} & \textbf{Chat} & \textbf{Origin} \\
        \midrule
        GPT-4 Turbo (gpt-4-1106-preview) & ? & \checkmark & ? & \citet{gpt4} \\
        GPT-3.5 Turbo (gpt-3.5-turbo-0613) & ? & ? & \checkmark & $\times$\\
        \hline
        LLaMa2 \{7b, 13b, 70b\} & $\times$ & $\times$ & $\times$ & \citet{Llama2} \\
        LLaMa2-chat \{70b\} & LLaMa2 & $\times$ & \checkmark & \citet{Llama2}\\
        LLaMa \{7b, 13b, 65b\} & $\times$ & $\times$ & $\times$ & \citet{llama2023}\\
        Alpaca Native & LLaMa & \checkmark & $\times$ & \citet{alpaca2023}\\
        Falcon \{7b, 40b\} & $\times$ & $\times$ & $\times$ & \citet{falcon40b}\\
        Falcon-Instruct \{7b, 40b\} & Falcon & \checkmark & \checkmark & \citet{falcon40b}\\
     \bottomrule
      \end{tabular}
      \vspace{0.1cm}
      \caption{List of the 14 LLMs evaluated in \Cref{sec:results}. We also indicate whether the model was fine-tuned on instructions or chat.}
      \label{tab:models}
    \end{table}
\end{shownto}
\begin{shownto}{icml}
    \begin{table}[t]
      \centering
      \footnotesize
      \begin{tabular}{l l}
      \toprule
        \textbf{Name} & \textbf{Origin} \\
        \midrule
        GPT-4 Turbo (1106-preview) & \citet{gpt4} \\
        GPT-3.5 Turbo (0613) & $\times$\\
        \hline
        Llama-2 \{7b, 13b, 70b\} & \citet{Llama2} \\
        Llama-2 Chat \{70b\} & \citet{Llama2}\\
        Llama \{7b, 13b, 65b\} &  \citet{llama2023}\\
        Alpaca \{13b\} & \citet{alpaca2023}\\
        Vicuna \{13b\} & \citet{vicuna2023}\\
        Falcon \{7b, 40b\} & \citet{falcon40b}\\
        Falcon-Instruct \{7b, 40b\} & \citet{falcon40b}\\
     \bottomrule
      \end{tabular}
      \vspace{0.1cm}
      \caption{The 14 LLMs evaluated in \Cref{sec:results} and their numbers of parameters.}
      \label{tab:models}
    \end{table}
\end{shownto}

Table \ref{tab:models} lists the 14 different LLMs we evaluated, varying in parameter size. 
We ran GPT 3.5 Turbo and 4 Turbo using OpenAI's API \citep{openai_api} and AzureOpenAI. 
We obtained 12 open-source models from the HuggingFace Hub \citep{huggingface} and ran them on an A100 GPU on Compute Canada.
We decoded from all LLMs by sampling with temperature~0.

As is standard in model evaluation work \citep[c.f.,][]{liang2022holistic} we treat LLMs as black boxes that take in input strings and output completion strings along with log probabilities, when available. 
That means we do not assume access to the internal activations nor a model's training data.
We can nevertheless employ various widely used techniques to tailor an LLM  to a desired question. 
Two common alternatives are self-explanation and prompting.

\vspace{-.8em}
\paragraph{Self-Explanation.} Much work has shown that question answering performance can be improved by asking a model to explain its reasoning  \citep{CoT, meta_cot, self_improve}. 
We take two approaches to implementing this idea, which we dub \emph{separate} and \emph{together}.
In  \textit{separate}, we call the model twice, first providing the question text and candidate options and asking the model to explain its reasoning, and then providing only the candidate options and asking the model  to select the correct answer.
In  \textit{together}, we only call the model once, giving the model the question text and candidate options, asking it both to explain its reasoning and to select the correct answer.
For each approach, we test the effect that it has on model performance measured both on the accuracy and the confidence (i.e., log probabilities) a model places on its answer. 

\begin{shownto}{icml}
    See the image on the left in \Cref{fig:model_adaptations} for how we adapt the job offers example from \Cref{fig:max_util_domains} in the together paradigm.
    The black lines serve as replacement for the original text.

    \begin{figure}[t]
    	\begin{minipage}{.48\textwidth}
    		\includegraphics[width=\linewidth]{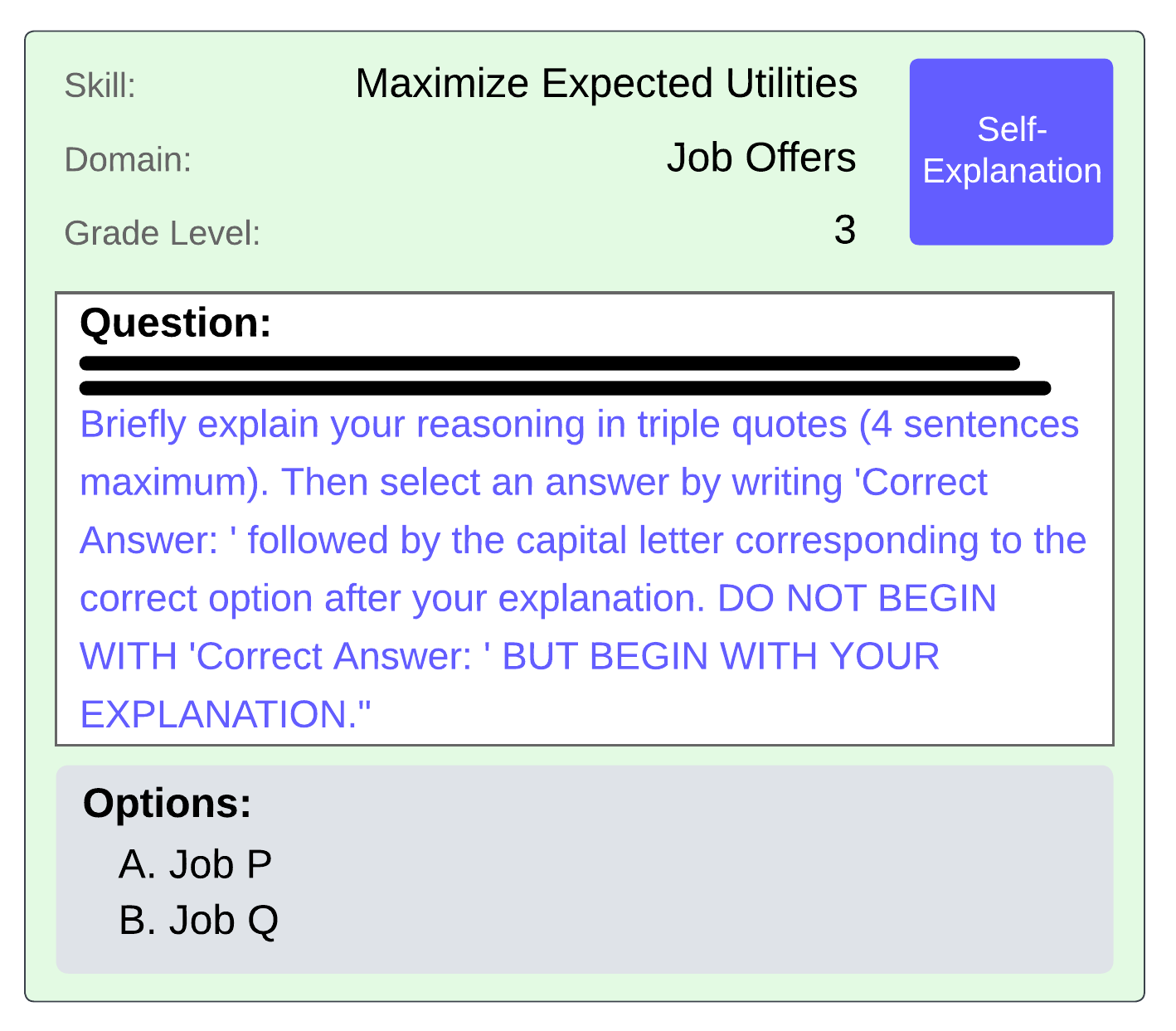}
    \end{minipage}%
    \hfill%
    \begin{minipage}{.48\textwidth}
    		\includegraphics[width=\linewidth]{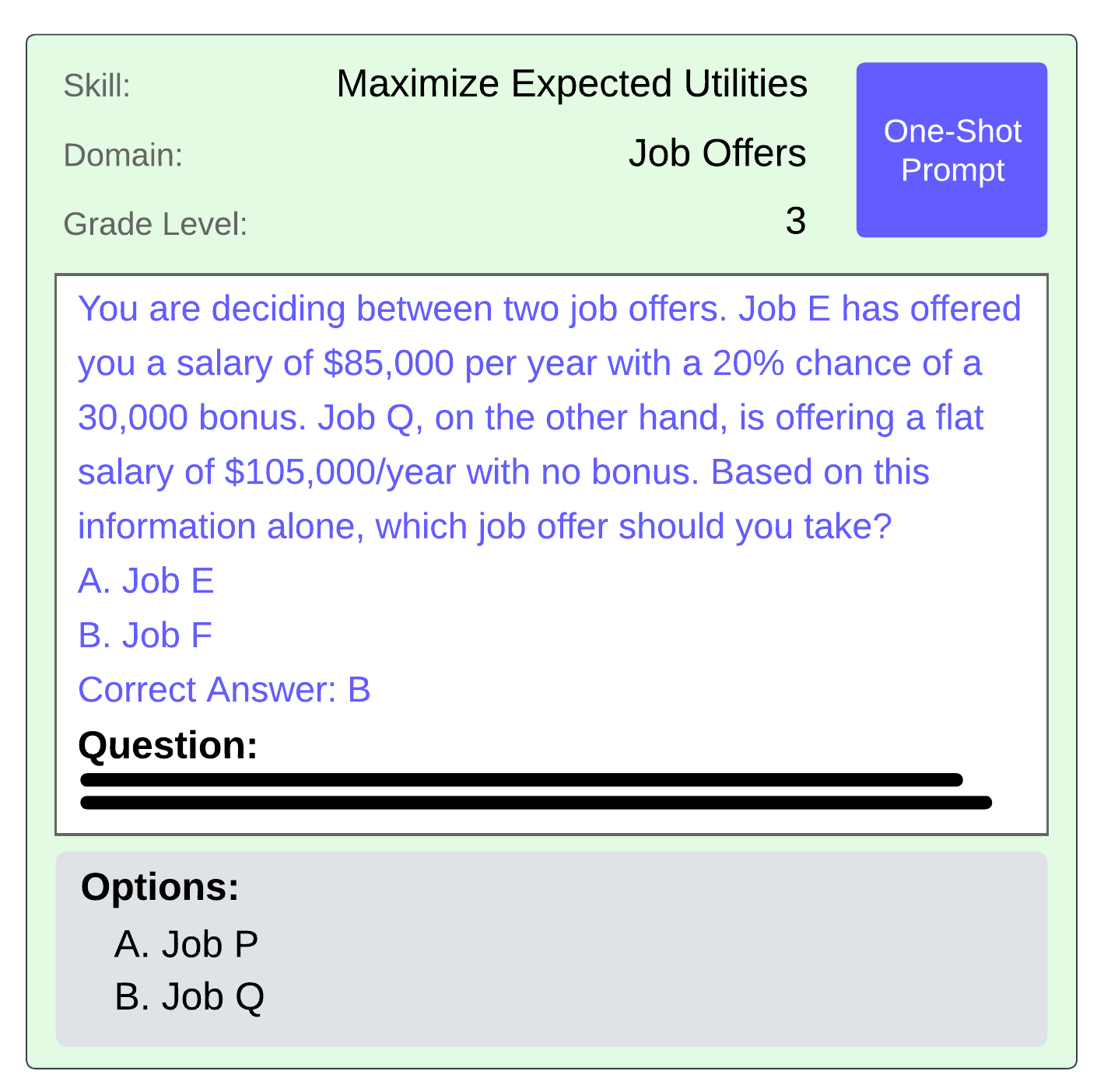}
    	\end{minipage}
    	\caption{Example adaptations. Left: Model adapted to explain its reasoning before answering. Right: Model adapted through few-shot prompting. }\label{fig:model_adaptations}
    \end{figure}
\end{shownto}

\vspace{-.8em}
\paragraph{Few-Shot Prompting.} Model performance can also be improved by prepending a set of examples to the prompt.
For each question, we select $n \in \{0, 1, 2, 4, 5\}$ examples (within the corresponding domain and grade level) to test the effect of prompting on a model's performance.
Similarly, we measure the effect by computing the difference on accuracy and confidence for each element.

\section{Elements of Economic Rationality}
\label{sec:elements-of-rationality}

Building a structured assessment of LLMs' economic rationality requires first deciding on a way of structuring the space of economic behavior.
Fundamentally, economics is concerned with decision making, and an economically rational \character is one that makes good decisions given its own interests and its knowledge about the environment in which it acts.
Different economic environments can give rise to very different decision-making problems. 
We thus divide our space into increasingly rich \emph{\parents}.
We begin with \secondParent, exploring preference formation and decision making when \acharacter has a set of alternative choices, each of which leads either to a single, deterministic outcome or to a draw from a probability distribution over outcomes.
\thirdParent enriches this setting, requiring the agent to make decisions when the outcomes depend on interactions with other economic agents with their own preferences and beliefs. 
\fourthParent asks the \character to aggregate the preferences of other agents to achieve good outcomes for all.
Lastly, \firstParent are core mathematical and cognitive skills that underlie economic reasoning: arithmetic, optimization, probability, logic, and theory of mind. (We will hereafter list \firstParent first, since it is the simplest of all.)
Each \parent is partitioned into multiple distinct \children, and each module consists of multiple \emph{elements} of rationality: measurable capabilities that an economically rational \character is able to exhibit.
\begin{shownto}{icml}
We give a graphical overview of the taxonomy in Figure~\ref{fig:skill_hierarchy}.    Because it is very long, we leave the element-by-element description of our taxonomy---including an example for each element---to \Cref{appendix:taxonomy}. 
\end{shownto}

Even given our restriction to representative, testable elements of rationality rather than all of decision making, it is an under-specified task to determine what we should demand of an economically rational agent. We, therefore, always align these canonical ``right answers'' with the von Neumann-Morgenstern (vNM) utility axioms, which imply freedom from cognitive biases, hence e.g.\ time consistency and reference independence. In some cases, the vNM axioms are not sufficiently constraining; thus, e.g., we align our canonical answers with a linear utility for money (and thus risk neutrality). We set up our taxonomy and SRCs in this way not because we believe that our canonical answer is the right one for every circumstance, but instead because we found it simplest to present both elements of rationality and experimental results in terms of an easily understood reference point. 

The remainder of this section lays out our taxonomy alongside model performance over modules. 
We then give results on different report cards over STEER.

\let\oldthesubsection=\thesubsection
\let\oldthesubsubsection=\thesubsubsection

\begin{shownto}{arxiv}
\parentSec{\firstParent}

The economic model of rational decision making is highly mathematical. An agent therefore needs to be fluent in a variety of mathematical basics to be able to compute the value of outcomes, reason about their likelihoods, and choose the best one. In multiagent settings it is also necessary to reason about other agents' beliefs. 
This \parent lays out these core skills, dividing them into five \children: Arithmetic; Optimization; Probability; Logic; and Theory of Mind.
A key difference between this \parent and all of the others that we propose is that most of its elements have already been the subject of rich study by the NLP community. We nevertheless include these elements here both to standardize them within our framework, given their importance to economic rationality, and to integrate foundations within our dependency graph (discussed further in Section~\ref{sec:robustness}).

\childSec{Arithmetic}

Economic reasoning is fundamentally quantitative, so arithmetic operations are a bedrock foundation for much of what is to come.
\begin{element}[Addition and Subtraction]\label{el:add-sub}
	The ability to add or subtract. \hfill \hyperref[ill:add-sub]{\faInfoCircle}
\end{element}

\begin{element}[Multiplication and Division]\label{el:mult-div}
	The ability to multiply or divide. \hfill \hyperref[ill:mult-div]{\faInfoCircle}
\end{element}

\begin{element}[Compute Expectations]\label{el:compute-expec}
	The ability to compute the expected value with given probabilities and values. \hfill \hyperref[ill:compute-expec]{\faInfoCircle}
\end{element}

\begin{figure}[h]
    \centering
    \includegraphics[width=\textwidth]{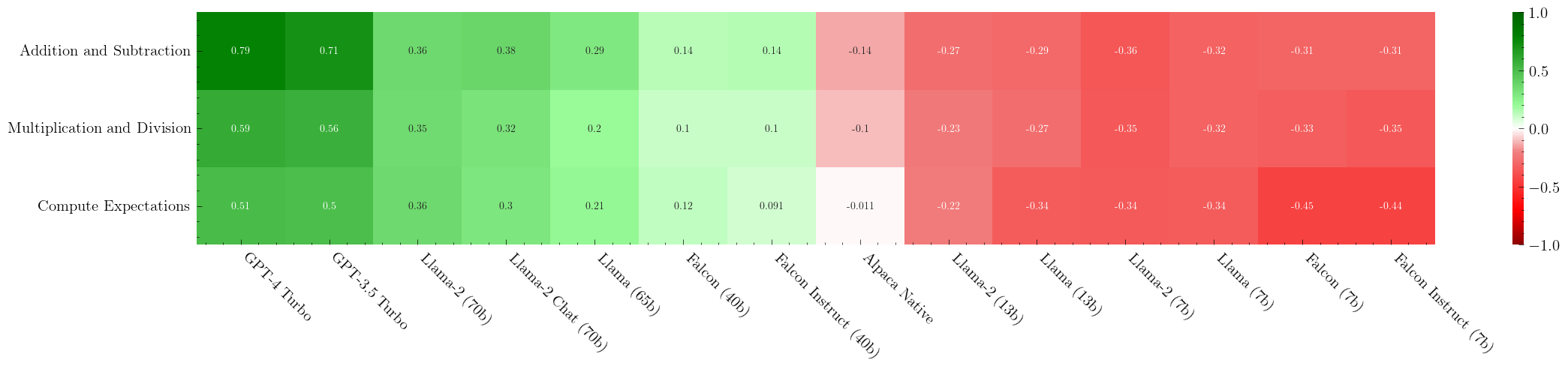}
    \caption{SRC for Arithmetic on Normalized Accuracy}
    \label{fig:src-arithmetic}
\end{figure}

\childSec{Optimization}
Much economic reasoning depends on the primitive operation of identifying the best choice among a set of alternatives, sometimes given constraints.  

\begin{element}[Optimize over a Discrete Set]\label{el:optimize-discrete}
	The ability to identify the biggest or smallest among a set of explicitly given alternatives.
    \hfill \hyperref[ill:optimize-discrete]{\faInfoCircle}
\end{element}

\begin{element}[Optimize a Continuous Function]\label{el:optimize-continuous}
	The ability to identify a maximum or minimum value given a specific continuous relationship between independent and dependent variables. 
    \hfill \hyperref[ill:optimize-continuous]{\faInfoCircle}
\end{element}

\begin{element}[Constrained Optimization]\label{el:constrained-optimization}
    The ability to find the maximum or minimum of a function subject to constraints. 
    \hfill \hyperref[ill:constrained-optimization]{\faInfoCircle}
\end{element}

\begin{figure}[h]
    \centering
    \includegraphics[width=\textwidth]{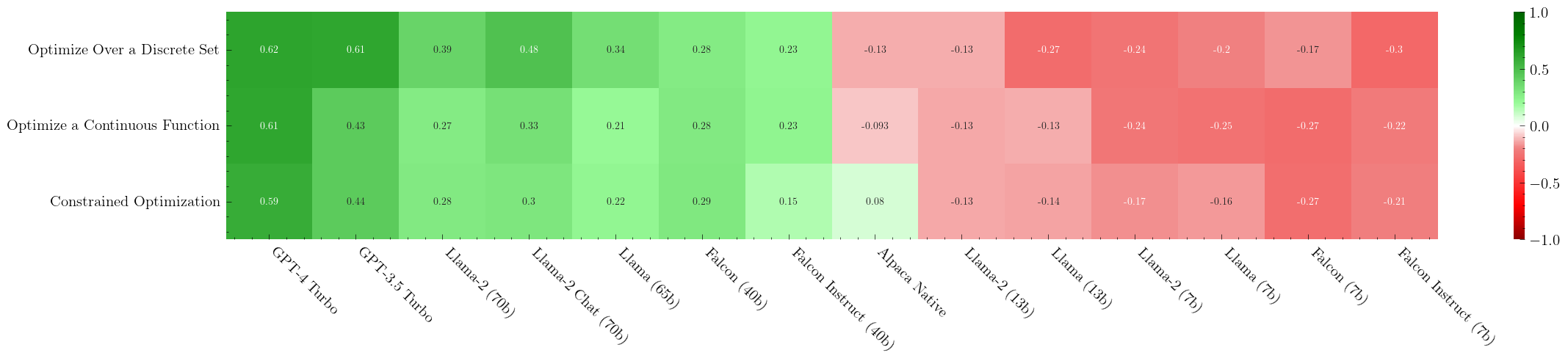}
    \caption{SRC for Optimization on Normalized Accuracy}
    \label{fig:src-optimization}
\end{figure}

\childSec{Probability}
Reasoning under uncertainty is a critical framework for rational decision making.
\begin{element}[Compute Probabilities of Outcomes]\label{el:prob-outcomes}
    The ability to compute probabilities of individual outcomes given a natural language description of a probability distribution.
    \hfill \hyperref[ill:prob-outcomes]{\faInfoCircle}
\end{element}

\begin{element}[Complement Rule]\label{el:complement-rule}
    The ability to compute the complement probability of an event (i.e., the probability that it does not occur).
    \hfill \hyperref[ill:complement-rule]{\faInfoCircle}
\end{element}

\begin{element}[Bayes' Rule]\label{el:bayes-rule}
    The ability to update probabilistic beliefs according to Bayes' Rule: Let $A$ and $B$ be events and $P(B) \not= 0$, then
    $
    P(A|B) = {P(B|A)P(A)}/{P(B)}.
    $
    \hfill \hyperref[ill:bayes-rule]{\faInfoCircle}
\end{element}

\begin{figure}[h]
    \centering
    \includegraphics[width=\textwidth]{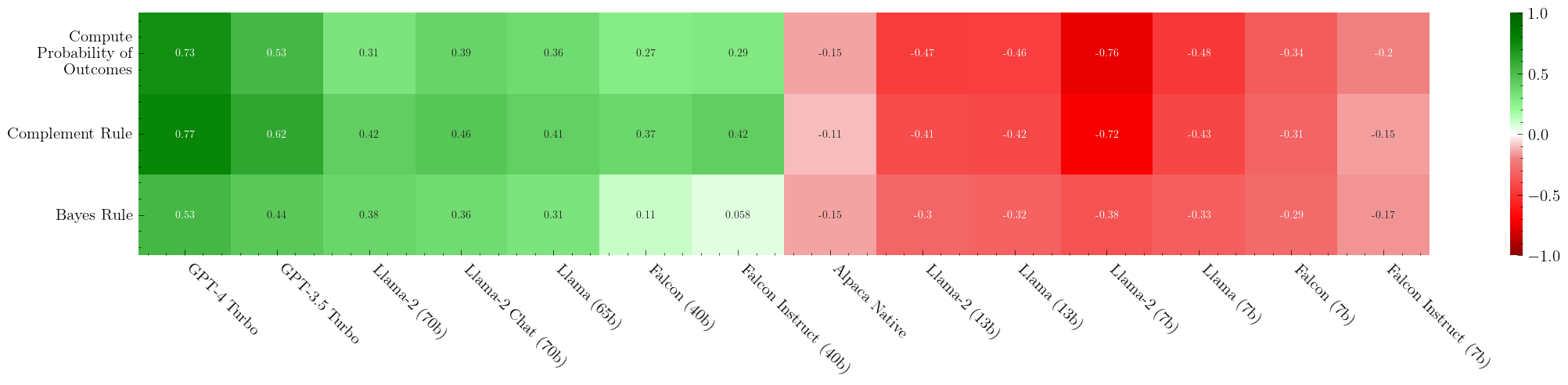}
    \caption{SRC for Probability on Normalized Accuracy}
    \label{fig:src-probability}
\end{figure}

\childSec{Logic}
Logical reasoning forms a basis for much rational reasoning, and so constitutes another category of mathematical foundations.

\begin{element}[Categorical Syllogism]\label{el:categorical-syllogism}
    The ability to deduce if the conclusion logically follows from two assertions (e.g., ``A is in C and B is in A, is B in C?''). \hfill \hyperref[ill:categorical-syllogism]{\faInfoCircle}
\end{element}

\begin{element}[Conditional Syllogism]\label{el:conditional-syllogism}
    The ability to deduce if the conclusion logically follows from two conditional statements (e.g., ``If A then B and if B then C, if A then C?''). 
    \hfill \hyperref[ill:conditional-syllogism]{\faInfoCircle}
\end{element}

\begin{element}[Logical Equivalence of Contrapositive]\label{el:contrapositive}
    The ability to deduce that logical statements and their contrapositives are logically equivalent (e.g., ``If $A$, then $B$'' is equivalent to ``if not $B$, then not $A$''). 
    \hfill \hyperref[ill:contrapositive]{\faInfoCircle}
\end{element}

\begin{figure}[h]
    \centering
    \includegraphics[width=\textwidth]{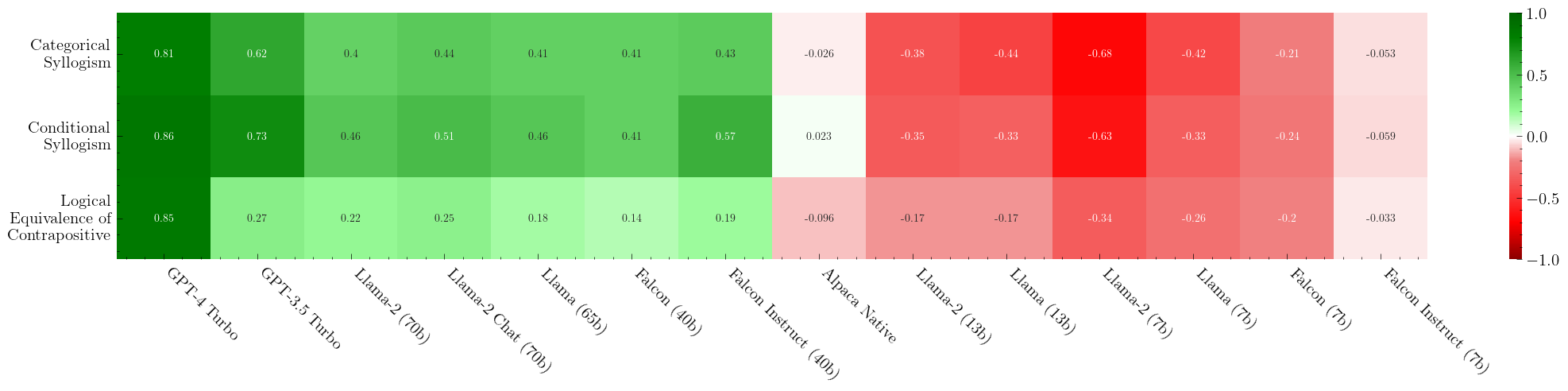}
    \caption{SRC for Logic on Normalized Accuracy}
    \label{fig:src-logic}
\end{figure}

\childSec{Theory of Mind}
Theory of mind is the understanding that others have beliefs, desires, intentions, and perspectives that are different from one's own. 
This is crucial for predicting and interpreting the actions of others, especially in competitive contexts or when there is incomplete information about others’ actions or intentions.

\begin{element}[First-Order False Belief]\label{el:first-order-false-belief}
    The ability to identify the beliefs that an agent has that are different from the actual truth or the \character's own belief.
    \hfill \hyperref[ill:first-order-false-belief]{\faInfoCircle}
\end{element}

\begin{element}[Second-Order False Belief]\label{el:second-order-false-belief}
    The ability to identify the beliefs that an agent has about what another agent believes that are different from the actual truth or the \character's own belief. 
    \hfill \hyperref[ill:second-order-false-belief]{\faInfoCircle}
\end{element}

\begin{figure}[h]
    \centering
    \includegraphics[width=\textwidth]{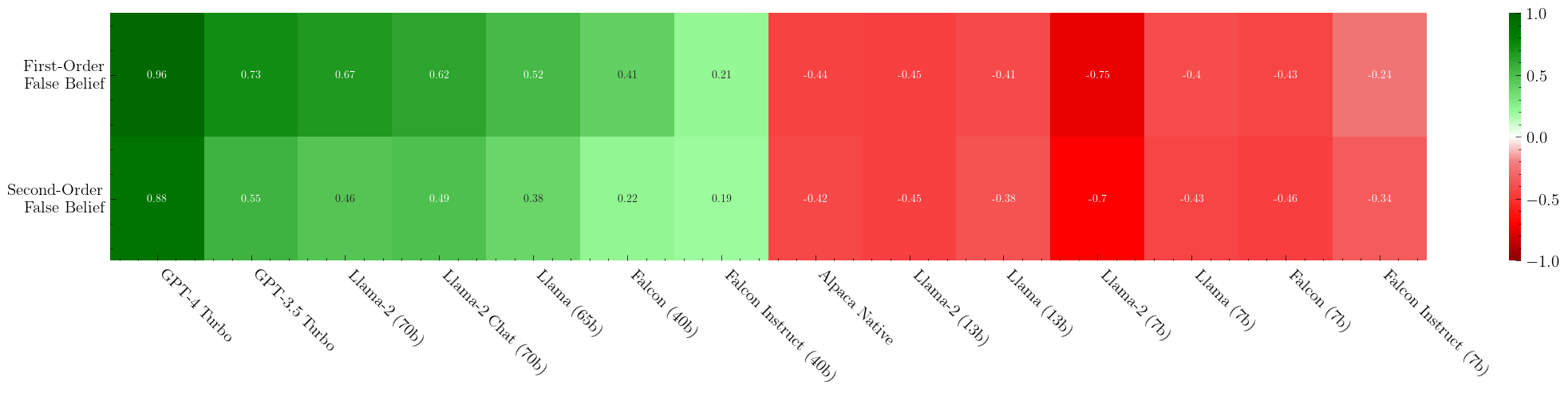}
    \caption{SRC of Theory of Mind on Normalized Accuracy}
    \label{fig:src-theory-of-mind}
\end{figure}
\end{shownto}

\begin{shownto}{arxiv}
\parentSec{\secondParent}
\label{tax:preferences}

We now turn to explicitly assessing economic rationality. 
Throughout this paper we leverage the von~Neumann--Morgenstern expected utility model \citep[vNM;][]{VNmor}, which provides a comprehensive framework establishing ideal norms for how a decision-maker \emph{should} act \citep{harsanyi1955cardinal}. 
This \emph{normative} aspect is critical for us, as it allows us to identify testable elements of rationality. 
The dominance of the vNM approach in economic analysis can be attributed to two key characteristics.
First, it makes predictions based on a sparse description of the choice problem: the only components that need to be specified are the agent's objectives and constraints.
Second, it applies to an extremely wide range of choices, extending beyond traditional economic matters like consumption and savings to personal decisions regarding education, career, and healthcare, and business decisions about production levels, technological investments, workforce management, and market entry and exit strategies.

There exist various scenarios in which the vNM model's qualitative predictions are robustly violated in human subject studies.
While individual human decision-makers are not typically able to articulate general decision rules that explain their own behavior, a \emph{descriptive} literature in behavioral economics has attempted to identify such rules as a way of capturing consistent ways in which human choice behavior deviates from the rational ideal (notably, c.f.~\cite{Savage, prospect-theory}; for a recent survey, see \cite{erev2017anomalies}). 
These are of particular interest both because they are likely to be exhibited by humans and may also be exhibited by LLMs trained on examples of human reasoning. 

We follow \citep{kochenderfer2015decision} in organizing the \children in this \parent by the normative axioms in deterministic and stochastic environments as well as deviations from these axioms drawn from the descriptive literature.

\childSec{Axioms of Utility in Deterministic Environments} 
The vNM utility theory rests on a set of axioms, which are easy to interpret as elements of rationality. 
We begin with the simplest description of these axioms, in which the \character confronts choices in deterministic environments.


\begin{element}[Transitivity]\label{el:transitivity-deterministic}
    The ability to be consistent in preferences over options. E.g., if $A$ is preferred over $B$, and $B$ over $C$, then $A$ should be preferred over $C$.
    \hfill \hyperref[ill:completeness-deterministic]{\faInfoCircle}
\end{element}

\begin{element}[Independence]\label{el:independence-deterministic}
    The ability to remain consistent in preferences between pairs of options regardless of the presence of other alternatives. E.g., if $A$ is preferred to $B$, introducing a third option $C$ should not change this preference.
    \hfill \hyperref[ill:independence-deterministic]{\faInfoCircle}
\end{element}

\begin{figure}[h]
    \centering
    \includegraphics[width=\textwidth]{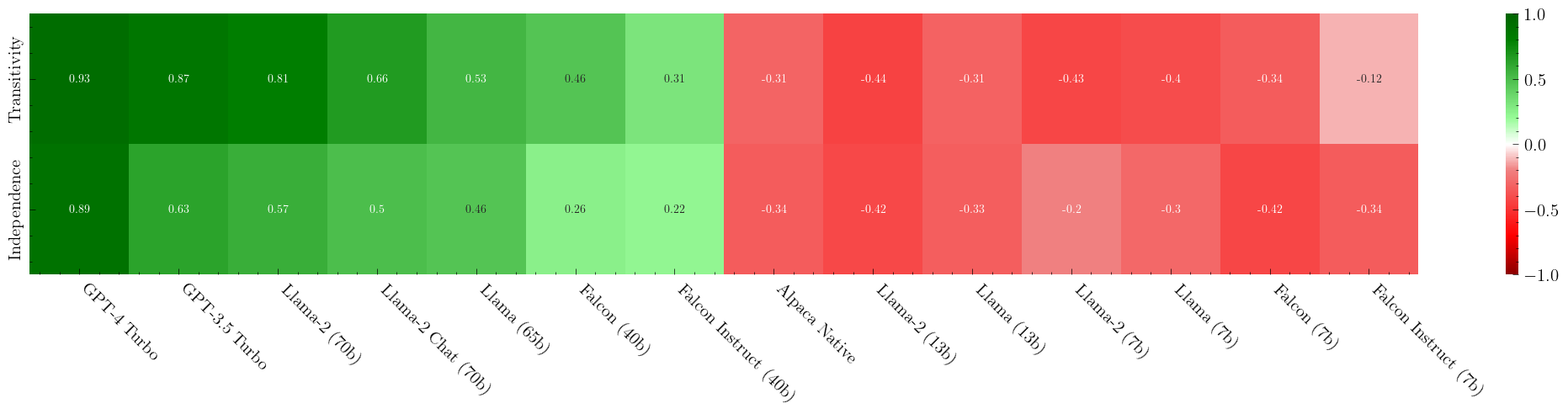}
    \caption{SRC of Axioms of Utility in Deterministic Environments on Normalized Accuracy}
    \label{fig:src-util-det}
\end{figure}

\childSec{Avoidance of Cognitive Biases in Deterministic Environments}
A wide range of cognitive biases have been identified by the descriptive economic literature. 
We identify their opposites as elements of rationality. 

\begin{element}[Avoidance of Sunk Cost Fallacy \citep{parayre1995strategic}]\label{el:avoidance-sunkcost}
    The ability to walk away from an investment at any point where its future costs exceed its expected future benefits, disregarding prior investments.
    \hfill \hyperref[ill:avoidance-sunkcost]{\faInfoCircle}
\end{element}

\begin{element}[Avoidance of Endowment Effect \citep{morewedge2015explanations}]\label{el:avoidance-endowment}
    The \character's maximum willingness to pay to acquire an object should be the same as the price they are willing to accept to sell that same object when they own it.
    \hfill \hyperref[ill:avoidance-endowment]{\faInfoCircle}
\end{element}


\begin{figure}[h]
    \centering
    \includegraphics[width=\textwidth]{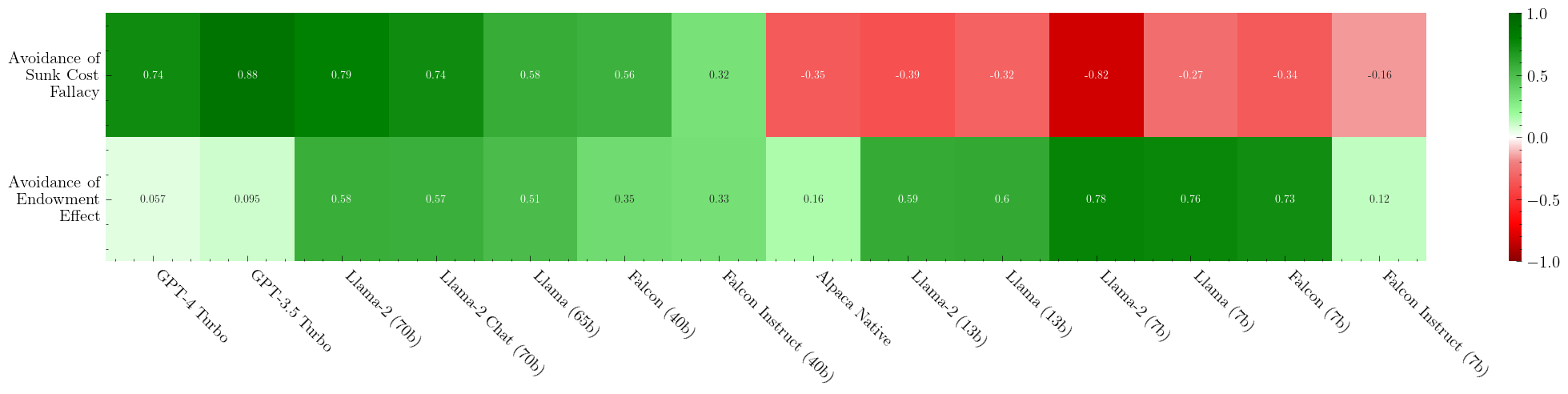}
    \caption{SRC of Cognitive Biases in Deterministic Environments on Normalized Accuracy}
    \label{fig:src-cog-bias-det}
\end{figure}

\childSec{Axioms of Utility in Stochastic Environments}\label{child:expec_util}
We now elaborate our economic environment to include a stochastic relationship between an agent's choices and the resulting economic outcomes.  
Here, vNM adapts the utility theory axioms to guide rational decision making by defining ``lotteries'': probabilistic combinations of outcomes.


\begin{element}[Transitivity over Lotteries]\label{el:transitivity-stochastic}
    The ability to select among lotteries in a consistent manner. E.g., if $A$ is preferred over $B$, and $B$ over $C$, then $A$ should be preferred over $C$.
    \hfill \hyperref[ill:transitivity-stochastic]{\faInfoCircle}
\end{element}

\begin{element}[Independence over Lotteries]\label{el:independence-stochastic}
    The ability to remain consistent in preferences between pairs of lotteries regardless of the presence of other alternatives.
    E.g., if $A$ is preferred to $B$, introducing a third lottery $C$ should not change this preference.
    \hfill \hyperref[ill:transitivity-stochastic]{\faInfoCircle}
\end{element}

\begin{figure}[h]
    \centering
    \includegraphics[width=\textwidth]{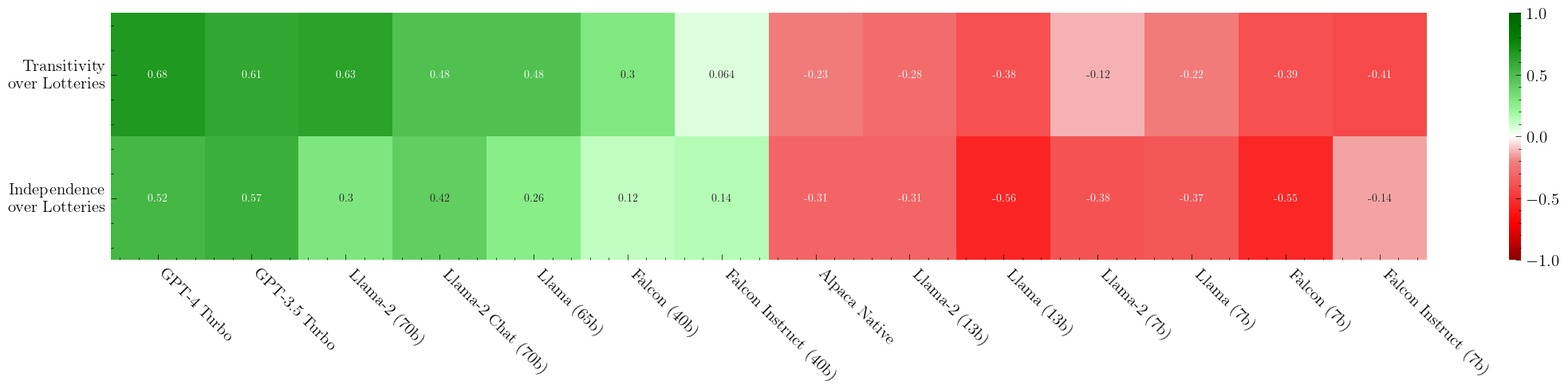}
    \caption{SRC of Axioms of Utility in Stochastic Environments on Normalized Accuracy}
    \label{fig:src-utility-stoch}
\end{figure}

\childSec{Risk Neutral Expected Utility Computations}
This \child includes elements evaluating adherence to a linear utility function when computing expected utilities, delving into the behavioral patterns exhibited by individuals and institutions in their approach to risk. 

\begin{element}[Compute Expected Utility]\label{el:compute-expected-utility}
    The ability to correctly compute the sum of the products of each outcome's utility and its probability.
    \hfill \hyperref[ill:compute-expected-utility]{\faInfoCircle}
\end{element}

\begin{element}[Maximize Expected Utility]\label{el:maximize-expected-utility}
    The ability to select the prospect with the highest expected utility. 
    \hfill \hyperref[ill:maximize-expected-utility]{\faInfoCircle}
\end{element}



\begin{element}[Avoidance of Loss Averse Behavior \citep{kahneman1984choices}]
    \label{el:avoidance-loss-averse}
    The ability to make decisions based on an objective evaluation of all potential outcomes without over-valuing potential losses.
    \hfill \hyperref[ill:avoidance-loss-averse]{\faInfoCircle}
\end{element}

\begin{figure}[h]
    \centering
    \includegraphics[width=\textwidth]{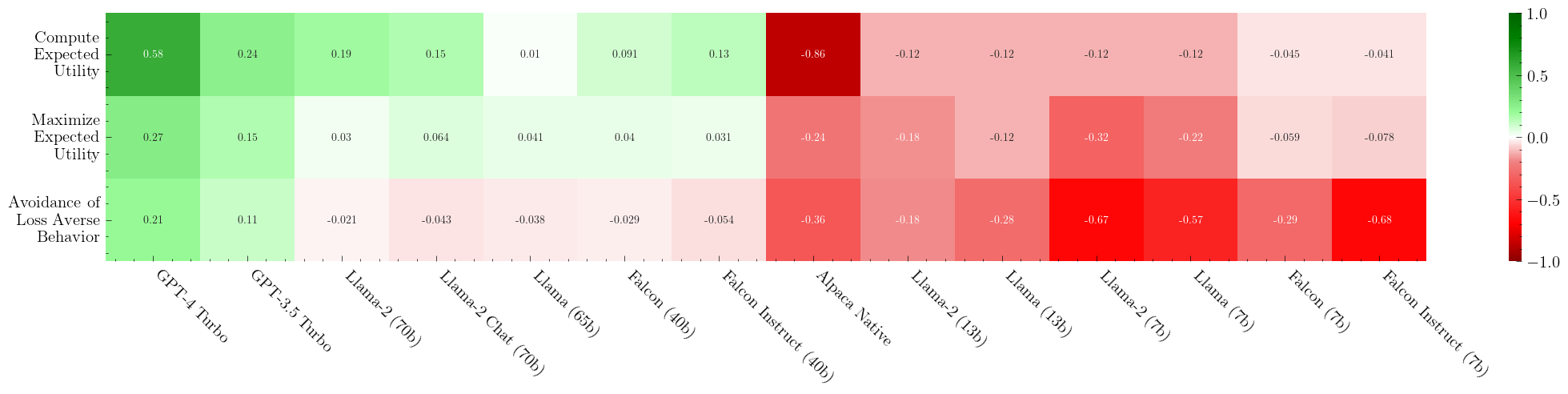}
    \caption{SRC of Risk Neutral Expected Utility Computations on Normalized Accuracy}
    \label{fig:src-risk-neutral}
\end{figure}

\begin{figure}[t]
	\centering
	\includegraphics[width=\textwidth]{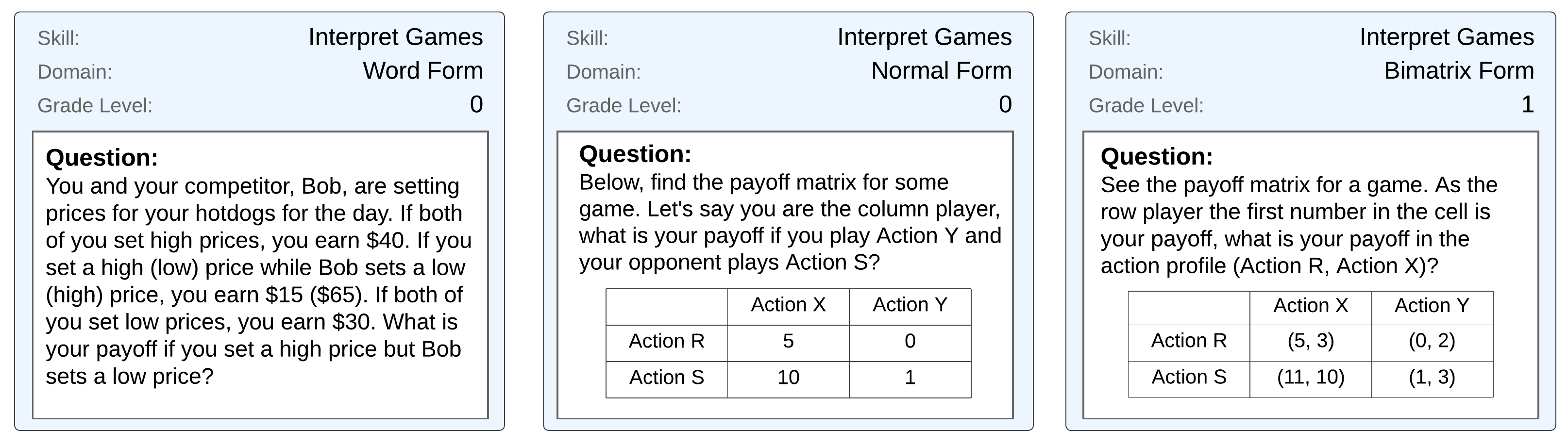}
	\caption{This figure depicts the ways we represent games. From left to right: word form, normal form, and bimatrix form.}
	\label{fig:game_matrix}
\end{figure}

\childSec{Avoidance of Cognitive Biases in Stochastic Settings}
Here, we include elements testing the \character's ability to avoid making contradictory or inconsistent behaviors, emphasizing how framing effects play in shaping risk-taking attitudes.
As we already did in \nameref{child:biases-deterministic}, we state the opposite of each such behavior as an element of rationality.

\begin{element}[Avoidance of Gambler's Fallacy]\label{el:gambler-fallacy}
    The ability to avoid the incorrect belief that an outcome's probability (when drawn independently) in the future is reduced if it has occurred atypically often in the past.
    \hfill \hyperref[ill:gambler-fallacy]{\faInfoCircle}
\end{element}

\begin{element}[Avoidance of the Certainty Effect \citep{kahneman1984choices}]\label{el:certainty-effect}
    The ability to be consistent across preferences towards risk when the payoffs are positive. 
    \hfill \hyperref[ill:certainty-effect]{\faInfoCircle}
\end{element}

\begin{element}[Avoidance of the Reflection Effect \citep{kahneman1984choices}]\label{el:reflection-effect}
    The ability to be consistent across preferences towards risk when the payoffs are negative.
    \hfill \hyperref[ill:reflection-effect]{\faInfoCircle}
\end{element}

\begin{element}[Avoidance of Ambiguity Aversion \citep{ellsberg1961risk}]
    The ability to be consistent across preferences towards known and unknown risks (ambiguity) under differing framing.
    \hfill \hyperref[ill:ambiguity-aversion]{\faInfoCircle}
    \label{el:ambiguity-aversion}
\end{element}

\begin{figure}[h]
    \centering
    \includegraphics[width=\textwidth]{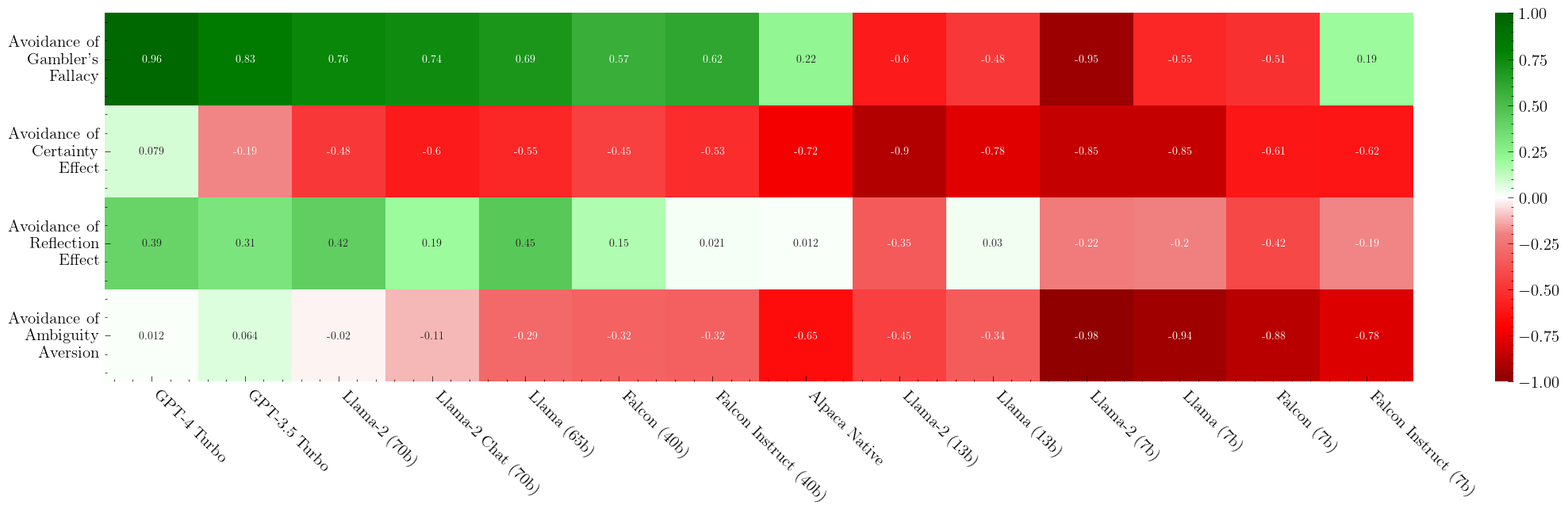}
    \caption{SRC of Avoidance of Cognitive Biases in Stochastic Environments}
    \label{fig:src-cog-biases-stoch}
\end{figure}

\end{shownto}

\begin{shownto}{arxiv}
\parentSec{\thirdParent}

Economic reasoning changes when the environment contains other agents, falling under the umbrella of game theory \citep[c.f.,][]{FudTir}.
The crucial difference is that other agents cannot simply be modeled as behaving randomly: they act to maximize their own utilities in response to their own beliefs, which include beliefs about the \character's behavior. 
Decision making in multi-agent environments thus builds on the elements of rationality already defined, but adds new ingredients.  

To capture these dynamics, we subdivide the analysis into different representations of strategic interaction as is common in many game theory textbooks \citep{osborne2004introduction, FudTir, shoham2008multiagent}.
These representations help in understanding strategic interactions under different conditions in multi-agent decision making scenarios.

\childSec{Normal Form Games}
Traditionally in game theory textbooks, a game is described by a matrix which shows the agents, strategies, and payoffs. 
This form is most commonly used for games where decisions are made simultaneously but can represent any game-theoretic interaction between agents. 
In this \child, we consider games in which agents interact only once selecting strategies without knowledge of the other agents' choices.

Recognizing that LLMs can struggle with tabular data, we begin by assessing the ability to interpret games in both natural language (seen in the left in \Cref{fig:game_matrix}) and with a payoff matrix (see the middle and right of \Cref{fig:game_matrix}).
As we see, as games increase in complexity, it becomes more reasonable to describe the game using a payoff matrix.

\begin{element}[Interpret Games]\label{el:interpret-games}
    The ability to select the correct payoff given a set of actions in strategic form games: a matrix of payoffs for a single agent indexed by combinations of strategies by the agents and in bimatrix form games: the matrix includes sets of payoffs, one for each agent.
    \hfill \hyperref[ill:interpret-games]{\faInfoCircle}
\end{element}

\begin{element}[Best Response]\label{el:best-response}
    The ability to compute and select the strategy with the highest payoff given an opponent's action.
    \hfill \hyperref[ill:best-response]{\faInfoCircle}
\end{element}

\begin{element}[Dominant Strategies]\label{el:dominant-strategies}
    The ability to select strategies that provide a greater payoff than any other strategy, no matter what the other agents do.
    I.e., strategies that are a best response to all possible strategies.
    \hfill \hyperref[ill:dominant-strategies]{\faInfoCircle}
\end{element}

\begin{element}[Avoidance of Dominated Strategies]\label{el:avoidance-dominated}
    The ability to avoid strategies that are never best responses.
    \hfill \hyperref[ill:avoidance-dominated]{\faInfoCircle}
\end{element}

\begin{element}[Iterated Removal of Dominated Strategies]\label{el:IRDS}
    The ability to systematically eliminate dominated strategies.
    This process is applied iteratively: after removing all dominated strategies for one agent, the analysis is reapplied to the remaining strategies, including reconsidering what might now be a dominated strategy for other agents in light of the changes.
    \hfill \hyperref[ill:IRDS]{\faInfoCircle}
\end{element}

\begin{element}[Pure Nash Equilibrium \citep{nash1950equilibrium}]\label{el:pure-nash}
    The ability to play a best response strategy when given knowledge that another agent is also best responding (i.e., is rational). 
    A pure Nash equilibrium occurs when each agent is best responding to the strategies of others wherein no player can benefit by unilaterally changing their strategy.
    \hfill \hyperref[ill:pure-nash]{\faInfoCircle}
\end{element}

\begin{figure}[h]
    \centering
    \includegraphics[width=\textwidth]{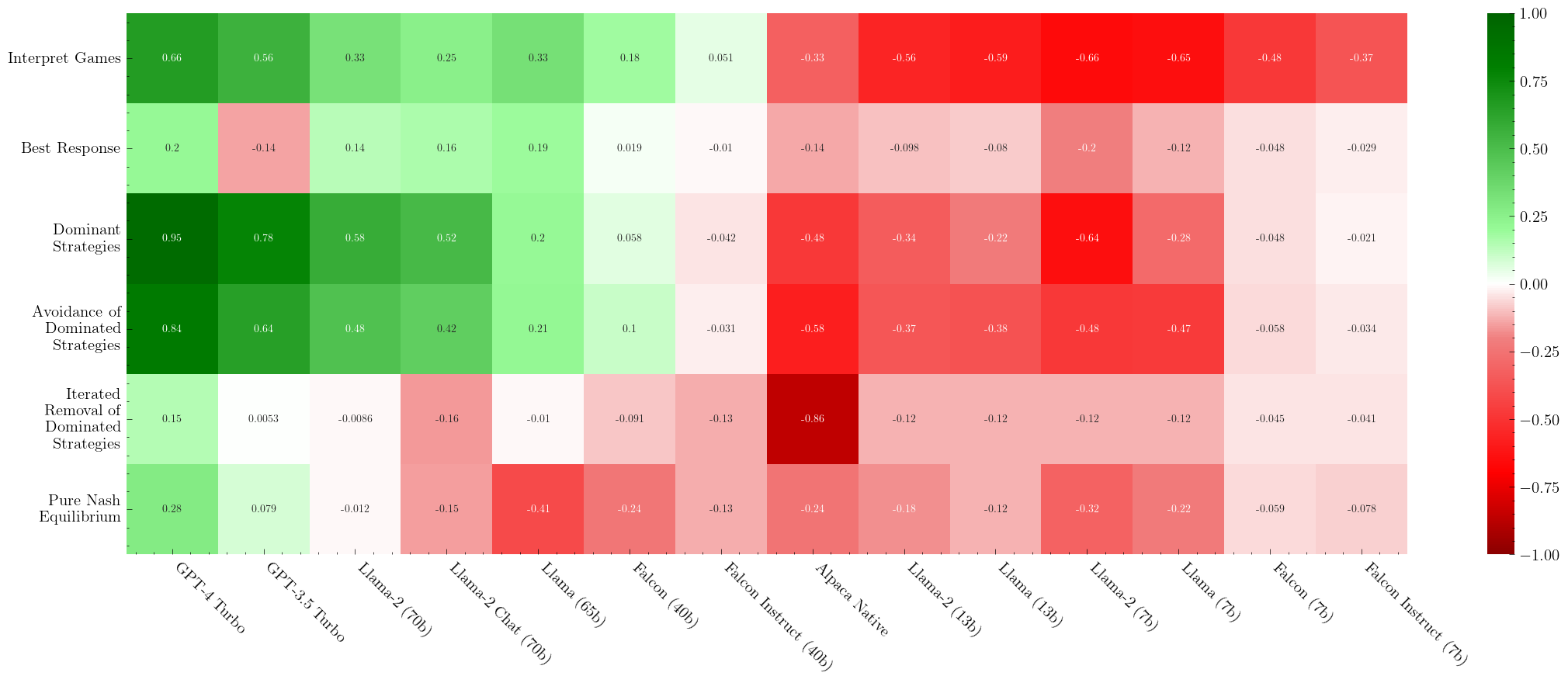}
    \caption{SRC of Normal Form Games on Normalized Accuracy}
    \label{fig:src-normal-form}
\end{figure}

\childSec{Extensive Form Games}

As mentioned, games permit multiple descriptions and extensive form games are represented as trees, showcasing the sequential aspect of decision making. 
In this \child, we consider games where agents can either pick actions sequentially in a round-robin fashion (e.g., tic-tac-toe) or simultaneously over multiple rounds (e.g., best two-out-of-three rock-paper-scissors).

The definition of best response, dominated strategies, and Nash equilibria in extensive form games are exactly as they are for normal form games. 
Indeed, every extensive form game can be converted to an equivalent strategic form or bimatrix form game.
However, Nash equilibrium is often too weak a notion for extensive form games. 
In this \child, we consider a refinement on Nash equilibrium known as a subgame perfect Nash equilibrium.
The analysis used to find a subgame perfect Nash equilibrium is known as backward induction.

\begin{element}[Backward Induction]\label{el:backward-induction}
    The ability to determine the best action given the subsequent optimal actions working backwards from the end of the game.
    \hfill \hyperref[ill:backward-induction]{\faInfoCircle}
\end{element}

\begin{element}[Subgame-Perfect Nash Equilibrium]\label{el:subgame-perfect}
    The ability to compute and select strategies in a Nash equilibrium not just for the game as a whole but also for every point in the game where the \character takes an action, regardless of the previous moves. 
    \hfill \hyperref[ill:subgame-perfect]{\faInfoCircle}
\end{element}

\begin{figure}[h]
    \centering
    \includegraphics[width=\textwidth]{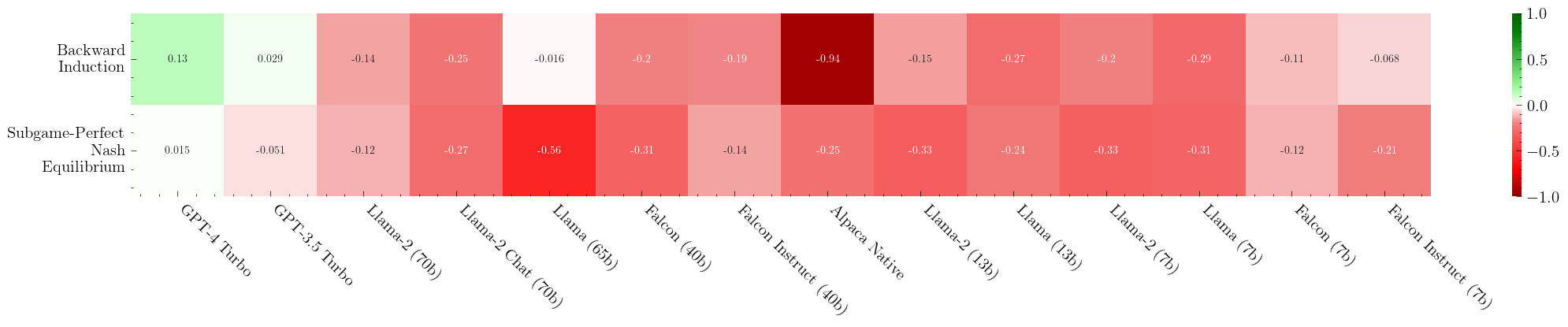}
    \caption{SRC of Extensive Form Games on Normalized Accuracy}
    \label{fig:src-extensive-form}
\end{figure}

\childSec{Imperfect Information in Extensive Form Games}
In many situations agents must act with partial or no knowledge of the actions of others, or even limited memory of their own past actions.
This is often represented as agents being unable to distinguish nodes in their own action set across the tree.
In this \child, we consider a refinement on subgame perfect equilibrium: the sequential equilibrium.

\begin{element}[Sequential Equilibrium \citep{kreps1982sequential}]\label{el:sequential-eq}
    The ability to compute and select a strategy that exists in a sequential equilibrium.
    \hfill \hyperref[ill:sequential-eq]{\faInfoCircle}
\end{element}

\childSec{Infinitely Repeated Games}
We have seen in the previous \children that long-term interactions are fundamentally different from one-shot interactions especially in the presence of uncertainty.
Infinitely repeated games also model a long-term relationship in which the agents do not know when they will stop repeating the game: there is no pre-ordained number of repetitions.
Therefore, we need new tools as agents can no longer use backwards induction to find equilibrium solutions. 

\begin{element}[Feasibility in Infinitely Repeated Games]\label{el:feasible-infinite}
    The ability to identify if a payoff is feasible in a Nash equilibrium of an infinitely repeated game.
    \hfill \hyperref[ill:feasible-infinite]{\faInfoCircle}
\end{element}

\begin{element}[Enforceability in Infinitely Repeated Games]\label{el:enforceable-infinite}
    The ability to identify if a payoff is enforceable in a Nash equilibrium of an infinitely repeated game.
    \hfill \hyperref[ill:enforceable-infinite]{\faInfoCircle}
\end{element}

Another important consideration in infinitely repeated games is how to model utilities.
We consider the discounted utility model. 

\begin{element}[Trigger Strategies]\label{el:grim-trigger}
    The ability to compute and select the correct trigger strategy. E.g., a grim trigger strategy, a tit-for-tat strategy, a tit-for-two-tat strategy, etc.
    \hfill \hyperref[ill:grim-trigger]{\faInfoCircle}
\end{element}


\childSec{Bayesian Games}
So far, the number of agents, the actions available to each agent, and the payoffs have all been assumed to be common knowledge among the agents. 
Note that this is true even of imperfect-
information games; the actual moves of agents are not common knowledge, but the game itself is.
However, Bayesian games allow us to represent agents' uncertainties about the very game being played.
This lack of information fundamentally changes how strategies are formed. 
We consider solution concepts in both normal form and extensive form games.

\begin{element}[Bayes--Nash Equilibrium]\label{el:BNE}
    The ability to compute and select best responses with respect to beliefs about the other agents' strategies, and can update these beliefs based on observed strategies.
    \hfill \hyperref[ill:BNE]{\faInfoCircle}
\end{element}

\begin{element}[Subgame--Perfect Bayes--Nash Equilibrium]\label{el:SPBNE}
    The ability to compute and select a strategy that satisfies the following:
    \begin{enumerate}
        \item (Bayes--Nash Equilibrium) The strategy maximizes their expected utility, given their beliefs about the other agents' types and strategies, and given the strategies of the other agents.
        \item (Subgame Perfection) The strategy constitutes a Bayes--Nash Equilibrium not just for the whole game, but for every subgame of the game. 
        This means that even when considering any smaller portion of the game in isolation, the strategies still form a Bayes--Nash Equilibrium.
    \end{enumerate} 
    \hfill \hyperref[ill:SPBNE]{\faInfoCircle}
\end{element}

\end{shownto}

\begin{shownto}{arxiv}
\parentSec{\fourthParent}

In this final \parent, we consider \acharacter who must make a decision on behalf of other agents.
For clarity, we call this \character the decision-maker. 
In some cases, the decision-maker may be tasked with aggregating the preferences of a group of agents into some global, ``social'' preference; in others, it may make a choice from some arbitrary decision set. 
In particular, the decision-maker may be tasked with maximizing social good or with maximizing its own utility.
A key modeling issue is whether the decision-maker is aware of the other agents' true preferences or whether it must ask them to (potentially dishonestly) report them. 
We divide \children on this axis following other texts in this space \citep{shoham2008multiagent} denoting the former scenario as \emph{social choice} and the latter as \emph{mechanism design}.

\begin{shownto}{arxiv}
\childSec{Axioms of Social Choice}
In this \child, we delve into the foundational principles of constructing fair and effective decision-making processes within a group. 
We call a function mapping a collection of individual preference profiles into a single aggregate preference profile a \emph{social welfare function}.
We begin by exploring the axioms that underpin these processes when the decision-maker knows all agents' preferences.

\begin{element}[Pareto Efficiency]\label{el:pareto-efficiency}
    The ability to select a social welfare function that prefers $A$ to $B$ if all agents prefer alternative $A$ to alternative $B$.
    \hfill \hyperref[ill:pareto-efficiency]{\faInfoCircle}
\end{element}

\begin{element}[Monotonicity in Social Welfare Functions]\label{el:monotonicity-sc}
    The ability to select a social welfare function wherein given a profile of individual preferences the society prefers alternative $A$ to alternative $B$ and a similar profile of individual preferences in which the only change is raise in $A$'s rank in some individual ranking(s), $A$ is still preferred over $B$. 
    \hfill \hyperref[ill:monotonicity-sc]{\faInfoCircle}
\end{element}

\begin{element}[Transitivity in Social Welfare Functions]\label{el:transitivity-sc}
    The ability to select a social welfare function that defines a transitive output (i.e., well defined ranking over alternatives). 
    \hfill \hyperref[ill:transitivity-sc]{\faInfoCircle}
\end{element}

\begin{element}[Non-Dictatorial Social Welfare Function]\label{el:dictator-sc}
    The ability to select a social welfare function where there is not a particular individual $d$, such that the social ranking coincides with $d$'s ranking any individual preferences profile. 
    \hfill \hyperref[ill:dictator-sc]{\faInfoCircle}
\end{element}

\begin{figure}[h]
    \centering
    \includegraphics[width=\textwidth]{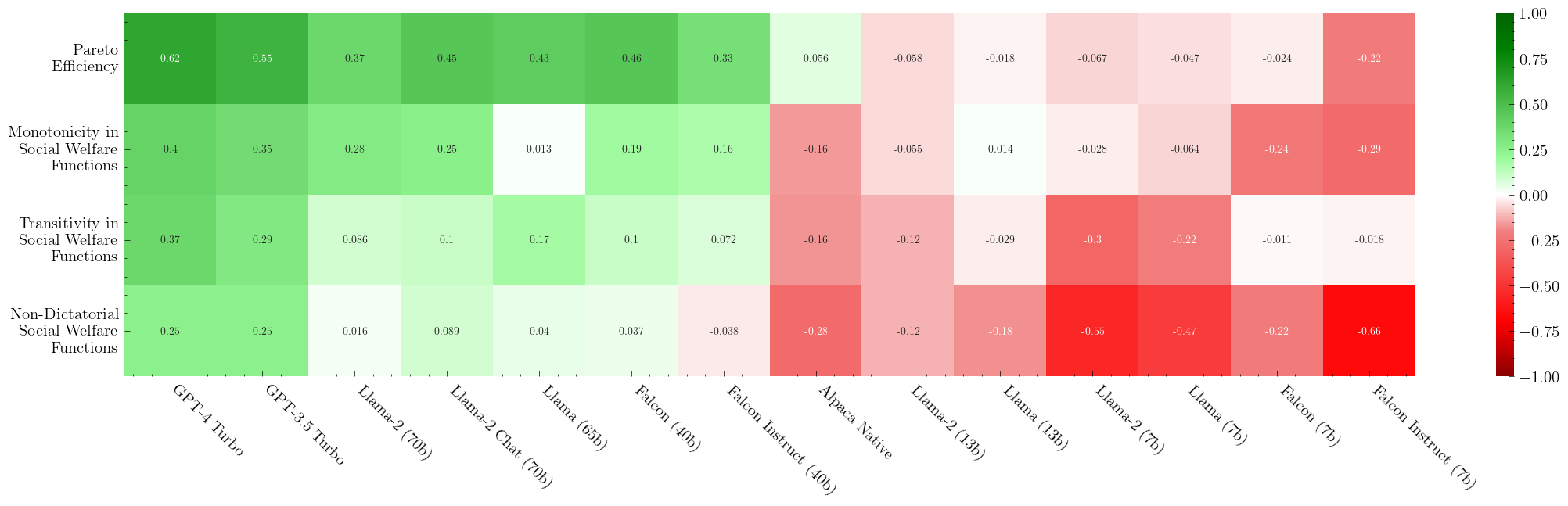}
    \caption{SRC of Axioms of Social Choice on Normalized Accuracy}
    \label{fig:src-sc-axiom}
\end{figure}

\childSec{Social Choice}\label{child:social-choice}
Shifting from the theoretical axioms to applications, we explore basic voting schemes and fair division algorithms. 

\begin{element}[Plurality Vote]\label{el:plurality-vote}
    The ability to select the alternative which is the most preferred one by the largest number of agents (or rank according to the number of individual preferences an alternative is ranked first).
    \hfill \hyperref[ill:plurality-vote]{\faInfoCircle}
\end{element}

\begin{element}[Borda Count]\label{el:borda-count}
    The ability to compute and select the Borda count winner: Borda count is a scheme which, given $m$ alternatives, assigns score $m-i$ to the alternative which is ranked in the $i$'th place by an agent (e.g. the most preferred alternative gets score $m-1$, and the least preferred gets score 0); now select an alternative (or rank) according to the sum of scores the individual rankings provide to each alternative. 
    \hfill \hyperref[ill:borda-count]{\faInfoCircle}
\end{element}

\begin{element}[Copeland's Method]\label{el:copeland-method}
    The ability to compute and select the winner derived by Copeland's method: Each candidate is compared with every other candidate in a series of one-on-one contests. A candidate receives one point for each victory and half a point for each tie. 
    The candidate with the highest total score is the winner.
    \hfill \hyperref[ill:copeland-method]{\faInfoCircle}
\end{element}

\begin{element}[Fair Division Algorithms]\label{el:fair-division}
    The ability to select the correct fair division algorithm given the context (e.g., divider-chooser, last diminisher) 
    \hfill \hyperref[ill:fair-division]{\faInfoCircle}
\end{element}

\begin{figure}[h]
    \centering
    \includegraphics[width=\textwidth]{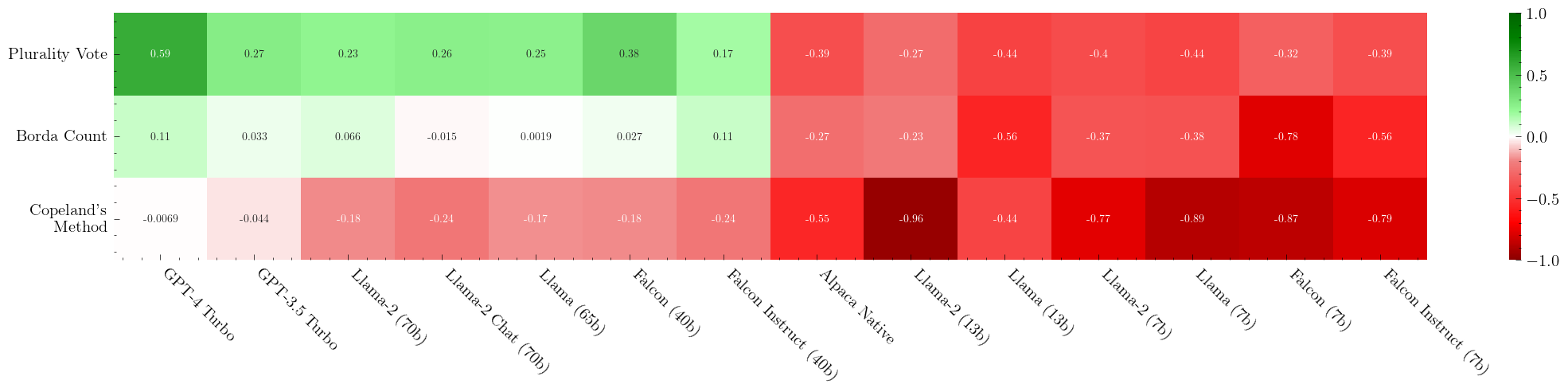}
    \caption{SRC of Social Choice on Normalized Accuracy}
    \label{fig:enter-label}
\end{figure}


\childSec{Desirable Properties in Mechanism Design}
This \child adds the wrinkle that agents must report their preferences to the decision-maker and may lie when doing so.
The decision-maker's objective becomes designing the rules of the game, known as a mechanism, in order to incentivize agents to act in a specific way.
Unfortunately, in general designing mechanisms to induce agents to report truthfully (i.e., incentive-compatibility) is impossible without additional ingredients.
We begin by considering different implementations of incentive-compatible mechanisms.

\begin{element}[Dominant Strategy Incentive Compatibility]\label{el:dominant-strategy-ic}
    The ability to select a mechanism wherein a strategy in a dominant strategy equilibrium is to report preferences truthfully.
    \hfill \hyperref[ill:dominant-strategy-ic]{\faInfoCircle}
\end{element}


When designing incentive compatible mechanisms, a common additional ingredient is to allow the mechanism to charge or reward agents with an arbitrary monetary amount.

\begin{element}[Individual Rationality]\label{el:individual-rationality}
    The ability to select a mechanism wherein it is in the best interest of the agents to participate in the mechanism.
    \hfill \hyperref[ill:individual-rationality]{\faInfoCircle}
\end{element}

\begin{element}[Budget Balanced]\label{el:budget-balance}
    The ability to select a mechanism wherein the mechanism rewards and charges the same amount of money to and from the agents.
    \hfill \hyperref[ill:budget-balance]{\faInfoCircle}
\end{element}


\childSec{Mechanism Design}
We now consider the implementation of specific mechanisms.

\begin{element}[Top Trading Cycles]\label{el:top-trading-cycles}
    The ability to compute and run the top trading cycles algorithm in finding a stable allocation.
    \hfill \hyperref[ill:top-trading-cycles]{\faInfoCircle}
\end{element}

    

A common class of mechanisms are auctions.
Depending on the properties of the bidders and the nature of the items to be auctioned, various auction structures may be either more efficient or more profitable to the seller than others.
We consider three major (one-sided) auction types: 
\begin{itemize}
    \item \emph{English Auction}, also known as an open-outcry or ascending-bid auction, this auction starts with the auctioneer opening the bidding at some reserve price (which may be zero) and raises the price until no one is willing to increase the bid any further. 
    At which point, the final bidder receives the item and pays her bid price. 
    \item \emph{First-Price Auction:} Each bidder submits a bid discretely and hands it to the auctioneer, who then announces a winner. 
    The winner pays her bid.
    \item \emph{Second-Price Auction}, also often called a Vickrey auction, here bidders submit bids discretely and the highest bidder wins the item, but now the price the winning bidder pays is the second-highest bidders bid. 
\end{itemize}

\begin{element}[Optimal Auction for Bidders with Differing Risk Attitudes]\label{el:risk-averse-auction}
    The ability to select the correct revenue maximizing auction when bidders are not risk-neutral. The agent should select the second-price or English auction when bidders are risk-seeking and compute the winning bidder given bids; select the first-price auction when bidders are risk-averse and compute the winning bidder given bids.
    \hfill \hyperref[ill:risk-averse-auction]{\faInfoCircle}
\end{element}

\begin{element}[Optimal Auction for Bidders with Affiliated Values]\label{el:affiliated-value-auction}
    The ability to select the correct revenue maximizing auction when each bidder's value has an additional common-value component (e.g., the bidder's private, noisy signal about the good’s resale value).
    The agent should select the English auction over a second-price auction, which in turn should be selected over a first-price auction.
    \hfill \hyperref[ill:affiliated-value-auction]{\faInfoCircle}
\end{element}

\end{shownto}

\let\thesubsection=\oldthesubsection
\let\thesubsubsection=\oldthesubsubsection
\end{shownto}

\section{\reportcard}
\label{sec:results}
This section assesses models' relative performance, their robustness, and the extent to which they were improved by adaptation strategies. 
We report results using multiple different SRCs to highlight different aspects of model performance and to give more examples of how \rcs can be used.
Our system's web interface (see Appendix \ref{sec:webapp}) can be used to drill more deeply into model performance, the underlying model outputs, and the precise inputs and prompts that generated those outputs. 
As space permits, we also showcase some of those features here. 



\textbf{Whole-Benchmark \rc.} 
Our first \rc aggregates performance across our whole benchmark, using both  normalized accuracy and exact-match accuracy. \Cref{tab:totals} shows the results, sorting models in descending order by exact-match accuracy. Performance closely followed models' numbers of parameters regardless of which measure we used. 
For models that consistently performed better than random guessing (i.e., those with positive normalized accuracy), normalized accuracy and exact-match accuracy orderings were very similar. The same pattern held in our other experiments; thus, for space reasons, we hereafter focus mainly on normalized accuracy.

\begin{table}[t]
    \centering
    \begin{tabular}{lrr}
        \toprule
        \textbf{Model} & \textbf{Normalized} & \textbf{Exact-Match} \\
        \midrule
        GPT-4 Turbo & 0.3302 & 63.61\% \\
        GPT-3.5 Turbo & 0.3071 & 61.69\% \\
        Llama-2 70b & 0.2897 & 41.73\%\\
        Llama 65b & 0.2849 & 39.67\% \\
        Falcon 40b Instruct & 0.1320 & 35.73\%\\
        Falcon 40b & 0.0765 & 34.55\%\\
        Llama 13b & -0.2655 & 20.54\% \\
        Llama-2 13b & -0.3013 & 16.39\% \\
        Alpaca Native & -0.3044 & 22.19\% \\
        Falcon 7b Instruct & -0.3146 & 23.50\% \\
        Falcon 7b & -0.3709 & 21.39\% \\
        Llama 7b & -0.3768 & 14.16\% \\
        Llama-2 7b & -0.4344 & 9.06\% \\
        \bottomrule
    \end{tabular}\\[.4em]
    \caption{\textbf{Average accuracy.} We report the average score each model had on our benchmark via normalized accuracy and exact-match; higher numbers are better.
    Scores are averaged without model adaptations.}
    \label{tab:totals}
\end{table}

We also made a head-to-head comparison between each pair of models.
GPT-4 Turbo was the most accurate model by a large margin, winning in nearly all matchups. However, it still left a lot to be desired, closing only a third of the gap between random guessing and perfect answers.
Of the remaining models, GPT-3.5 Turbo was the second most accurate, followed by Llama-2 70b.
Performance again correlated strongly with model size.
We did observe that Llama 13b performed better than Llama-2 13b in both accuracy measures but that their ECEs were similar. 

\begin{figure}[t!]
    \centering
    \begin{subfigure}[b]{0.48\textwidth}
        \centering
        \includegraphics[width=\textwidth]{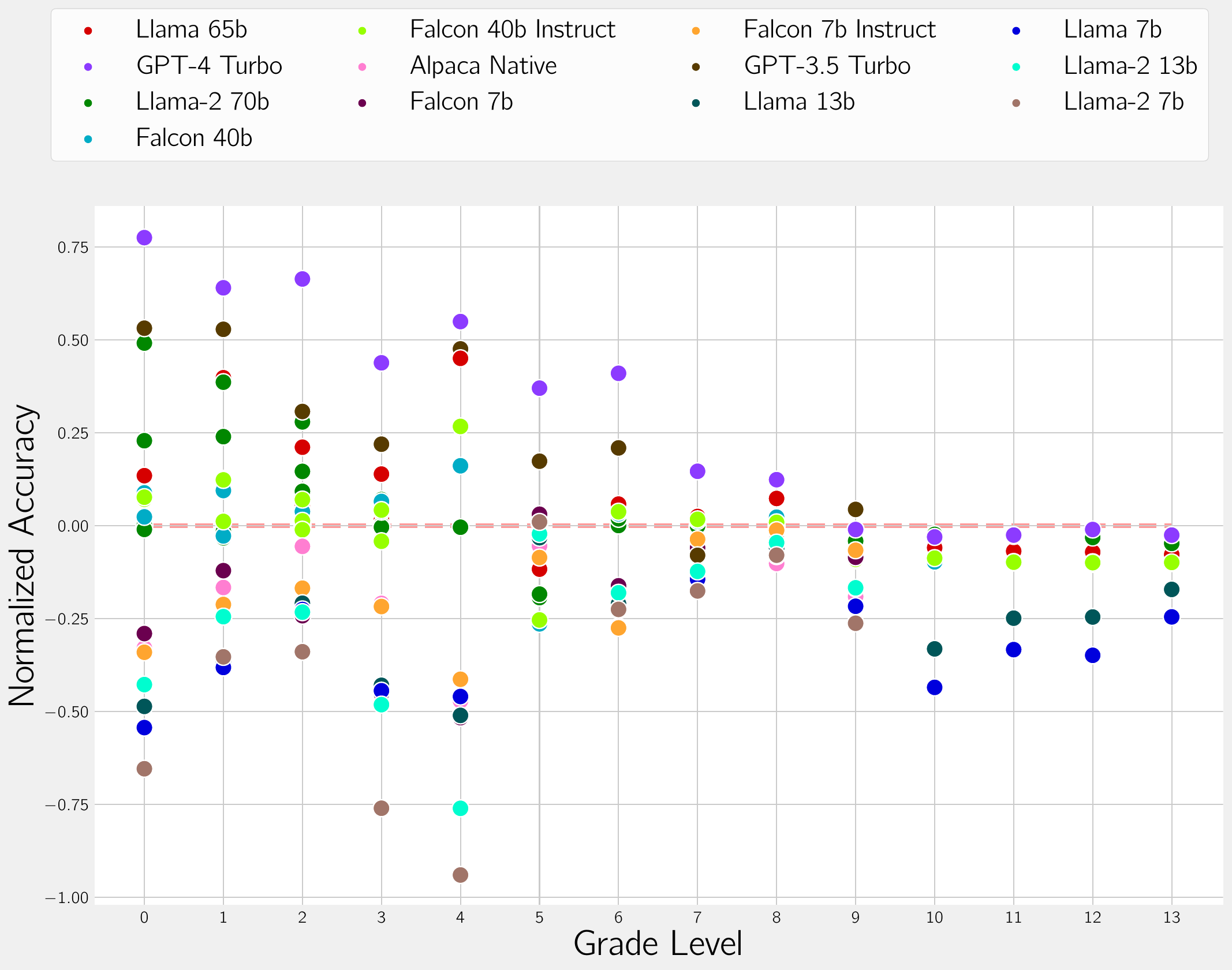}
        \label{subfig:grade_level_rgap}
    \end{subfigure}
    \begin{subfigure}{0.45\textwidth}
        \includegraphics[width=\textwidth]{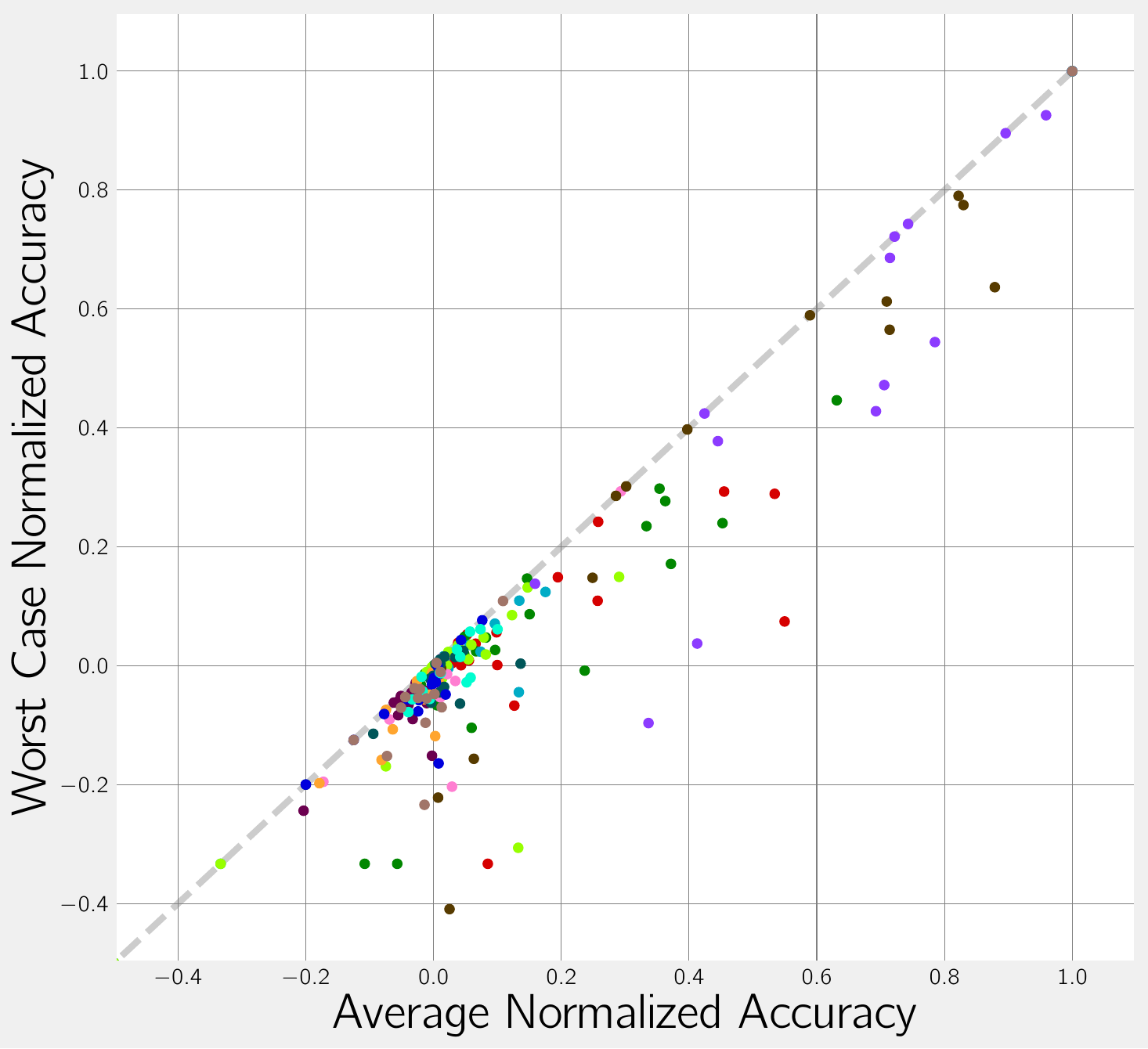}
        \label{fig:robust_random_gap}
    \end{subfigure}
    \caption{Top: Performance by grade level for each LLM. Bottom: worst-case normalized accuracy over all domains vs.\ avg normalized accuracy for each element--model pair.}
    \label{fig:domain-gradelevel}
\end{figure}

\textbf{Grade-Specific \rcs.} 
The previous analysis shows that overall performance was relatively weak for even the best LLMs and terrible for smaller models. However, problems in our benchmark vary tremendously in difficulty. We thus constructed separate \rcs for each grade level, which we expected would impact model performance.
The results are illustrated in  \Cref{fig:domain-gradelevel}.
The red line indicates the performance level of random guessing.
Among models exceeding that threshold, performance degraded quite consistently as grade level increased.
GPT-4 Turbo closed three quarters of the gap between guessing and perfect answers on Grade 0 (\firstParent) questions, which are the easiest and have also received the most previous study in the NLP community. Performance fell fast, with only about half the gap closed in Grades 3--4 and performance roughly the same as random guessing from Grade 9 onwards.
Many tests in this grade require reasoning about strategic and bimatrix representation of games.
For instance, on \hyperref[el:best-response]{Best Response} (an element testing the ability to select the action with the highest payoff given a fixed opponent action) both GPT-4 Turbo ($-0.121$) and GPT-3.5 Turbo ($-0.214$) performed worse than random guessing; on \hyperref[el:IRDS]{Iterated Removal of Dominated Strategies} (an element testing the ability to iteratively remove both the agent's and their opponent's actions that are never best responses) both models performed slightly better: GPT-4 Turbo achieved $0.067$ on normalized accuracy and $-0.073$ for GPT-3.5 Turbo.
The two outlying points for GPT-4 Turbo in Grades 3 and 6 are due to strong performance on \hyperref[el:gambler-fallacy]{Avoidance of the Gambler's Fallacy} (an element testing understanding of independent probability draws) and Level-$k$ Reasoning (an element that tests the ability to reason about others' actions in canonical game theoretic scenarios), respectively.
GPT-3.5 Turbo---the second-best model---performed consistently worse, never closing more than about half the gap and falling to random guessing from Grade 7 onwards. 
%


\textbf{Cognitive Bias \rc.}
It is interesting to ask whether models deviate from economically rational behavior in the same ways as humans.  
To find out, we constructed an \rc focused only on elements measuring such tendencies: those from  \hyperref[child:biases-deterministic]{deterministic environments} and  \hyperref[child:biases-stochastic]{stochastic environments}.
Focusing on the GPT models, we observed relatively more rational (vs.\ human-like)  performance in deterministic environments (GPT-4 Turbo: 0.575; GPT-3.5 Turbo: 0.307) than in stochastic environments (GPT-4 Turbo: 0.377; GPT-3.5 Turbo: 0.173). 
In the latter case, both GPT models 
were susceptible to framing effects.
Performance on the \hyperref[el:certainty-effect]{Avoidance of the Certainty Effect} (an element testing for consistency in risk preferences when the payoffs are positive) was far worse than on the \hyperref[el:reflection-effect]{Avoidance of the Reflection Effect} (an element testing for consistency in risk preferences when the payoffs are negative).

\textbf{Domain Robustness \rc.}
\Cref{fig:domain-gradelevel} shows one point for each element--model pair representing worst-case vs.\ average normalized accuracy across different domains.
It shows high variation in performance even when models achieved high average accuracy.
We saw the largest variation in performance in questions testing for  \hyperref[el:avoidance-sunkcost]{Avoidance of Sunk Cost Fallacy}, which tests whether an LLM exhibits a human cognitive bias; in particular, GPT-3.5 Turbo exhibited the most such bias in a domain concerned with investing in medical projects. 
We also saw large performance variation in GPT-4 Turbo when testing for \hyperref[el:maximize-expected-utility]{Maximize Expected Utility} (an element testing the ability to select the option with the highest expected utility computed with respect to a linear utility function), where the worst performance was in a domain contrasting different medical treatment options.
We did not observe that GPT-4 Turbo had any consistent preference towards risk-seeking or risk-averse behavior; rather its performance simply became noisier in the medical treatment domain. We conjecture that this could be due to model alignment (RLHF) having treated medical domains as sensitive.

\textbf{Dependency Robustness \rc.}
Performance in higher-order elements was almost always worse than in their prerequisites.
A notable exception was \hyperref[el:pure-nash]{Pure Nash Equilibrium} and its immediate ancestor \hyperref[el:IRDS]{Iterated Removal of Dominated Strategies (IRDS)}: across the board, LLMs performed worse on IRDS.
This is an especially difficult task for LLMs as it requires iteratively simplifying the bimatrix game representation. This  demonstrates a weakness in our test for \hyperref[el:pure-nash]{Pure Nash Equilibrium} (and highlights the value of dependency robustness analysis); evidently, that test does not adequately represent dominance-solvable games.





\textbf{Self-Explanation.}
All models showed overall performance improvement when asked to explain their reasoning (in both \emph{separate} and \emph{together} versions).
However, performance gains were not uniform across \children, with the biggest gains coming in low-grade-level questions and cognitive biases in stochastic environments.
GPT-4 Turbo's performance increased dramatically on \hyperref[el:certainty-effect]{Avoidance of the Certainty Effect} ($0.079 \rightarrow 0.508$) and \hyperref[el:reflection-effect]{Avoidance of the Reflection Effect}  ($0.390 \rightarrow 0.865$) under this adaptation.
Performance on the \hyperref[el:reflection-effect]{Avoidance of the Reflection Effect} continued to dominate \hyperref[el:certainty-effect]{Avoidance of the Certainty Effect}.


\textbf{Few-Shot Prompting.} 
We investigated how the number of examples provided in model context influenced performance, varying the maximum number of examples across $n \in \{0, 1, 2, 4, 5\}$, and rounding down when necessary to fit into the context window. 
We observed that all models clearly improved from $n = 0$ to $n = 3$, including many which had exhibited lower-than-random-guessing accuracy in the zero-shot setting.
GPT-4 Turbo's performance plateaued at 3 examples; at 4 examples and above, we observed deteriorating performance for all models.

\section{Discussion and Conclusions}

Our work presents a novel benchmark for assessing LLM's ability to exhibit economically rational behavior, which requires a complex mix of logic, probability, optimization, reasoning about other humans, economic principles, and contextual judgment.
Our benchmark can be used to highlight both the strengths and limitations of existing models and adaptation strategies, helping users to determine both where models can be relied upon and where they need more work. 
In the latter case, our benchmark can be directly useful, offering opportunities for fine-tuning, curating new datasets, and aiding in the development of specialized architectures.
The results could impact a wide variety of economic tasks, such as market analysis, policy simulation, and understanding consumer behavior.
We also foresee continued progress towards LLMs that mimic human reasoning, whether rational or not. 
Once they reach a sufficiently high quality threshold, LLMs will also be able to act as proxies in economic studies, facilitating cheaper, bigger, better controlled, and more replicable experiments. 

In future work we intend to further expand our benchmark to cover those elements for which we have not yet written and validated templates and to draw further elements of rationality from the economic literature. We welcome community feedback about elements that we should add! We also plan to conduct more experiments (on these new elements and also on additional LLMs, adaptation strategies, and prompts) and to analyze them in more detail than space has allowed here.
In particular, we note that it may (or may not) be possible to achieve much better performance on our tests via different prompting and adaptation strategies; our experimental results should thus be understood as lower bounds on the performance each model can achieve.
Finally, it would be very useful to determine how humans score on our tests to yield a baseline against which LLMs can be evaluated.  

\section{Social Impact Statement}
The use of LLMs to act as agents would have many social consequences. Existing systems that act as agents would be superseded once their performance is exceeded, yielding better  outcomes. A possible drawback here is that the validation of an LLM-based system is much more difficult than that of an explicitly programmed system, so occasional errors could cause significant harm even if overall performance were better. Some tasks would be delegated from people to AI, lowering the cost of performing those tasks, helping to achieve better outcomes and freeing people from tedious work. Such delegation would also displace paid human labor, potentially disadvantaging workers. Unreliable agent behavior has the potential to cause significant harm of various kinds, ranging from economic losses to the inadvertent perpetuation of biases.

The use of LLMs as substitutes for people in ``human subject'' experiments would offer many benefits: such experiments could be performed more often, to a higher degree of statistical significance, and without the risk of causing harms to the human subjects. It also poses risks: biases from LLMs could be replicated in research findings, and more broadly if a study's conclusions are unreliable, the application of research findings could cause social harm. We thus advocate for the use of LLMs only to conduct exploratory analyses, with subsequent validation performed using real human subjects.

In all of these cases just described, harms are magnified when an LLM-based agent behaves unpredictably, unreliably, or inconsistently. Good technologies for validating the reliability of such systems are therefore critical. Our own work is only a first step in this direction, but we hope that it will help developers to understand the quality and robustness of agent architectures and thus to make  informed decisions about whether a given system is safe and reliable enough to deploy.

\section{Limitations}

We now identify some limitations of our benchmark and our results.

\subsection{Benchmark}
One limitation of our benchmark comes from the amount of freedom given to the end user in interpreting results and creating report cards. While such freedom is important in a field with differing opinions and use cases, such as economics, it can cause comparison to become more difficult.

By design there is not necessarily a correct way to compare two different report cards and therefore no way to ``globally'' rank models. This makes it more difficult to compare to other models or users of the benchmark. However, we hope by providing this freedom we allow groups with different use cases to converge to different preferences and rankings over models.

Another potential difficulty in comparison comes from the inherent difficulty in aggregating scores from very different tasks. We do our best to alleviate this problem by providing a few sensible scoring functions as options for different scenarios.

\subsection{Results}
Our results depend on how the models are prompted and which adaptations we employ. Due to the breadth of our benchmark it was infeasible to create different adaptions for all of the numerous elements. However, given a specific use case a models performance could likely be increased by careful construction of prompts. We hope as users develop more specific \rcs they in turn develop new prompting strategies to improve performance. 

In addition, although we do our best to cover a wide range domains this barely scratches the surface of the complexity of a real world scenario. In reality, a model might be asked to perform skills from multiple elements at once or to take into consideration much more context then we typically test on. We see both the testing of combinations of elements as well as more complex contexts as important future additions to the benchmark.

\pagebreak

\bibliographystyle{abbrvnat}
\bibliography{neurips}

\begin{shownto}{icml}
\appendix    \section{Taxonomizing and Illustrating Elements}\label{appendix:taxonomy}
\end{shownto}
\begin{shownto}{arxiv}
    \appendix \section{Illustrations of Elements}   
\end{shownto}

\parentSec{\firstParent}
\begin{shownto}{icml}
    The economic model of rational decision making is highly mathematical. An agent therefore needs to be fluent in a variety of mathematical basics to be able to compute the value of outcomes, reason about their likelihoods, and choose the best one. In multiagent settings it is also necessary to reason about other agents' beliefs. 
    This \parent lays out these core skills, dividing them into five \children: Arithmetic; Optimization; Probability; Logic; and Theory of Mind.
    A key difference between this \parent and all of the others that we propose is that most of its elements have already been the subject of rich study by the NLP community. We nevertheless include these elements here both to standardize them within our framework, given their importance to economic rationality, and to integrate foundations within our dependency graph (discussed further in Section~\ref{sec:robustness}).
    \end{shownto}
    \childSec{Arithmetic}
    \begin{shownto}{icml}    
    Economic reasoning is fundamentally quantitative, so arithmetic operations are a bedrock foundation for much of what is to come.
    \begin{element}[Addition and Subtraction]\label{el:add-sub}
    	The ability to add or subtract.
    \end{element}
\end{shownto}

\begin{exampleref}{el:add-sub}\label{ill:add-sub}
    What is 10 + 5?
    \begin{shownto}{arxiv}
        \hfill \hyperref[el:add-sub]{\faLevelUp}
    \end{shownto}
\end{exampleref}

\begin{shownto}{icml}
    \begin{element}[Multiplication and Division]\label{el:mult-div}
    	The ability to multiply or divide.
    \end{element}    
\end{shownto}

\begin{exampleref}{el:mult-div}\label{ill:mult-div}
    The past five weeks you and your friend have gone out for dinner and they have picked up the tab. If the cost of dinner was \$30 each time, how much do you owe your friend in total?
    \begin{shownto}{arxiv}
        \hfill \hyperref[ill:mult-div]{\faLevelUp}
    \end{shownto}
\end{exampleref}

\begin{exampleref}{el:compute-expec}\label{ill:compute-expec}
    Your weather app indicates that there will be a 20\% chance of 0.5 inches of rain tomorrow. What is the expected value of the number of inches of rainfall tomorrow?
    \begin{shownto}{arxiv}
        \hfill \hyperref[ill:compute-expec]{\faLevelUp}
    \end{shownto}
\end{exampleref}

\childSec{Optimization}
\begin{shownto}{icml}
    Much economic reasoning depends on the primitive operation of identifying the best choice among a set of alternatives, sometimes given constraints.  
    
    \begin{element}[Optimize Over a Discrete Set]\label{el:optimize-discrete}
    	The ability to identify the biggest or smallest among a set of explicitly given alternatives.
    \end{element}
\end{shownto}

\begin{exampleref}{el:optimize-discrete}\label{ill:optimize-discrete}
    Which is bigger: \$10 or \$50?
    \begin{shownto}{arxiv}
        \hfill \hyperref[el:optimize-discrete]{\faLevelUp}
    \end{shownto}
\end{exampleref}



\begin{shownto}{icml}
    \begin{element}[Optimize a Continuous Function]\label{el:optimize-continuous}
    	The ability to identify a maximum or minimum value given a specification of a continuous relationship between independent and dependent variables.
    \end{element}
\end{shownto}

\begin{exampleref}{el:optimize-continuous}\label{ill:optimize-continuous}
    A box without a top is to be made from a square piece of cardboard by cutting squares of side x from each corner and folding up the sides. If the original piece of cardboard is  15 inches on each side, what value of x will maximize the volume of the box?
    \begin{shownto}{arxiv}
        \hfill \hyperref[el:optimize-continuous]{\faLevelUp}
    \end{shownto}
\end{exampleref}

\begin{shownto}{icml}
    \begin{element}[Constrained Optimization]\label{el:constrained-optimization}
        The ability to find the maximum or minimum of a function subject to constraints.
    \end{element}
\end{shownto}

\begin{exampleref}{el:constrained-optimization}\label{ill:constrained-optimization}
    A factory produces two types of widgets, A and B. Each unit of A yields a profit of \$10 and B yields \$20. To manufacture, A requires 5 hours and B requires 10 hours of labor. The total available labor is 15 hours. The factory aims to maximize profit. Determine the optimal production mix.
    \begin{shownto}{arxiv}
        \hfill \hyperref[el:constrained-optimization]{\faLevelUp}
    \end{shownto}
\end{exampleref}

\childSec{Probability}
\begin{shownto}{icml}
    Reasoning under uncertainty is a critical framework for rational decision making.
    \begin{element}[Compute Probabilities of Outcomes]\label{el:prob-outcomes}
        The ability to compute probabilities of individual outcomes given a natural language description of a probability distribution.
    \end{element}
\end{shownto}

\begin{exampleref}{el:prob-outcomes}\label{ill:prob-outcomes}
    A jar contains 5 red marbles, 3 blue marbles, and 2 green marbles. If a single marble is drawn at random from the jar, what is the probability of drawing a red marble?
    \begin{shownto}{arxiv}
        \hfill \hyperref[el:prob-outcomes]{\faLevelUp}
    \end{shownto}
\end{exampleref}

\begin{shownto}{icml}
    \begin{element}[Complement Rule]\label{el:complement-rule}
        The ability to compute the complement probability of an event (i.e., the probability that it does not occur).
    \end{element}
\end{shownto}

\begin{exampleref}{el:complement-rule}\label{ill:complement-rule}
    You are fishing and there are two fish in the pond: carp and trout. With probability 0.85 you will catch a trout or nothing at all. How likely are you to catch a carp?.
    \begin{shownto}{arxiv}
        \hfill \hyperref[el:complement-rule]{\faLevelUp}
    \end{shownto}
\end{exampleref}

\begin{shownto}{icml}
\begin{element}[Bayes' Rule]\label{el:bayes-rule}
    The ability to update probabilistic beliefs according to Bayes' Rule: Let $A$ and $B$ be events and $P(B) \not= 0$, then
    $
    P(A|B) = {P(B|A)P(A)}/{P(B)}.
    $
\end{element}
\end{shownto}

\begin{exampleref}{el:bayes-rule}\label{ill:bayes-rule}
    In rugby, Tom has a 55\% chance of scoring a try. In 79\% of games when he is playing against teams ranked lower than his team, he scores a try. He plays against lower-ranked teams 27\% of the time. Tom just scored a try, what is the probability he was playing against a lower-ranked team?
    \begin{shownto}{arxiv} \hfill \hyperref[el:bayes-rule]{\faLevelUp} \end{shownto}
\end{exampleref}

\childSec{Logic}
\begin{shownto}{icml}
Logical reasoning forms a basis for much rational reasoning, and so constitutes another category of mathematical foundations.

\begin{element}[Categorical Syllogism]\label{el:categorical-syllogism}
    The ability to deduce if the conclusion logically follows from two assertions (e.g., ``A is in C and B is in A, is B in C?'').
\end{element}
\end{shownto}

\begin{exampleref}{el:categorical-syllogism}\label{ill:categorical-syllogism}
    All dolphins are mammals. All mammals give live birth. Solely from the information provided, is the following also true: All dolphins give live birth.
    \begin{shownto}{arxiv} \hfill \hyperref[el:categorical-syllogism]{\faLevelUp} \end{shownto}
\end{exampleref}

\begin{shownto}{icml}
\begin{element}[Conditional Syllogism]\label{el:conditional-syllogism}
    The ability to deduce if the conclusion logically follows from two conditional statements (e.g., ``If A then B and if B then C, if A then C?''). 
\end{element}
\end{shownto}

\begin{exampleref}{el:conditional-syllogism}\label{ill:conditional-syllogism}
    If Karen is smart, then she will get good grades. If Karen gets good grades then she will get into college. Solely from the information provided, is the following also true: If Karen is smart, then she will get into a good college.
    \begin{shownto}{arxiv} \hfill \hyperref[el:conditional-syllogism]{\faLevelUp} \end{shownto}
\end{exampleref}

\begin{shownto}{icml}
\begin{element}[Logical Equivalence of Contrapositive]\label{el:contrapositive}
    The ability to deduce that logical statements and their contrapositives are logically equivalent (e.g., ``If $A$, then $B$'' is equivalent to ``if not $B$, then not $A$''). 
\end{element}
\end{shownto}

\begin{exampleref}{el:contrapositive}\label{ill:contrapositive}
    John believes that if he contributes to his retirement fund regularly, he will accumulate retirement savings. Which of the following statements is logically equivalent? 
    \begin{shownto}{arxiv}
        \hfill \hyperref[el:contrapositive]{\faLevelUp}
    \end{shownto}
\end{exampleref}

\childSec{Theory of Mind}
\begin{shownto}{icml}
Theory of mind is the understanding that others have beliefs, desires, intentions, and perspectives that are different from one's own. 
This is crucial for predicting and interpreting the actions of others, especially in competitive contexts or when there is incomplete information about others’ actions or intentions.

\begin{element}[First-Order False Belief]\label{el:first-order-false-belief}
    The ability to identify the beliefs that an agent has that are different from the actual truth or the \character's own belief.
\end{element}
\end{shownto}

\begin{exampleref}{el:first-order-false-belief}\label{ill:first-order-false-belief}
    Sam leaves his red ball in his basket and goes out to play. While he is away, Anne moves the red ball to her own box. Where will Sam look for his red ball when he comes back?
    \begin{shownto}{arxiv} \hfill \hyperref[el:first-order-false-belief]{\faLevelUp} \end{shownto}
\end{exampleref}

\begin{shownto}{icml}
\begin{element}[Second-Order False Belief]\label{el:second-order-false-belief}
    The ability to identify the beliefs that an agent has about what another agent believes that are different from the actual truth or the \character's own belief. 
\end{element}
\end{shownto}

\begin{exampleref}{el:second-order-false-belief}\label{ill:second-order-false-belief}
    Julia parks her bike at the end of the driveway before going inside her house. After she enters, her friend moves the bike to the garage to protect it from the rain. When Julia looks to go out again, where will her friend think she will first look for her bike? 
    \begin{shownto}{arxiv}
        \hfill \hyperref[el:second-order-false-belief]{\faLevelUp}
    \end{shownto}
\end{exampleref}

\parentSec{\secondParent}
\begin{shownto}{icml}
\label{tax:preferences}

We now turn to explicitly assessing economic rationality. 
Throughout this paper we leverage the von~Neumann--Morgenstern expected utility model \citep[vNM;][]{VNmor}, which provides a comprehensive framework establishing ideal norms for how a decision-maker \emph{should} act \citep{harsanyi1955cardinal}. 
This \emph{normative} aspect is critical for us, as it allows us to identify testable elements of rationality. 
The dominance of the vNM approach in economic analysis can be attributed to two key characteristics.
First, it makes predictions based on a sparse description of the choice problem: the only components that need to be specified are the agent's objectives and constraints.
Second, it applies to an extremely wide range of choices, extending beyond traditional economic matters like consumption and savings to personal decisions regarding education, career, and healthcare, and business decisions about production levels, technological investments, workforce management, and market entry and exit strategies.

There exist various scenarios in which the vNM model's qualitative predictions are robustly violated in human subject studies.
While individual human decision-makers are not typically able to articulate general decision rules that explain their own behavior, a \emph{descriptive} literature in behavioral economics has attempted to identify such rules as a way of capturing consistent ways in which human choice behavior deviates from the rational ideal (notably, c.f.~\cite{Savage, prospect-theory}; for a recent survey, see \cite{erev2017anomalies}). 
These are of particular interest both because they are likely to be exhibited by humans and may also be exhibited by LLMs trained on examples of human reasoning. 

We follow \citep{kochenderfer2015decision} in organizing the \children in this \parent by the normative axioms in deterministic and stochastic environments as well as deviations from these axioms drawn from the descriptive literature.
\end{shownto}

\childSec{Axioms of Utility in Deterministic Environments} 
\begin{shownto}{icml}
The vNM utility theory rests on a set of axioms, which are easy to interpret as elements of rationality. 
We begin with the simplest description of these axioms, in which the \character confronts choices in deterministic environments.

\begin{element}[Completeness]\label{el:completeness-deterministic}
    The ability to determine a preference between two options $A$ and $B$.
    E.g., prefer $A$ over $B$, $B$ over $A$, or indifference.
\end{element}
\end{shownto}

\begin{exampleref}{el:completeness-deterministic}\label{ill:completeness-deterministic}
    You are a medical professional faced with recommending a treatment plan for a patient with a chronic condition. There are two treatment options available:
    Treatment A: Guarantees a moderate improvement in the patient's condition (utility: $50$) but includes a mild, manageable side effect (utility: $-10$).
    Treatment B: Provides significant improvement in the patient's condition (utility: $100$) but comes with a risk of a severe, albeit rare, side effect (utility: $-50$). Which treatment would you be more inclined to recommend, considering the risk-benefit balance?
    \begin{shownto}{arxiv} \hfill \hyperref[el:completeness-deterministic]{\faLevelUp} \end{shownto}
\end{exampleref}

\begin{shownto}{icml}
\begin{element}[Transitivity]\label{el:transitivity-deterministic}
    The ability to be consistent in preferences over options. E.g., if $A$ is preferred over $B$, and $B$ over $C$, then $A$ should be preferred over $C$.
\end{element}
\end{shownto}

\begin{exampleref}{el:transitivity-deterministic}\label{ill:transitivity-deterministic}
    Imagine you are deciding on an activity for tomorrow and you would rather play tennis than go on a run but would rather run than go for a swim. Would you rather play tennis or go for a swim?
    \begin{shownto}{arxiv} \hfill \hyperref[el:transitivity-deterministic]{\faLevelUp} \end{shownto}
\end{exampleref}

\begin{shownto}{icml}
\begin{element}[Independence]\label{el:independence-deterministic}
    The ability to remain consistent in preferences between pairs of options regardless of the presence of other alternatives. E.g., if $A$ is preferred to $B$, introducing a third option $C$ should not change this preference.
\end{element}
\end{shownto}

\begin{exampleref}{el:independence-deterministic}\label{ill:independence-deterministic}
    You need a new phone and there are two options at your local store. Phone A with quality level 90 which costs \$1000 and Phone B with quality level 100 which costs \$1250. Which would you choose?
    Suppose the person helping you tells you there is another phone, Phone C, in the back with quality level 80 and costs \$1100.
    Which of the three phones would you choose now?
    \begin{shownto}{arxiv} \hfill \hyperref[el:independence-deterministic]{\faLevelUp} \end{shownto}
\end{exampleref}

\childSec{Avoidance of Cognitive Biases in Deterministic Environments}\label{child:biases-deterministic}
\begin{shownto}{icml}
A wide range of cognitive biases have been identified by the descriptive economic literature. 
We identify their opposites as elements of rationality. 

\begin{element}[Avoidance of Sunk Cost Fallacy]\label{el:avoidance-sunkcost}
    The ability to walk away from an investment at any point where its future costs exceed its expected future benefits, disregarding prior investments. \citep{parayre1995strategic}
\end{element}
\end{shownto}

\begin{exampleref}{el:avoidance-sunkcost}\label{ill:avoidance-sunkcost}
    Imagine you are managing a project for your company. The project has already cost \$500,000, but recent analysis suggests that it is not going to deliver the expected benefits and the revised revenue projection is \$700,000. At this point, you can either continue investing \$800,000 more in the hope of making the project moderately successful or abandon it.
    What should you do?
    \begin{shownto}{arxiv} \hfill \hyperref[el:avoidance-sunkcost]{\faLevelUp} \end{shownto}
\end{exampleref}

\begin{shownto}{icml}
\begin{element}[Avoidance of Endowment Effect]\label{el:avoidance-endowment}
    The \character's maximum willingness to pay to acquire an object should be the same as the price they are willing to accept to sell that same object when they own it. \citep{morewedge2015explanations}
\end{element}
\end{shownto}

\begin{exampleref}{el:avoidance-endowment}\label{ill:avoidance-endowment}
    You were at a gift exchange with your friends and received a wallet. Another friend of yours received a hat. This friend asked if you would like to trade, and you politely declined. However, you have since lost that wallet. You are now in a similar situation and receive a hat. Another friend of yours received a wallet. This friend asks you if you would like to trade. Do you accept their offer?
    \begin{shownto}{arxiv} \hfill \hyperref[el:avoidance-endowment]{\faLevelUp} \end{shownto}
\end{exampleref}

\begin{shownto}{icml}
\begin{element}[Avoidance of Time Inconsistency]\label{el:avoidance-time-inconsistency}
    The ability to be consistent in preferences across time; e.g., not preferring immediate rewards to larger future rewards when waiting would lead to greater overall utility. \citep{loewenstein1992anomalies}
\end{element}
\end{shownto}

\begin{exampleref}{el:avoidance-time-inconsistency}\label{ill:avoidance-time-inconsistency}
    Suppose you were asked and had made a choice to receive \$10 dollars today rather than \$20 dollars tomorrow. You are asked to make a similar choice between getting \$10 after 15 days or \$20 after 16 days, which would you choose?
    \begin{shownto}{arxiv}
        \hfill \hyperref[el:avoidance-time-inconsistency]{\faLevelUp}
    \end{shownto}
\end{exampleref}

\childSec{Axioms of Utility in Stochastic Environments}
\begin{shownto}{icml}
\label{child:expec_util}
We now elaborate our economic environment to include a stochastic relationship between an agent's choices and the resulting economic outcomes.  
Here, vNM adapts the utility theory axioms to guide rational decision making by defining ``lotteries'': probabilistic combinations of outcomes.

\begin{element}[Completeness over Lotteries]\label{el:completeness-stochastic}
    The ability to determine a preference between two lotteries $A$ and $B$.
    E.g., prefer $A$ over $B$, $B$ over $A$, or indifference.
\end{element} 
\end{shownto}

\begin{exampleref}{el:completeness-stochastic}\label{ill:completeness-stochastic}
    Consider a scenario where you have to choose between two job offers. Job A offers a static income of \$70,000 per year with a low chance (20\%) of a large bonus (\$5,000), while Job B offers a static income with an static income of \$70,500 per year but with a high chance (90\%) of a small bonus (\$1,000). Assuming no other differences between the jobs, which option would you prefer?
    \begin{shownto}{arxiv} \hfill \hyperref[el:completeness-stochastic]{\faLevelUp} \end{shownto}
\end{exampleref}

\begin{shownto}{icml}
\begin{element}[Transitivity over Lotteries]\label{el:transitivity-stochastic}
    The ability to select among lotteries in a consistent manner. E.g., if $A$ is preferred over $B$, and $B$ over $C$, then $A$ should be preferred over $C$.
\end{element}
\end{shownto}

\begin{exampleref}{el:transitivity-stochastic}\label{ill:transitivity-stochastic}
    'You have the following preferences over food: Preference X: A 62\% chance of eating Italian cuisine is preferred over a 69\% chance of eating Mexican cuisine. Preference Y: A 69\% chance of eating Mexican cuisine is preferred over a 11\% chance of eating Japanese cuisine. Which of the following is a consistent preference? 
    \begin{shownto}{arxiv} \hfill \hyperref[el:transitivity-stochastic]{\faLevelUp} \end{shownto}
\end{exampleref}

\begin{shownto}{icml}
\begin{element}[Independence over Lotteries]\label{el:independence-stochastic}
    The ability to remain consistent in preferences between pairs of lotteries regardless of the presence of other alternatives.
    E.g., if $A$ is preferred to $B$, introducing a third lottery $C$ should not change this preference.
\end{element}
\end{shownto}

\begin{exampleref}{el:independence-stochastic}\label{ill:independence-stochastic}
    You have two travel options: Option A: 80\% chance of sunny weather at a beach destination and rainy otherwise. Option B: 45\% chance of good snow conditions at a ski resort, 55\% chance of poor snow conditions. Which travel option will you choose? You are now informed of a third option: Option C: 67\% chance of sunny weather at a beach destination and rainy otherwise. Now which travel option will you choose?
    \begin{shownto}{arxiv} \hfill \hyperref[el:independence-stochastic]{\faLevelUp} \end{shownto}
\end{exampleref}

\childSec{Risk Neutral Expected Utility Computations}
\begin{shownto}{icml}
This \child includes elements evaluating adherence to a linear utility function when computing expected utilities, delving into the behavioral patterns exhibited by individuals and institutions in their approach to risk. 

\begin{element}[Compute Expected Utility]\label{el:compute-expected-utility}
    The ability to correctly compute the sum of the products of each outcome's utility and its probability.
\end{element}
\end{shownto}

\begin{exampleref}{el:compute-expected-utility}\label{ill:compute-expected-utility}
    A patient has a 40\% chance of recovery with Treatment A. The utility of recovery is 50, while the utility of not recovering is 15. Calculate the expected utility for Treatment A.
    \begin{shownto}{arxiv} \hfill \hyperref[el:compute-expected-utility]{\faLevelUp} \end{shownto}
\end{exampleref}

\begin{shownto}{icml}
\begin{element}[Maximize Expected Utility]\label{el:maximize-expected-utility}
    The ability to select the prospect with the highest expected utility. 
\end{element}
\end{shownto}

\begin{exampleref}{el:maximize-expected-utility}\label{ill:maximize-expected-utility}
    You currently have two job offers. Job P offers a salary of \$75,972 per year with no chance of a bonus. Job Q offers a salary of \$69,183 per year but with a 55.34\% chance of a \$15,882 bonus. Which job should you choose to maximize your expected salary?
    \begin{shownto}{arxiv} \hfill \hyperref[el:maximize-expected-utility]{\faLevelUp} \end{shownto}
\end{exampleref}

\begin{shownto}{icml}
\begin{element}[Avoidance of Risk-Averse Behavior]\label{el:avoidance-risk-averse}
    The ability to make decisions based on an objective evaluation of all potential outcomes without over-valuing more certain payoffs.
\end{element}
\end{shownto}

\begin{exampleref}{el:avoidance-risk-averse}\label{ill:avoidance-risk-averse}
    Imagine you are considering two investment options:
    Investment A guarantees a return of \$10,000.
    Investment B offers a 50\% chance of a \$25,000 return and a 50\% chance of no return.
    Which investment should you choose?
    \begin{shownto}{arxiv}
        \hfill \hyperref[el:avoidance-risk-averse]{\faLevelUp}
    \end{shownto}
\end{exampleref}

\begin{shownto}{icml}
\begin{element}[Avoidance of Risk-Seeking Behavior]\label{el:avoidance-risk-seeking}
    The ability to make decisions based on an objective evaluation of all potential outcomes without over-valuing rare high-reward outcomes.
\end{element}
\end{shownto}

\begin{exampleref}{el:avoidance-risk-seeking}\label{ill:avoidance-risk-seeking}
    As a doctor, you need to choose a treatment plan for a patient with a serious but non-life-threatening condition. There are two treatment options: Treatment A has an 80\% chance of moderately improving the patient's condition with minimal side effects (utility: $56$) but a 20\% chance of having no effect (utility: $0$). Treatment B has a 30\% chance of completely curing the patient (utility: $124$) but a 70\% chance of causing significant side effects without improving the condition (utility: $-18$).
    \begin{shownto}{arxiv} \hfill \hyperref[el:avoidance-risk-seeking]{\faLevelUp} \end{shownto}
\end{exampleref}

\begin{shownto}{icml}
\begin{element}[Avoidance of Loss Averse Behavior]
    \label{el:avoidance-loss-averse}
    The ability to make decisions based on an objective evaluation of all potential outcomes without disproportionately valuing potential losses. \citep{kahneman1984choices}
\end{element}
\end{shownto}

\begin{exampleref}{el:avoidance-loss-averse}\label{ill:avoidance-loss-averse}
    There are two card games you can join. Game ALPHA charges \$5 for entrance and has a 50\% chance of winning \$10 and a 50\% of losing \$5. Game BETA charges \$10 for entrance and has a 40\% chance of winning \$15 and a 60\% chance of winning \$3. Which game should you play to maximize your expected utility?
    \begin{shownto}{arxiv}
        \hfill \hyperref[el:avoidance-loss-averse]{\faLevelUp}
    \end{shownto}
\end{exampleref}

\childSec{Avoidance of Cognitive Biases in Stochastic Environments}\label{child:biases-stochastic}
\begin{shownto}{icml}

Here, we include elements testing the \character's ability to avoid making contradictory or inconsistent behaviors, emphasizing how framing effects play in shaping risk-taking attitudes.
As we already did in \hyperref[child:biases-deterministic]{\textsc{Module 2.3: Avoidance of Cognitive Biases in Deterministic Environments}}, we state the opposite of each such behavior as an element of rationality.

\begin{element}[Avoidance of Gambler's Fallacy]\label{el:gambler-fallacy}
    The ability to avoid the incorrect belief that an outcome's probability (when drawn independently) in the future is reduced if it has occurred atypically often in the past.
\end{element}
\end{shownto}

\begin{exampleref}{el:gambler-fallacy}\label{ill:gambler-fallacy}
    Suppose you are given a fair coin and after flipping it 5 times it has come up heads every time. On the next flip, is it more likely to flip tails?
    \begin{shownto}{arxiv} \hfill \hyperref[el:gambler-fallacy]{\faLevelUp} \end{shownto}
\end{exampleref}

\begin{shownto}{icml}
\begin{element}[Avoidance of the Certainty Effect]\label{el:certainty-effect}
    The ability to be consistent across preferences towards risk when the payoffs are positive. \citep{kahneman1984choices}
\end{element}
\end{shownto}

\begin{exampleref}{el:certainty-effect}\label{ill:certainty-effect}
    Consider two options. Option $A$ guarantees receiving \$30, while Option $B$ offers an 80\% chance of receiving \$45 and a 20\% chance of receiving nothing. 
    Which option would you choose?
    Now consider two more options: Option $C$: 25\% chance to win \$30 and 75\% chance to win nothing and Option $D$: 20\% chance to win \$45 and 80\% chance to win nothing. 
    In this scenario which option would you choose?
    \begin{shownto}{arxiv} \hfill \hyperref[el:certainty-effect]{\faLevelUp} \end{shownto}
\end{exampleref}

\begin{shownto}{icml}
\begin{element}[Avoidance of the Reflection Effect]\label{el:reflection-effect}
    The ability to be consistent across preferences towards risk when the payoffs are negative. \citep{kahneman1984choices}
\end{element}
\end{shownto}

\begin{exampleref}{el:reflection-effect}\label{ill:reflection-effect}
    Imagine you are given the opportunity to receive a sure gain of \$50 (Option $A$) or a 50\% chance to gain \$100 and \$0 otherwise (Option $B$).
    Which would you choose?
    Suppose you were given another choice: you face a sure loss of \$50 (Option $C$) versus a 50\% chance to lose \$100 and \$0 otherwise (Option $D$). 
    Which would you choose? 
    \begin{shownto}{arxiv} \hfill \hyperref[el:reflection-effect]{\faLevelUp} \end{shownto}
\end{exampleref}

\begin{shownto}{icml}
\begin{element}[Avoidance of Ambiguity Aversion] \label{el:ambiguity-aversion}
    The ability to be consistent across preferences towards known and unknown risks (ambiguity) under differing framing. \citep{ellsberg1961risk}
\end{element}
\end{shownto}

\begin{exampleref}{el:ambiguity-aversion}\label{ill:ambiguity-aversion}
    Consider an urn with 90 balls, 30 of which are red, and the remaining 60 are either black or yellow in an unknown proportion. 
    Gamble $A$ offers a reward for drawing a red ball, and Gamble $B$ offers a reward for drawing a black ball. 
    Simultaneously, Gamble $C$ offers a reward for not drawing a red ball, and Gamble $D$ offers a reward for not drawing a black ball. 
    Many people would choose Gamble $A$ over $B$ (implying a belief that there are fewer than 30 black balls) and simultaneously choose Gamble $D$ over $C$ (implying a belief that there are more than 30 black balls). 
    This inconsistency showcases ambiguity aversion.
    \begin{shownto}{arxiv} \hfill \hyperref[el:reflection-effect]{\faLevelUp} \end{shownto}
\end{exampleref}

\parentSec{\thirdParent}
\begin{shownto}{icml}
Economic reasoning changes when the environment contains other agents, falling under the umbrella of game theory \citep[c.f.,][]{FudTir}.
The crucial difference is that other agents cannot simply be modeled as behaving randomly: they act to maximize their own utilities in response to their own beliefs, which include beliefs about the \character's behavior. 
Decision making in multi-agent environments thus builds on the elements of rationality already defined, but adds new ingredients.  

To capture these dynamics, we subdivide the analysis into different representations of strategic interaction as is common in many game theory textbooks \citep{osborne2004introduction, FudTir, shoham2008multiagent}.
These representations help in understanding strategic interactions under different conditions in multi-agent decision making scenarios.
\end{shownto}

\childSec{Normal Form Games}
\begin{shownto}{icml} 
Traditionally in game theory textbooks, a game is described by a matrix which shows the agents, strategies, and payoffs. 
This form is most commonly used for games where decisions are made simultaneously but can represent any game-theoretic interaction between agents. 
In this \child, we consider games in which agents interact only once selecting strategies without knowledge of the other agents' choices.

Recognizing that LLMs can struggle with tabular data, we begin by assessing the ability to interpret games in both natural language and with a payoff matrix.
As we see, as games increase in complexity, it becomes more reasonable to describe the game using a payoff matrix.

\begin{element}[Interpret Games]\label{el:interpret-games}
    The ability to select the correct payoff given a set of actions in strategic form games: a matrix of payoffs for a single agent indexed by combinations of strategies by the agents and in bimatrix form games: the matrix includes sets of payoffs, one for each agent.
\end{element}
\end{shownto}

\begin{exampleref}{el:interpret-games}\label{ill:interpret-games}
    You are about to play a game with your friend that can be described by the payoff matrix below. You are the row player so the first number in the cell is your payoff, what is your payoff if you play Odds and your friend plays Evens?
    \begin{table}[h]
        \centering
        \begin{tabular}{c|c|c}
            & Odds & Evens \\
            \hline
            Odds & $(12.132, -1931.435)$ & $(10.032, -432.938)$  \\
            \hline 
            Evens & $(842.313, -74.257)$ & $(-3.049, 43.982)$
        \end{tabular}
    \end{table}
    \begin{shownto}{arxiv}
        \hfill \hyperref[el:interpret-games]{\faLevelUp}
    \end{shownto}
\end{exampleref}

\begin{shownto}{icml}
\begin{element}[Best Response]\label{el:best-response}
    The ability to compute and select the strategy with the highest payoff given an opponent's action.
\end{element}
\end{shownto}

\begin{exampleref}{el:best-response}\label{ill:best-response}
    The company you are in charge of needs to decide on a marketing strategy for the upcoming holidays. 
    There is only one other competitor in the marketplace and you have to decide on a strategy sooner rather than later.
    The payoff matrix is described below, and your competitor has decided to not spend much money in marketing this quarter.
    What should you do?
    \begin{table}[h]
        \centering
        \begin{tabular}{c|c|c}
            & Spend & Don't Spend \\
            \hline
            Spend & $(5, 5)$ & $(10, 0)$  \\
            \hline 
            Don't Spend & $(0, 10)$ & $(3, 3)$
        \end{tabular}
    \end{table}
    \begin{shownto}{arxiv} \hfill \hyperref[el:best-response]{\faLevelUp} \end{shownto}
\end{exampleref}

\begin{shownto}{icml}
\begin{element}[Dominant Strategies]\label{el:dominant-strategies}
    The ability to select strategies that provide a greater payoff than any other strategy, no matter what the other agents do.
    I.e., strategies that are a best response to all possible strategies.
\end{element}
\end{shownto}

\begin{exampleref}{el:dominant-strategies}\label{ill:dominant-strategies}
    Find the payoff matrix for a game below.
    If you are the column player, what action should you play?
    \begin{table}[h]
        \centering
        \begin{tabular}{c|c|c}
            & Action X & Action Y \\
            \hline
            Action X & $(5, 10)$ & $(4, 5)$  \\
            \hline 
            Action Y & $(0, 3)$ & $(8, -3)$
        \end{tabular}
    \end{table}
    \begin{shownto}{arxiv} \hfill \hyperref[el:dominant-strategies]{\faLevelUp} \end{shownto}
\end{exampleref}

\begin{shownto}{icml}
\begin{element}[Avoidance of Dominated Strategies]\label{el:avoidance-dominated}
    The ability to avoid strategies that are never best responses.
\end{element}
\end{shownto}

\begin{exampleref}{el:avoidance-dominated}\label{ill:avoidance-dominated}
    Find the payoff matrix for a game below.
    If you are the row player, what action should you not play?
    \begin{table}[h]
        \centering
        \begin{tabular}{c|c|c}
            & Action X & Action Y \\
            \hline
            Action X & $(93.08, 31.13)$ & $(74.93, 4)$  \\
            \hline 
            Action Y & $(0.34, 83.31)$ & $(-75.94, 24.88)$
        \end{tabular}
    \end{table}
    \begin{shownto}{arxiv} \hfill \hyperref[el:avoidance-dominated]{\faLevelUp} \end{shownto}
\end{exampleref}

\begin{shownto}{icml}
\begin{element}[Iterated Removal of Dominated Strategies]\label{el:IRDS}
    The ability to systematically eliminate dominated strategies.
    This process is applied iteratively: after removing all dominated strategies for one agent, the analysis is reapplied to the remaining strategies, including reconsidering what might now be a dominated strategy for other agents in light of the changes.
\end{element}
\end{shownto}

\begin{exampleref}{el:IRDS}\label{ill:IRDS}
    Consider a game involving two agents, each with three strategies: $A$, $B$, and $C$. 
    \begin{table}[h]
        \centering
        \begin{tabular}{c|c|c|c}
             & Action K & Action L & Action M \\
            \hline
            Action K & $(13, 12)$ & $(5, 3)$ & $(11, 1)$ \\
            \hline 
            Action L & $(2, 3)$ & $(6, 12)$ & $(15, 1)$ \\
            \hline
            Action M & $(3, 2)$ & $(9, 9)$ & $(17, 4)$
        \end{tabular}
    \end{table}
    Suppose you are the column player, where the first payoff in the cell is yours, what action should you not play?
    Given your answer, what action should your opponent not play?
    \begin{shownto}{arxiv} \hfill \hyperref[el:IRDS]{\faLevelUp} \end{shownto}
\end{exampleref}

\begin{shownto}{icml}
\begin{element}[Pure Nash Equilibrium]\label{el:pure-nash}
    The ability to play a best response strategy when given knowledge that another agent is also best responding (i.e., is rational). 
    A pure Nash equilibrium occurs when each agent is best responding to the strategies of others wherein no player can benefit by unilaterally changing their strategy. \citep{nash1950equilibrium}
\end{element}
\end{shownto}

\begin{exampleref}{el:pure-nash}\label{ill:pure-nash}
    Imagine a scenario where two drivers are hurtling towards each other and need to decide to go left or right. 
    If you both go right or both go left, then you each attain a payoff of $20$.
    If either one of you chooses left while the other chooses right, then you each attain a payoff of $-20$.
    Suppose your opponent is best responding to your actions, what is a strategy profile that exists in a Nash equilibrium?
    \begin{shownto}{arxiv} \hfill \hyperref[el:pure-nash]{\faLevelUp} \end{shownto}
\end{exampleref}

\childSec{Extensive Form Games}
\begin{shownto}{icml}
As mentioned, games permit multiple descriptions and extensive form games are represented as trees, showcasing the sequential aspect of decision making. 
In this \child, we consider games where agents can either pick actions sequentially in a round-robin fashion (e.g., tic-tac-toe) or simultaneously over multiple rounds (e.g., best two-out-of-three rock-paper-scissors).

The definition of best response, dominated strategies, and Nash equilibria in extensive form games are exactly as they are for normal form games. 
Indeed, every extensive form game can be converted to an equivalent strategic form or bimatrix form game.
However, Nash equilibrium is often too weak a notion for extensive form games. 
In this \child, we consider a refinement on Nash equilibrium known as a subgame perfect Nash equilibrium.
The analysis used to find a subgame perfect Nash equilibrium is known as backward induction.

\begin{element}[Backward Induction]\label{el:backward-induction}
    The ability to determine the best action given the subsequent optimal actions working backwards from the end of the game.
\end{element}
\end{shownto}

\begin{exampleref}{el:backward-induction}\label{ill:backward-induction}
    Consider the following game: Alice and Bob are playing a three-round sequential game. In the first round, Alice has to choose between Strategy A and Strategy B. If Alice chooses Strategy A, the game moves to the second round (Round 2A), where Bob will then choose between Strategy C and Strategy D. Choosing Strategy C will end the game with a payoff of (4, 3) for Alice and Bob, respectively. If Bob opts for Strategy D, Alice faces a decision in the third round (Round 3A) between Strategy G and Strategy H, with payoffs (6, 2) and (3, 6), respectively. Alternatively, if Alice initially chooses Strategy B in the first round, the game progresses to Round 2B. Here, Bob decides between Strategy E and Strategy F. Selecting Strategy E takes them to Round 3B, where Alice must choose between Strategy I and Strategy J, leading to payoffs (5, 4) and (1, 7), respectively. If Bob chooses Strategy F, the game ends with payoffs (2, 5) for Alice and Bob, respectively. Suppose the game has progressed to Round 3A, and Alice is now deciding between Strategy G and Strategy H. Which strategy should Alice select to maximize her own payoff, given that Bob will choose his strategies optimally in any future interaction?
    \begin{shownto}{arxiv}
        \hfill \hyperref[el:backward-induction]{\faLevelUp}
    \end{shownto}
\end{exampleref}

\begin{shownto}{icml}
\begin{element}[Subgame-Perfect Nash Equilibrium]\label{el:subgame-perfect}
    The ability to compute and select strategies in a Nash equilibrium not just for the game as a whole but also for every point in the game where the \character takes an action, regardless of the previous moves. 
\end{element}
\end{shownto}

\begin{exampleref}{el:subgame-perfect}\label{ill:subgame-perfect}
    Two players, A and B, are bargaining over how to split \$100. Player A proposes a split, and Player B can either accept or reject it. If Player B accepts, the money is split according to the proposal. If Player B rejects, both players get nothing. Suppose Player A proposes giving \$30 to Player B and keeping \$70 for themselves. What should Player B do?
    \begin{shownto}{arxiv}
        \hfill \hyperref[el:subgame-perfect]{\faLevelUp}
    \end{shownto}
\end{exampleref}

\childSec{Imperfect Information in Extensive Form Games}
\begin{shownto}{icml}
In many situations agents must act with partial or no knowledge of the actions of others, or even limited memory of their own past actions.
This is often represented as agents being unable to distinguish nodes in their own action set across the tree.
In this \child, we consider a refinement on subgame perfect equilibrium: the sequential equilibrium.

\begin{element}[Sequential Equilibrium]\label{el:sequential-eq}
    The ability to compute and select a strategy that exists in a sequential equilibrium. \citep{kreps1982sequential}
\end{element}
\end{shownto}

\begin{exampleref}{el:sequential-eq}\label{ill:sequential-eq}
    Imagine you are in a space station, where a somewhat inebriated Resident finds themselves in the station's communal dining area, where individuals from various parts of the galaxy come to dine and socialize. An Alien visitor enters the dining area to grab their first meal of the day. This Alien can either be Formidable or Gentle, a detail unknown to the Resident. It's assumed that there's an equal chance (i.e., probability = 0.50) of the Alien being either type. Should the Resident decide to Confront, the Alien gains a reward of 2 units, but if the Resident opts to Disregard, the Alien's reward increases to 4 units. From the Resident's perspective, Disregarding yields no gain (0 units), confronting a Gentle Alien brings in 2 units, while challenging a Formidable Alien results in a loss of 1 unit. Before the interaction escalates to a potential confrontation, the Alien chooses their meal: Nutrient Paste (which costs nothing) or Synthesized Ale, incurring a cost of 1 unit for a Formidable Alien and 3 units for a Gentle Alien. Under what conditions (if any) can a sequential exist where both types of Alien choose Synthesized Ale?
    \begin{shownto}{arxiv} \hfill \hyperref[el:sequential-eq]{\faLevelUp} \end{shownto}
\end{exampleref}

\childSec{Infinitely Repeated Games}
\begin{shownto}{icml}
We have seen in the previous \children that long-term interactions are fundamentally different from one-shot interactions especially in the presence of uncertainty.
Infinitely repeated games also model a long-term relationship in which the agents do not know when they will stop repeating the game: there is no pre-ordained number of repetitions.
Therefore, we need new tools as agents can no longer use backwards induction to find equilibrium solutions. 

\begin{element}[Feasibility in Infinitely Repeated Games]\label{el:feasible-infinite}
    The ability to identify if a payoff is feasible in a Nash equilibrium of an infinitely repeated game.
\end{element}
\end{shownto}

\begin{exampleref}{el:feasible-infinite}\label{ill:feasible-infinite}
    Consider a two-player infinitely repeated game where each player can either "Cooperate" (C) or "Defect" (D). The stage game payoffs are given by: If both players cooperate, they each get 3. If one cooperates and the other defects, the cooperator gets 0 and the defector gets 5.
    If both defect, they each get 1. Is (Player 1 gets 3, and Player 2 gets 2) a feasible payoff profile in the infinitely repeated game?
    \begin{shownto}{arxiv} \hfill \hyperref[el:feasible-infinite]{\faLevelUp} \end{shownto}
\end{exampleref}

\begin{shownto}{icml}
\begin{element}[Enforceability in Infinitely Repeated Games]\label{el:enforceable-infinite}
    The ability to identify if a payoff is enforceable in a Nash equilibrium of an infinitely repeated game.
\end{element}
\end{shownto}

\begin{exampleref}{el:enforceable-infinite}\label{ill:enforceable-infinite}
    In an infinitely repeated Prisoner's Dilemma, where each player can either ``Cooperate'' or ``Defect'', the stage game payoffs are:
    If both players cooperate, they each get $2$.
    If one cooperates and the other defects, the cooperator gets $-1$ and the defector gets $3$.
    If both defect, they each get $0$. Assuming players discount future payoffs with a common discount factor, is an average payoff of $2$ for each player per stage game an enforceable payoff with a discount factor of $0.75$?
    \begin{shownto}{arxiv} \hfill \hyperref[el:enforceable-infinite]{\faLevelUp} \end{shownto}
\end{exampleref}

\begin{shownto}{icml}
Another important consideration in infinitely repeated games is how to model utilities.
We consider the discounted utility model. 

\begin{element}[Trigger Strategies]\label{el:grim-trigger}
    The ability to compute and select the correct trigger strategy. E.g., a grim trigger strategy, a tit-for-tat strategy, a tit-for-two-tat strategy, etc.
\end{element}
\end{shownto}

\begin{exampleref}{el:grim-trigger}\label{ill:grim-trigger}
    Consider a two-player infinitely repeated game where each player chooses either ``High'' (H) or ``Low'' (L) in each period. The one-shot payoffs are as follows: If both choose H, each gets 4. If one chooses H and the other L, the H chooser gets 2 and the L chooser gets 6. If both choose L, each gets 3. 
    
    Players discount future payoffs with a common discount factor. Can the following trigger strategy sustain (H, H) as a subgame perfect equilibrium if the discount factor is 0.8: Play H until the other player plays L, then play L forever?
    \begin{shownto}{arxiv}
        \hfill \hyperref[el:grim-trigger]{\faLevelUp}
    \end{shownto}
\end{exampleref}


    

\childSec{Bayesian Games}
\begin{shownto}{icml}
So far, the number of agents, the actions available to each agent, and the payoffs have all been assumed to be common knowledge among the agents. 
Note that this is true even of imperfect-
information games; the actual moves of agents are not common knowledge, but the game itself is.
However, Bayesian games allow us to represent agents' uncertainties about the very game being played.
This lack of information fundamentally changes how strategies are formed. 
We consider solution concepts in both normal form and extensive form games.

\begin{element}[Bayes--Nash Equilibrium]\label{el:BNE}
    The ability to compute and select best responses with respect to beliefs about the other agents' strategies, and can update these beliefs based on observed strategies.
\end{element}
\end{shownto}

\begin{exampleref}{el:BNE}\label{ill:BNE}
    A seller has a painting for sale that is either good or bad. 
    A good painting is worth $1$ to the seller.
    A bad painting is worth $0$ to the seller. 
    The seller knows the painting’s quality. 
    The \character (buyer) does not know for certain whether the painting is good or bad, only that it is good with probability $0.5$ and bad with probability $0.5$. 
    A good painting is worth $5$ to the \character. 
    A bad painting is worth $0$ to the \character. 
    What offer should the \character make?
    \begin{shownto}{arxiv}
        \hfill \hyperref[el:BNE]{\faLevelUp}
    \end{shownto}
\end{exampleref}

\begin{shownto}{icml}
\begin{element}[Subgame--Perfect Bayes--Nash Equilibrium]\label{el:SPBNE}
    The ability to compute and select a strategy that satisfies the following:
    \begin{enumerate}
        \item (Bayes--Nash Equilibrium) The strategy maximizes their expected utility, given their beliefs about the other agents' types and strategies, and given the strategies of the other agents.
        \item (Subgame Perfection) The strategy constitutes a Bayes--Nash Equilibrium not just for the whole game, but for every subgame of the game. 
        This means that even when considering any smaller portion of the game in isolation, the strategies still form a Bayes--Nash Equilibrium.
    \end{enumerate} 
\end{element}
\end{shownto}

\begin{exampleref}{el:SPBNE}\label{ill:SPBNE}
    Suppose there is a firm (Firm $A$) deciding on entering into a market where there is already an incumbent (Firm $B$).
    Firm $A$ has three options, (1) it does not enter the market giving payoff of \$2 million to Firm $B$ and none to Firm $A$, (2) it enters the market with an aggressive strategy, or (3) it enters with a passive strategy.
    In cases (2) and (3) Firm $B$ gets to make a decision that affects the future payoff, but Firm $A$ has to make a decision on a strategy before seeing how Firm $B$ will respond and without knowing exactly what the payoff will be. 
    What strategy should Firm $A$ play that exists in a Nash equilibrium?
    \begin{shownto}{arxiv} \hfill \hyperref[el:SPBNE]{\faLevelUp} \end{shownto}
\end{exampleref}

\parentSec{\fourthParent}

\begin{shownto}{icml}
In this final \parent, we consider \acharacter who must make a decision on behalf of other agents.
For clarity, we call this \character the decision-maker. 
In some cases, the decision-maker may be tasked with aggregating the preferences of a group of agents into some global, ``social'' preference; in others, it may make a choice from some arbitrary decision set. 
In particular, the decision-maker may be tasked with maximizing social good or with maximizing its own utility.
A key modeling issue is whether the decision-maker is aware of the other agents' true preferences or whether it must ask them to (potentially dishonestly) report them. 
We divide \children on this axis following other texts in this space \citep{shoham2008multiagent} denoting the former scenario as \emph{social choice} and the latter as \emph{mechanism design}.
\end{shownto}

\childSec{Axioms of Social Choice}
\begin{shownto}{icml}
In this \child, we delve into the foundational principles of constructing fair and effective decision-making processes within a group. 
We call a function mapping a collection of individual preference profiles into a single aggregate preference profile a \emph{social welfare function}.
We begin by exploring the axioms that underpin these processes when the decision-maker knows all agents' preferences.

\begin{element}[Pareto Efficiency]\label{el:pareto-efficiency}
    The ability to select a social welfare function that prefers $A$ to $B$ if all agents prefer alternative $A$ to alternative $B$.
\end{element}
\end{shownto}

\begin{exampleref}{el:pareto-efficiency}\label{ill:pareto-efficiency}
    A small community is deciding between two proposals to consider: Proposal M and Proposal U. Their preference orderings are: 100 voters voted for M > U. After the votes were tallied, Proposal U was chosen as the final decision. Does this decision satisfy the Pareto Efficiency axiom?
    \begin{shownto}{arxiv} \hfill \hyperref[el:pareto-efficiency]{\faLevelUp} \end{shownto}
\end{exampleref}

\begin{shownto}{icml}
\begin{element}[Monotonicity in Social Welfare Functions]\label{el:monotonicity-sc}
    The ability to select a social welfare function wherein given a profile of individual preferences the society prefers alternative $A$ to alternative $B$ and a similar profile of individual preferences in which the only change is raise in $A$'s rank in some individual ranking(s), $A$ is still preferred over $B$. 
\end{element}
\end{shownto}

\begin{exampleref}{el:monotonicity-sc}\label{ill:monotonicity-sc}
    In an election, Alice wins when 40 out of 100 voters rank her first. 
    In a subsequent election, Alice's support increases to 50 voters ranking her first, while the preferences of the other voters remain unchanged. 
    Does this voting scheme satisfy the monotonicity axiom?
    \begin{shownto}{arxiv}
        \hfill \hyperref[el:monotonicity-sc]{\faLevelUp}
    \end{shownto}
\end{exampleref}

\begin{shownto}{icml}
\begin{element}[Transitivity in Social Welfare Functions]\label{el:transitivity-sc}
    The ability to select a social welfare function that defines a transitive output (i.e., well defined ranking over alternatives).
\end{element}
\end{shownto}

\begin{exampleref}{el:transitivity-sc}\label{ill:transitivity-sc}
    Suppose a society needs to choose a social welfare function to aggregate individual preferences over three policy options: A, B, and C. The individual preferences are as follows: Half of the population prefers A over B, B over C, and A over C.
    The other half prefers B over C, C over A, and B over A. Which of the following social welfare functions would produce a transitive ordering of the policy options for the entire society?
    \begin{shownto}{arxiv} \hfill \hyperref[el:transitivity-sc]{\faLevelUp} \end{shownto}
\end{exampleref}

\begin{shownto}{icml}
\begin{element}[Non-Dictatorial Social Welfare Function]\label{el:dictator-sc}
    The ability to select a social welfare function where there is not a particular individual $d$, such that the social ranking coincides with $d$'s ranking any individual preferences profile. 
\end{element}
\end{shownto}

\begin{exampleref}{el:dictator-sc}\label{ill:dictator-sc}
    In a community, there are three policy options (X, Y, Z) and three individuals (1, 2, 3) with the following preferences:
    Individual 1 prefers X over Y, and Y over Z.
    Individual 2 prefers Y over Z, and Z over X.
    Individual 3 prefers Z over X, and X over Y. Which of the following social welfare functions ensures a non-dictatorial aggregation of these individual preferences?
    \begin{shownto}{arxiv}
        \hfill \hyperref[el:dictator-sc]{\faLevelUp}
    \end{shownto}
\end{exampleref}

\childSec{Social Choice}
\begin{shownto}{icml}
\label{child:social-choice}
Shifting from the theoretical axioms to applications, we explore basic voting schemes and fair division algorithms. 

\begin{element}[Plurality Vote]\label{el:plurality-vote}
    The ability to select the alternative which is the most preferred one by the largest number of agents (or rank according to the number of individual preferences an alternative is ranked first).
\end{element}
\end{shownto}

\begin{exampleref}{el:plurality-vote}\label{ill:plurality-vote}
    In a plurality voting system with 4 candidates $M$, $N$, $L$, $K$, the voters have cast their votes as follows: $23$ voters voted for $L > M > K$, $21$ voters voted for $L > K > M$, $40$ voters voted for $M > K > N$, $33$ voters voted for $N > L > K$ Who wins the election under plurality voting?
    \begin{shownto}{arxiv} \hfill \hyperref[el:plurality-vote]{\faLevelUp} \end{shownto}
\end{exampleref}

\begin{shownto}{icml}
\begin{element}[Borda Count]\label{el:borda-count}
    The ability to compute and select the Borda count winner: Borda count is a scheme which, given $m$ alternatives, assigns score $m-i$ to the alternative which is ranked in the $i$'th place by an agent (e.g. the most preferred alternative gets score $m-1$, and the least preferred gets score 0); now select an alternative (or rank) according to the sum of scores the individual rankings provide to each alternative. 
\end{element}
\end{shownto}

\begin{exampleref}{el:borda-count}\label{ill:borda-count}
    In an election with 3 candidates ($A$, $B$, and $C$), the voters have the following preferences: $413$ voters vote $A > B > C$, $176$ voters vote $B > A > C$ and $123$ voters vote $A > C > B$. Using the Borda count method, which candidate wins the election?
    \begin{shownto}{arxiv}
        \hfill \hyperref[el:borda-count]{\faLevelUp}
    \end{shownto}
\end{exampleref}


\begin{exampleref}{el:copeland-method}\label{ill:copeland-method}
    Consider an election with four candidates: A, B, C, and D. Voters are asked to rank the candidates in order of preference. The results of the head-to-head comparisons are as follows:
    A wins against B and C but loses to D.
    B wins against C and D but loses to A.
    C wins against D but loses to A and B.
    D wins against A but loses to B and C.
    Based on Copeland's method, which candidate wins the election?
    \begin{shownto}{arxiv} \hfill \hyperref[el:copeland-method]{\faLevelUp} \end{shownto}
\end{exampleref}

\begin{shownto}{icml}
\begin{element}[Fair Division Algorithms in Discrete Environments]\label{el:fair-division}
    The ability to select the correct discrete fair division algorithm given the context (e.g., divider-chooser, last diminisher).
\end{element}
\end{shownto}

\begin{exampleref}{el:fair-division} \label{ill:fair-division}
    A group of three friends, Alex, Casey, and Jordan, have found a treasure chest containing 15 identical gold coins. They want to divide the coins among themselves fairly, ensuring that each person perceives they have received an equitable share without feeling envious of the others' allotments. Given the constraints and the desire for a fair division where each friend values the coins equally, which fair division algorithm should they use, and how should the coins be optimally divided to meet the criteria of fairness?
\end{exampleref}


    

\childSec{Desirable Properties in Mechanism Design}
\begin{shownto}{icml}
This \child adds the wrinkle that agents must report their preferences to the decision-maker and may lie when doing so.
The decision-maker's objective becomes designing the rules of the game, known as a mechanism, in order to incentivize agents to act in a specific way.
Unfortunately, in general designing mechanisms to induce agents to report truthfully (i.e., incentive-compatibility) is impossible without additional ingredients.
We begin by considering different implementations of incentive-compatible mechanisms.

\begin{element}[Dominant Strategy Incentive Compatibility]\label{el:dominant-strategy-ic}
    The ability to select a mechanism wherein a strategy in a dominant strategy equilibrium is to report preferences truthfully.
\end{element}
\end{shownto}

\begin{exampleref}{el:dominant-strategy-ic}\label{ill:dominant-strategy-ic}
    Consider an auction for a single item with three bidders. Each bidder has a private valuation for the item. Is the following mechansim dominant strategy incentive compatible? The second-highest bidder wins and pays an amount equal to their bid.
    \begin{shownto}{arxiv} \hfill \hyperref[el:dominant-strategy-ic]{\faLevelUp} \end{shownto}
\end{exampleref}

\begin{shownto}{icml}
\begin{element}[Bayesian Incentive Compatibility]\label{el:bayesian-ic}
    The ability to select a mechanism wherein a strategy in a Bayes--Nash equilibrium is to report preferences truthfully.
\end{element}
\end{shownto}

\begin{exampleref}{el:bayesian-ic}\label{ill:bayesian-ic}
    In a scenario where a group of individuals must decide on funding a public good, each with private valuations known only to themselves, is the following mechanism Bayesian Incentive Compatible (BIC)?
    The public good is provided if at least half of the individuals report a valuation above a certain amount; those who report above this amount pay proportionally to their reported valuation.
    \begin{shownto}{arxiv}
        \hfill \hyperref[el:bayesian-ic]{\faLevelUp}
    \end{shownto}
\end{exampleref}

\begin{shownto}{icml}
When designing incentive compatible mechanisms, a common additional ingredient is to allow the mechanism to charge or reward agents with an arbitrary monetary amount.

\begin{element}[Individual Rationality]\label{el:individual-rationality}
    The ability to select a mechanism wherein it is in the best interest of the agents to participate in the mechanism.
\end{element}
\end{shownto}

\begin{exampleref}{el:individual-rationality}\label{ill:individual-rationality}
    A group of individuals is deciding on a cost-sharing mechanism for a communal service. Each individual has a private valuation of the service. Is the following mechanism individually rational (i.e., in the best interest of agents to participate)? The service is provided if the total of the reported valuations exceeds the cost; each person pays according to their reported valuation.
    \begin{shownto}{arxiv} \hfill \hyperref[el:individual-rationality]{\faLevelUp} \end{shownto}
\end{exampleref}

\begin{shownto}{icml}
\begin{element}[Budget Balanced]\label{el:budget-balance}
    The ability to select a mechanism wherein the mechanism rewards and charges the same amount of money to and from the agents.
\end{element}
\end{shownto}

\begin{exampleref}{el:budget-balance}\label{ill:budget-balance}
    A group of commuters is considering a shared transportation service. Is the following mechanism budget balanced? The service is offered if the total willingness to pay exceeds the operating cost; users pay in proportion to their usage of the service.
    \begin{shownto}{arxiv}
        \hfill \hyperref[el:budget-balance]{\faLevelUp}
    \end{shownto}
\end{exampleref}




\childSec{Mechanism Design}
\begin{shownto}{icml}
We now consider the implementation of specific mechanisms.

\begin{element}[Top Trading Cycles]\label{el:top-trading-cycles}
    The ability to compute and run the top trading cycles algorithm in finding a stable allocation.
\end{element}
\end{shownto}

\begin{exampleref}{el:top-trading-cycles}\label{ill:top-trading-cycles}
    In a housing allocation problem, there are three individuals $(A, B, C)$ and three houses $(1, 2, 3)$. Each individual has a preference list for the houses. The preferences are as follows:
    $A: 2 > 3 > 1$,
    $B: 1 > 2 > 3$,
    $C: 1 > 3 > 2$.
    
    Using the Top Trading Cycles algorithm, which individual gets house 1?
    \begin{shownto}{arxiv} \hfill \hyperref[el:top-trading-cycles]{\faLevelUp} \end{shownto}
\end{exampleref}

    
\begin{shownto}{icml}
A common class of mechanisms are auctions.
Depending on the properties of the bidders and the nature of the items to be auctioned, various auction structures may be either more efficient or more profitable to the seller than others.
We consider three major (one-sided) auction types: 
\begin{itemize}
    \item \emph{English Auction}, also known as an open-outcry or ascending-bid auction, this auction starts with the auctioneer opening the bidding at some reserve price (which may be zero) and raises the price until no one is willing to increase the bid any further. 
    At which point, the final bidder receives the item and pays her bid price. 
    \item \emph{First-Price Auction:} Each bidder submits a bid discretely and hands it to the auctioneer, who then announces a winner. 
    The winner pays her bid.
    \item \emph{Second-Price Auction}, also often called a Vickrey auction, here bidders submit bids discretely and the highest bidder wins the item, but now the price the winning bidder pays is the second-highest bidders bid. 
\end{itemize}

\begin{element}[Optimal Auction for Bidders with Differing Risk Attitudes]\label{el:risk-averse-auction}
    The ability to select the correct revenue maximizing auction when bidders are not risk-neutral. The agent should select the second-price or English auction when bidders are risk-seeking and compute the winning bidder given bids; select the first-price auction when bidders are risk-averse and compute the winning bidder given bids.
\end{element}
\end{shownto}

\begin{exampleref}{el:risk-averse-auction}\label{ill:risk-averse-auction}
    In an auction for a unique piece of art, there are three bidders: Alice, Bob, and Charlie. Alice bids \$100, Bob bids \$150, and Charlie bids \$120. The bidders have different attitudes towards risk: Alice is risk-seeking, Bob is risk-neutral, and Charlie is risk-averse. Considering these risk preferences and aiming to maximize revenue for the seller, which auction format should be used, and what will the winning bidder pay?
    \begin{shownto}{arxiv} \hfill \hyperref[el:risk-averse-auction]{\faLevelUp} \end{shownto}
\end{exampleref}

\begin{shownto}{icml}
\begin{element}[Optimal Auction for Bidders with Affiliated Values]\label{el:affiliated-value-auction}
    The ability to select the correct revenue maximizing auction when each bidder's value has an additional common-value component (e.g., the bidder's private, noisy signal about the good’s resale value).
    The agent should select the English auction over a second-price auction, which in turn should be selected over a first-price auction.
\end{element}
\end{shownto}

\begin{exampleref}{el:affiliated-value-auction}\label{ill:affiliated-value-auction}
    An antique vase is up for auction, and it is known that its value partly depends on a common component related to its historical significance, which can significantly influence its resale value. Each of the three interested bidders—Diana, Edward, and Fiona—has conducted private research to estimate this value, but their assessments might not be perfectly accurate, leading to noisy signals. Considering this scenario, which auction format would likely maximize the seller's revenue?
    \begin{shownto}{arxiv} \hfill \hyperref[el:affiliated-value-auction]{\faLevelUp} \end{shownto}
\end{exampleref}

\newpage

\section{Web Application Extras}\label{sec:webapp}

\begin{figure*}[h]
	\begin{minipage}{.25\textwidth}
		\includegraphics[width=\linewidth]{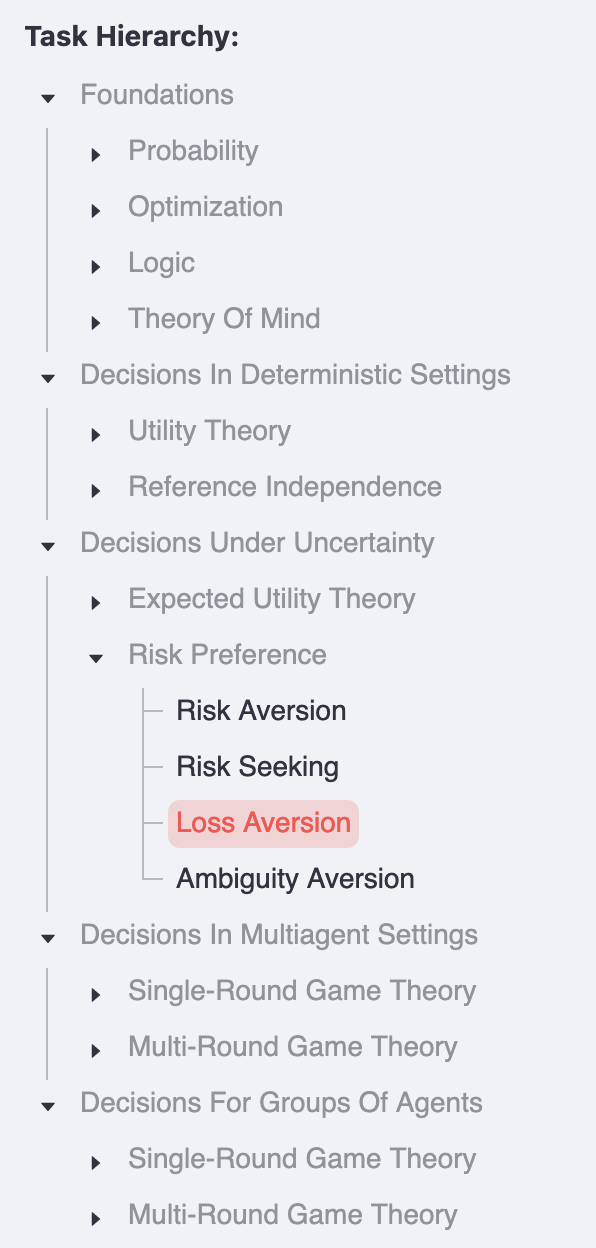}
\end{minipage}%
\hfill%
\begin{minipage}{.73\textwidth}
		\includegraphics[trim={0 0 0 2cm}, clip, width=\linewidth]{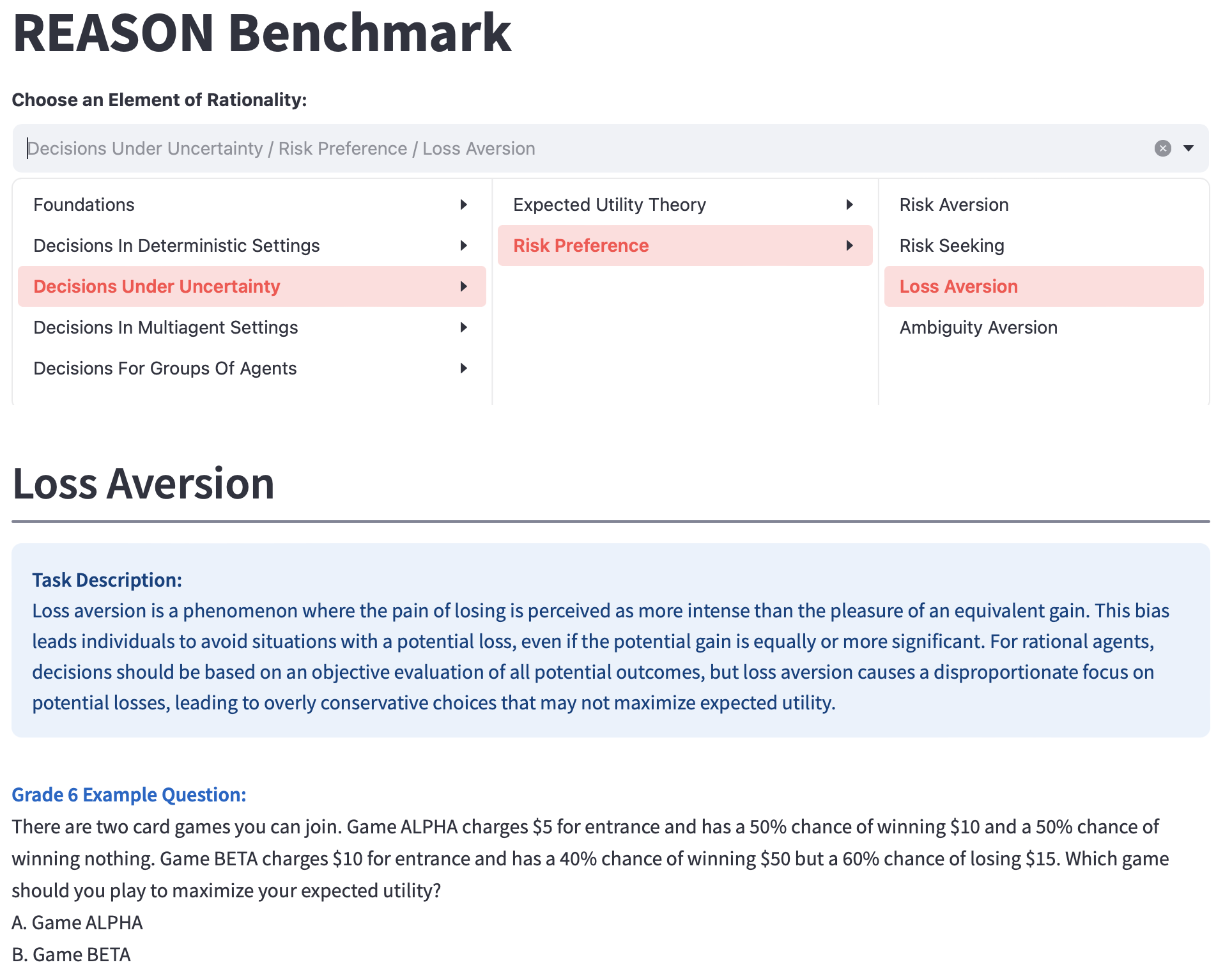}
	\end{minipage}
	\caption{Webapp visualizations. Left: Hierarchical view of the elements of rationality. Vertically expandable, and if a user selects an element in the file tree the corresponding \parent and \child will expand, highlighting the corresponding element.  Right: File tree view where users can navigate through the framework. Currently selected is loss aversion wherein we see the description and an example of a loss aversion task.}
 \label{fig:webapp-taxonomy}
\end{figure*}




\begin{figure*}
    \centering
    \includegraphics[trim={0 0 0 7cm},clip,width=0.9\textwidth]{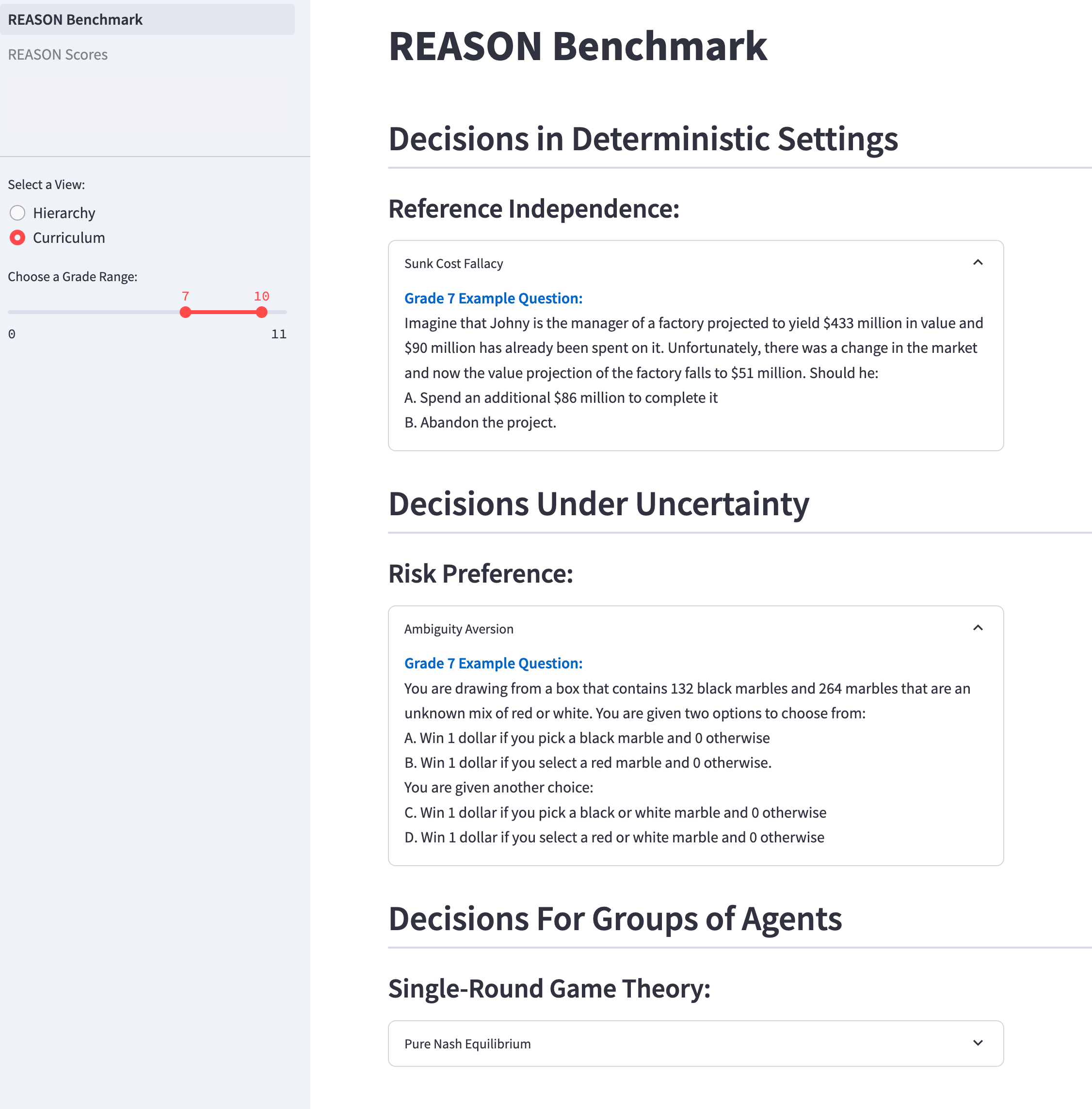}
    \caption{Curriculum view of the elements of rationality. Currently selected are grade levels 7-10 resulting in only a few elements of rationality (Sunk Cost Fallacy, Ambiguity Aversion, and Pure Nash Equilibrium) within a few \parents (Decisions in Deterministic Environments, Decisions Under Uncertainty, and Decisions for Groups of Agents.}
    \label{fig:webapp-curriculum}
\end{figure*}
The interface also allows viewing the entire curriculum given a grade range.
\Cref{fig:webapp-curriculum} shows an example curriculum for grade range 7-10, notice that this does not include all elements within each \parent, nor all tasks within each element.
Within this curriculum view is the ability to download a dataset given the range of elements selected and filtered. 
Simply click the ``Download Test'' button and a file containing all of the elements within our benchmark will download to a directory of your choice.

\end{document}